\documentclass[journal]{IEEEtran}
\ifCLASSINFOpdf
\else
\fi

\usepackage{multirow}
\usepackage{tabls}
\usepackage{wrapfig}

\usepackage{amsmath,amssymb} 
\usepackage{amsmath,amssymb}
 \usepackage[ruled,vlined,linesnumbered,resetcount]{algorithm2e}
 \usepackage[linesnumbered]{algorithm2e}
\usepackage{algorithmic}
\usepackage{caption}
\usepackage{subfigure}
\renewcommand{\algorithmicrequire}{ \textbf{Input:}} 
\renewcommand{\algorithmicensure}{ \textbf{Output:}} 
\usepackage[font=small,labelfont=bf]{caption}
   \usepackage{graphicx}
 \usepackage{booktabs}

\usepackage{epstopdf}
\usepackage{graphicx}

\usepackage{times}
\usepackage{epsfig}
\usepackage{graphicx}
\usepackage{amsmath}
\usepackage{amssymb,enumitem}

\usepackage{multirow}
\usepackage{tabls}
\usepackage{wrapfig}
\usepackage{hyperref}
\usepackage{amsmath,amssymb} 
\usepackage{amsmath,amssymb}
 \usepackage[ruled,vlined,linesnumbered,resetcount]{algorithm2e}
  \usepackage{booktabs}
 \usepackage[linesnumbered]{algorithm2e}
\usepackage{algorithmic}
\usepackage{caption}
\usepackage{subfigure}
\renewcommand{\algorithmicrequire}{ \textbf{Input:}} 
\renewcommand{\algorithmicensure}{ \textbf{Output:}} 
\usepackage[font=small,labelfont=bf]{caption}
   \usepackage{graphicx}
 \usepackage{booktabs}
\usepackage{epstopdf}
\usepackage{graphicx}
\usepackage{subfigure}
\usepackage{enumitem}
\usepackage{multirow}
\usepackage{tabls}
\usepackage{wrapfig}
\usepackage{hyperref}
\usepackage{amsmath,amssymb} 
\usepackage{amsmath,amssymb}
 \usepackage[ruled,vlined,linesnumbered,resetcount]{algorithm2e}
  \usepackage{booktabs}
 \usepackage[linesnumbered]{algorithm2e}
\usepackage{algorithmic}
\usepackage{caption}
\usepackage{subfigure}
\renewcommand{\algorithmicrequire}{ \textbf{Input:}} 
\renewcommand{\algorithmicensure}{ \textbf{Output:}} 
\usepackage[font=small,labelfont=bf]{caption}
   \usepackage{graphicx}
 \usepackage{booktabs}
\usepackage{epstopdf}
\usepackage{graphicx}
\usepackage{subfigure}
\usepackage{enumitem}
\usepackage{amsmath,amssymb} 
\usepackage{amsmath,amssymb}
\usepackage[ruled,vlined,linesnumbered,resetcount]{algorithm2e}
\usepackage[linesnumbered]{algorithm2e}
\usepackage{algorithmic}
\usepackage{caption}
\usepackage{subfigure}
\renewcommand{\algorithmicrequire}{ \textbf{Input:}} 
\renewcommand{\algorithmicensure}{ \textbf{Output:}} 
\usepackage[font=small,labelfont=bf]{caption}
\usepackage{graphicx}
\usepackage{booktabs}

\usepackage{epstopdf}
\usepackage{graphicx}

\begin{document}

\title{Joint Transmission Map Estimation and Dehazing using Deep Networks}

\author{He Zhang,~\IEEEmembership{Student Member, IEEE}, Vishwanath Sindagi,~\IEEEmembership{Student Member, IEEE}\\  
	Vishal M. Patel,~\IEEEmembership{Senior Member,~IEEE}
	\thanks{He Zhang was  with the Department of Electrical and Computer Engineering at Rutgers University, 
		Piscataway, NJ USA. email: he.zhang92@rutgers.edu}
	\thanks{Vishwanath Sindagi and Vishal M. Patel are with the Department of Electrical and Computer Engineering, Johns Hopkins University, Baltimore, MD, USA. Email: \{vsindag1, vpatel36\}@jhu.edu}
}
\maketitle

\begin{abstract}
Single image haze removal is an extremely challenging problem due to its inherent ill-posed nature.  Several prior-based and learning-based methods have been proposed in the literature to solve this problem and they have achieved visually appealing  results. However, most of the existing methods assume constant atmospheric light model and tend to follow a two-step procedure involving prior-based methods for estimating transmission map followed by calculation of dehazed image using the closed form solution. In this paper, we relax the constant atmospheric light assumption and propose a novel unified single image dehazing network that jointly estimates the transmission map and performs dehazing. In other words,  our new approach provides an end-to-end learning framework, where the inherent transmission map and dehazed result are learned jointly from the loss function. Extensive experiments evaluated on synthetic and real datasets with challenging hazy images demonstrate that the proposed method achieves significant improvements over the  state-of-the-art methods.

\end{abstract}
\section{Introduction}

\begin{figure}[t]
	\centering
	\begin{minipage}{.237\textwidth}
		\centering
		\includegraphics[width=4.3cm,height=3cm]{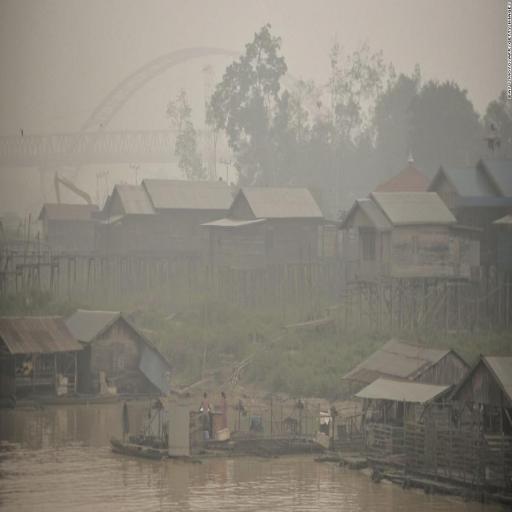}
		\captionsetup{labelformat=empty}
		\captionsetup{justification=centering}
	\end{minipage}
	\begin{minipage}{.237\textwidth}
		\centering
		\includegraphics[width=4.3cm,height=3cm]{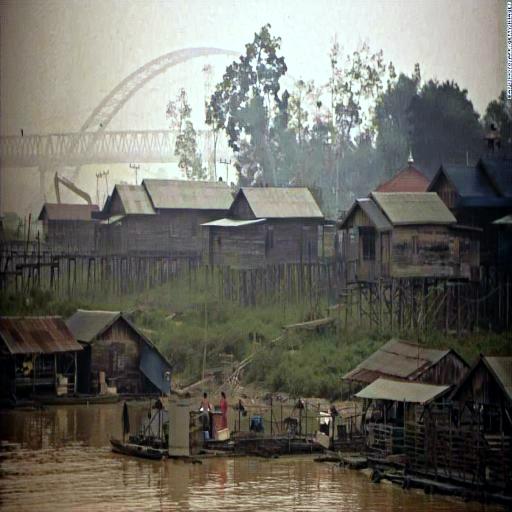}
		\captionsetup{labelformat=empty}
		\captionsetup{justification=centering}
	\end{minipage}	
		\vskip+4pt\caption{Sample image dehazing result using the proposed method. Left: Input hazy image.  Right: Dehazed result.} \label{fig:over}
\end{figure}

Haze is the obscuration of lower atmosphere, typically caused by the presence of suspended particles in the air such as dust, smoke and other dry particulates. The presence of haze usually reduces the visibility range, thus affecting quality of images captured by camera sensors that will be processed by computer vision systems. A sample hazy image is shown on the left side of Figure~\ref{fig:over}. It can be clearly observed that the existence of haze in an image greatly obscures the background scene.  The problem of estimating a clear image from a single hazy input image is commonly referred to as dehazing.   Image dehazing has attracted a significant interest in the computer vision and image processing communities in recent years \cite{dehaze_2016_eccv,dehaze_2016_tip,dehaze_2014_tang,dehaze_2016_dana,dehaz_2009_dark,dehaze_depth,dehaze_multi_image,dehaze_iccv17,yang2018towards,kim2013optimized,galdran2017fusion,wang2017fast, dehaze_2018, ren2019deep, zhang2018multi, li2018single, yang2018proximal, liu2019dual,xu2018strong, chen2019gated, sakaridis2018semantic}.  

The deterioration of image quality is captured by the following mathematical model \cite{Fattal_2008_siggraph}:
\begin{equation}
\mathbf{I}(\mathbf{x})=\mathbf{J(x)}t(\mathbf{x})+\mathbf{A(x)}(1-t(\mathbf{x})),
\label{eq: intro}
\end{equation}
where $\mathbf{x}$ is the location in the image co-ordinates, $\mathbf{I}$ represents the observed hazy image, $\mathbf{J}$ is the image before degradation, $\mathbf{A}$ is the global atmospheric light, and $t(\mathbf{x})$ is the transmission map. Transmission map contains the per-pixel attenuation information that affects the light reaching the camera sensor and it is a factor of depth as shown below:
\begin{equation}
	t\mathbf{(x)}=e^{-\beta d(\mathbf{x})},
	\label{eq: trans}
\end{equation}
where $\beta$ is attenuation coefficient of the atmosphere and $d\mathbf{(x)}$ is the depth map.  One can view \eqref{eq: intro} as the superposition of two components: 1. \emph{Direct attenuation} $\left(\mathbf{J(x)}t(\mathbf{x})\right)$, and 2. \emph{Airlight}  $\left(\mathbf{A(x)}(1-t(\mathbf{x})\right)$. Direct attenuation represents the effect of scattering of light and the eventual decay of light before it reaches the camera sensor. Airlight is a phenomenon that results from the scattering of environmental light causing a shift in the apparent brightness of the scene.  Note that Airlight is a function of scene depth and the global atmospheric light $\mathbf{A}$. As it can be observed from Eq. \ref{eq: intro}, image dehazing is an inherently ill-posed problem which has been addressed in different ways. Many previous methods overcome this issue by including extra prior assumption  such as multiple images of the same scene \cite{dehaze_multi_image} or depth information \cite{dehaze_depth} to determine a solution.  However, no extra information such as depth or multiple images is available for the problem of single image dehazing. {To tackle this issue, different prior information has to be considered into the optimization framework such as dark-channel prior \cite{dehaz_2009_dark}, contrast color-lines \cite{dehaze_2014acm}] and hazeline prior \cite{dehaze_2016_dana}. For example, based on the observation that there always exists one channel that is significant dark in the captured outdoor images, dark-channel prior \cite{dehaz_2009_dark} is leveraged in the optimization framework to guarantee  dehazed images are ``dark-channel".  Different from dark-channel prior, \cite{dehaze_2016_dana} leverage the haze-line prior in the framework, based on the observation that color cluster in the clear image can be approximated as the haze-line  in RGB space.}  More recently, several learning-based methods have also been proposed, where different learning algorithms such as random forest regression and Convolutional Neural Networks (CNNs) are trained for predicting the transmission map \cite{dehaze_2014_tang,dehaze_2016_eccv,dehaze_2016_tip,dehaze_iccv17}.  Many existing methods make an important assumption of constant atmospheric light \footnote{Meaning that the intensity of atmosphere light $\mathbf{A}$ is independent from its spatial location $\mathbf{x}$.} in the image degradation model \eqref{eq: intro} and tend to follow a two-step procedure.  First, they learn the mapping from input hazy image to its corresponding transmission map and  then using the estimated transmission map they calculate the clear image by reformulating Eq. \ref{eq: intro} as 
\begin{equation}
\mathbf{J(x)}=\frac{\mathbf{I(x)}-\mathbf{A(x)}(1-t(\mathbf{x}))}{t(\mathbf{x})}.
\label{eq: dehaze1}
\end{equation}


As a result, most of the previous methods consider the task of transmission map estimation and dehazing as two separate tasks, except the Li \emph{et al.} \cite{dehaze_iccv17}.  By doing so, they are unable to accurately capture the transformation between the transmission map and the dehazed image. Motivated by this observation, we relax the constant atmospheric light assumption \cite{dehaze_iccv15, dehaze_tif_2012_} and propose to jointly learn the transmission map and dehazed image from an input hazy image using a deep CNN-based network.  {Relaxed constant atmospheric light hypothesis within a certain adjustable limit} not only allows us to exploit the benefits of multi-task learning but it also enables us to regress on losses defined in the image space.  By enforcing the network to learn the transmission map, we still follow the popular image degradation model \eqref{eq: intro}. This joint learning enables the network to implicitly learn the atmospheric light and hence avoiding the need for manual calculation. On the other hand, previous learning-based CNN methods \cite{dehaze_2016_eccv,dehaze_2016_tip}  utilize Euclidean loss in generating the corresponding transmission map, which may result in blurry effect and hence poor quality dehazed images \cite{dehaze_vis_edge}.  To tackle this issue, we incorporate the gradient loss combined with the adversarial loss to generate better transmission map with sharper edges.   

\begin{figure*}[t]
\centering
		\includegraphics[width=0.99\textwidth]{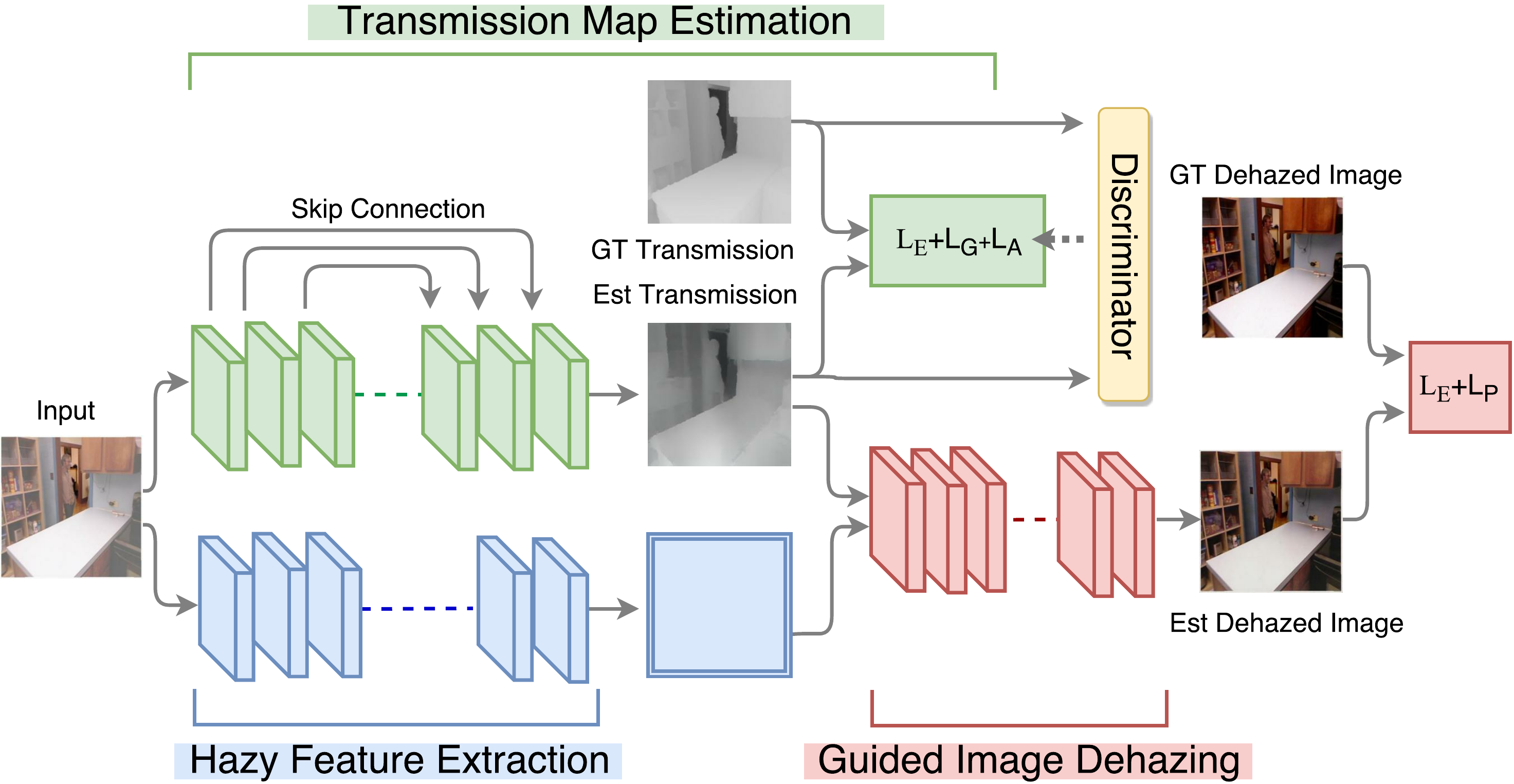}
	\vskip+4pt\caption{Overview of the proposed multi-task method for image dehazing. The proposed method consists of three modules: (a) Hazy feature extraction, (b) Transmission map estimation, and (c) Guided image dehazing. First, the transmission map is estimated from the input hazy image and it is concatenated with high dimensional feature map. These concatenated maps are fed into the guided dehazing module to estimate the dehazed image. The transmission map estimation module is trained using a GAN framework. The image dehazing module is trained by minimizing a combination of perceptual loss and Euclidean loss.}
\label{fig:overview}
\end{figure*}

Figure~\ref{fig:overview} gives an overview of the proposed single image dehazing method.  Our network consists of three parts: 1. Transmission map estimation, 2.  Hazy image feature extraction, and 3. Dehazing network guided by transmission map and hazy image features. The transmission map estimation is learned using a combination of adversarial loss, gradient loss and pixel-wise Euclidean loss. The transmission maps from this module are concatenated with the output of hazy image feature extraction module and processed by the dehazing network. Hence, the transmission maps are also involved in the dehazing procedure via the concatenation operator. The dehazing network is learned by optimizing a weighted combination of perceptual loss and pixel-wise Euclidean loss to generate perceptually better results.  Shown in Figure~\ref{fig:over} is a sample  dehazed image using the proposed method.  

This paper makes the following contributions:
\begin{itemize}[noitemsep]
  \item A novel joint transmission map estimation and image dehazing using deep networks is proposed. This is enabled by relaxing the constant atmospheric light assumption, thus allowing the network to implicitly learn the transformation from input hazy image to transmission map and transmission map to dehazed image. 
  \item We propose to use the recently introduced Generative Adversarial Network (GAN) framework for learning the transmission map.
  \item By performing a joint learning of transmission map and image dehazing, we are able to minimize losses defined in the image space such as perceptual loss and pixel-wise Euclidean loss, thereby generating perceptually better results with high quality details. 
  \item Extensive experiments on synthetic and real image datasets are conducted to demonstrate the effectiveness of the proposed method.
\end{itemize}

\section{Related Work}
We briefly review recent works on image dehazing and some commonly used losses in various CNN-based image reconstruction tasks. 

\subsection{Single Image Dehazing}
Early methods tend to address the dehazing problem via  including  certain  prior assumption. For example,  the authors in in \cite{dehaze_tan_2008} tend to recover  the contrast  for each patch relying  on the assumption that  that haze greatly decrease the contrast of the color images. Then, Kratz and Nishino \cite{dehaze_2009_kratz}  proposed to model the image with a factorial Markov random field in which the scene albedo and depth
are two statistically independent latent layers. He. \emph{et.al} in \cite{dehaz_2009_dark} proposed a dark-channel prior based on the surprising observation that RGB images from outdoor scene tend to have one channel  that in significantly dark.  Built on dark channel prior,   Meng \emph{et al.} \cite{dehaze_2013_meng}  imposing a specific boundary constraint during the estimation of transmission map.   More recently, Berman \emph{et al.} \cite{dehaze_2016_dana} proposed a non-local prior method based on the observation that the colors of a haze-free image can be well  represented  by a few hundred different colors that fall into several  tight clusters in the RGB space. 

The success of CNNs in modeling the non-learning mapping between input and output  has also  inspired researchers to explore CNN-based algorithms for low-level vision tasks such as image dehazing \cite{dehaze_2016_eccv,dehaze_2016_tip,dehaze_iccv17}. Unlike previous prior-based methods in the estimation of transmission map,  Cai \emph{et al.}  \cite{dehaze_2016_tip} train an end-to-end CNN network to  directly  estimate  the transmission map from the input haze image. More recently, Ren \emph{et al.} \cite{dehaze_2016_eccv} proposed a multi-scale deep architecture to directly regress the transmission maps via a course to fine fashion. However, the method of both Ren \emph{et al.} \cite{dehaze_2016_eccv} and Cai \emph{et al.}  \cite{dehaze_2016_tip} still leveraged a two-step procedure and hence the whole algorithm is not end-to-end optimized.  Most recently, Li \emph{et al} proposed an all-in-one dehazing network, where a linear embedding is leveraged to encode the transmission map and the atmospheric light into a single  variable.   Though these CNN-based learning methods achieve superior performance over the recent state-of-the-art methods, they limit their capabilities by learning a mapping only between the input hazy image and the transmission map. This is mainly due to the fact that these methods are based on the popular image degradation model given by \eqref{eq: intro} which assumes a constant atmospheric light.  In contrast,  we relax this assumption and thus enable the network to learn a transformation from the input hazy image to transmission map and transmission map to dehazed image. By doing this, we are also able to use losses defined in the image domain to learn the network. In the following sub-sections, two different losses that we use to improve the performance of the proposed network are reviewed.

\subsection{Loss Functions}
Loss functions form an important and integral part of a learning process, especially in CNN-based reconstruction tasks. Initial work on CNN-based image regression tasks optimized over pixel-wise L2-norm (Euclidean loss) or L1-norm between the predicted and ground truth images \cite{long2015fully,eigen2015predicting, fu2017clearing}. Since these losses operate at per-pixel level, their ability to capture high level perceptual/contextual details such as sharp edges and complicated contour are limited and they tend to produce blurred results. In order to overcome this issue, we use two different loss functions: adversarial loss and perceptual loss for learning the transmission map and dehazed image, respectively. 

\subsubsection{Adversarial loss} The adversarial loss,  formulated in the Generative Adversarial Networks(GAN) work by Goodfellow \textit{et al.} \cite{GAN}, has been widely used in generating realistic images. GAN consists of a generator and a discriminator that are jointly optimized. While the generator's goal is to synthesize images that are similar in distribution of the training images, the discriminator's job is to identify if the images fed to it are real or synthesized (fake). After the success of this method in generating realistic images, this concept has been explored as different formulations in various applications such as  data augmentation \cite{peng2018jointly}, paired and unpaired 2d/3d image to image translation \cite{zhang2018translating,pix2pix2016, zhang2018synthesis,WEIKAI_GAN, yamaguchi2018high }, image super-resolution \cite{GAN_sisr}, image inpainting \cite{yu2018generative,yu2018free,zhao2018identity} and image de-raining \cite{derain_2017_zhang}. In our work, we propose to use the GAN framework as an additional loss function to guide the learning of transmission map, which when optimized appropriately, will generated realistic transmission maps.

\subsubsection{Perceptual loss} Many researchers have argued and demonstrated through their results that it would be better to optimize a perceptual loss function in various applications \cite{nips-generating,perceptual_loss,luan2017deep,yulun_per}. The perceptual function is usually defined using high-level
features extracted from a pre-trained convolutional network. The aim is to minimize the perceptual difference between the reconstructed image and the ground truth image. Perceptually superior results were obtained for both super-resolution and artistic style-transfer \cite{li2016precomputed,gatys2015neural,zhang2018multi, xiong_inpainting}. In this work, a VGG-16 architecture \cite{vgg} based perceptual loss is used to train the network for performing dehazing.



\section{Proposed Method}


The proposed network is illustrated in Figure~\ref{fig:overview} which consists of the following modules: 1. \textit{Transmission map estimation}, 2. \textit{Hazy image feature extraction}, and 3. \textit{Transmission guided image dehazing},
where the first module learns to estimate transmission maps from corresponding input hazy images, the second module extracts haze relevant features from the input hazy image and the third module learns to perform image dehazing by combining the feature information extracted from the hazy image with the estimation transmission map.  In what follows, we explain these modules in detail.

\subsection{Transmission Map Estimation}
\label{ssec:tme}
The task of predicting transmission map from a given input hazy image is considered as a pixel-level image regression task. In other words, the aim is to learn a pixel-wise non-linear mapping from a given input image to the corresponding transmission map by minimizing the loss between them. In contrast to the method used by Ren \emph{et al.} in \cite{dehaze_2016_eccv}, our method uses adversarial loss in addition to pixel-wise Euclidean loss to learn better quality transmission maps. Also, the network architecture used in this work is very different from the one used in \cite{dehaze_2016_eccv}.

For incorporating the adversarial loss, the transmission map estimation is learned in the Conditional Generative Adversarial Network (CGAN) framework \cite{mirza2014conditional}. Similar to earlier works on GANs for image reconstruction tasks \cite{derain_2017_zhang,GAN_pix2pix2016,GAN_sisr}, the proposed network for learning the transmission map consists of two sub-networks: Generator $G$ and Discriminator $D$. The goal of GAN is to train $G$ to produce samples from training distribution such that the synthesized samples are indistinguishable from the actual distribution by the discriminator $D$. The sub-network $G$ is motivated by the success of encoder-decoder structure in pixel-wise image reconstruction \cite{unet,image_res_cnn,GAN_pix2pix2016}. In this work, we  adopt a `U-Net'-based structure \cite{unet} as the generator for the transmission map estimation. Rather than concatenating the symmetric layers during training, shortcut connections \cite{deep_residue} are used to connect the symmetric layers  with the aim of addressing the vanishing gradient problem for deep networks. To better capture the semantic information and make the generated transmission map indistinguishable from the ground truth transmission map,  a CNN-based differentiable discriminator is used as a  `guidance' to guide the generator in generating better transmission maps. The proposed generator network is as follows (the shortcut connection is neglected here): 

\noindent  \emph{CP(15)-CBP(30)-CBP(60)-CBP(120)-CBP(120)-CBP(120)-CBP(120)-CBP(120)-TCBR(120)-TCBR(120)-TCBR(120)-TCBR(120)-TCBR(60)-TCBR(30)-TCBR(15)-TC(1)-TanH},
 
\noindent where $C$ represents the convolutional layer, $TC$ represents transpose convolution layer, $P$ indicates Prelu \cite{prelu} and $B$ indicates batch-normalization \cite{batch_norm}. The number in the bracket represents the number of output feature maps of the corresponding layer.

To ensure that the estimated transmission map is indistinguishable from the ground truth image, a learned discriminator sub-network is designed to classify if each input image is real or fake. Inspired by the success of patch-discriminator in distinguish real from fake, we also adopt a 70$\times$ 70 patch discriminator, where $70\times 70$ indicates the receptive field of the discriminator, to generate visually pleasing and sharper results. \cite{yan2018fast} also explores other ways to make the images sharper.   The structure of the discriminator is defined as follows: 

\noindent \emph{CB(48)-CBP(96)-CBP(192)-CBP(384)-CBP(384)-C(1)-Sigmoid}.

Furthermore, we propose to employ gradient-based loss function in order to enforce consistency in the gradients between the estimated and target transmission map. The use of gradient loss function is inspired by its success in several other tasks such as depth estimation  \cite{depth_gra_zhou,depth_gra}.

\normalsize
\subsection{Hazy Feature Extraction and Guided Image Dehazing}
A possible solution to image dehazing is to directly learn an end-to-end non-linear mapping between the estimated transmission map and the desired dehazed output. However, as shown in  \cite{GAN_pix2pix2016}, while learning a mapping from transmission map-like to an RGB color image is possible, one may loose some information due to the absence of the albedo and the lighting information.  

To generate better dehazed image and enable the whole process (estimation of the transmission map and the dehazed image) end-to-end, we propose a deep transmission guided network for single image dehazing via relaxing the assumption of constant atmospheric light.  Inspired by \emph{guided filtering} \cite{deep_joint_guided,guided_1,guided_2}, where a guidance image is leveraged  to guided the generation of high-quality results (eg. depth map), a set of convolutional layers with symmetric skip connections are stacked in the front and they serve as a hazy image feature extractor. Basically, the hazy feature extraction part extract deep features from the input hazy image. Then,   These feature maps are concatenated with the estimated transmission map. Then the concatenation is fed into the guided image dehazing module. This module consists of another set of CNN layers with non-linearities and it essentially acts as a fusion CNN whose task is to learn a mapping from transmission map and high-dimensional feature maps to dehazed image. \footnote{Note that our network is quite different from the network proposed in \cite{deep_joint_guided} in the sense that the proposed network is a multi-task learning network with a single input while the network in \cite{deep_joint_guided} is a single-task network with two inputs.}   To learn this network, a perceptual loss function based on VGG-16 architecture \cite{vgg} is used in addition to pixel-wise Euclidean loss. The use of perceptual loss greatly enhances the visual appeal of the results. Details of the network structure for the hazy feature extraction and guided image dehazing module are as follows:\\

\noindent \emph{CP(20)-CBP(40)-CBP(80)-C(1)-Conca(2)-CP(80)-CBP(40)-CBP(20)-C(3)-TanH},

\noindent where $Conca$ indicates concatenation.

In summary, a non-linear mapping from the input hazy image and transmission map to dehazed image is learned in a multi-task end-to-end fashion. By learning this mapping, we enforce our network to implicitly learn the estimation of atmospheric light, thereby avoiding the ``manual" estimation as followed by some of the existing methods. 
\begin{figure*}[t]
	\centering
	\begin{minipage}{.185\textwidth}
		\centering
		\caption*{\quad}
		\includegraphics[width=1\textwidth]{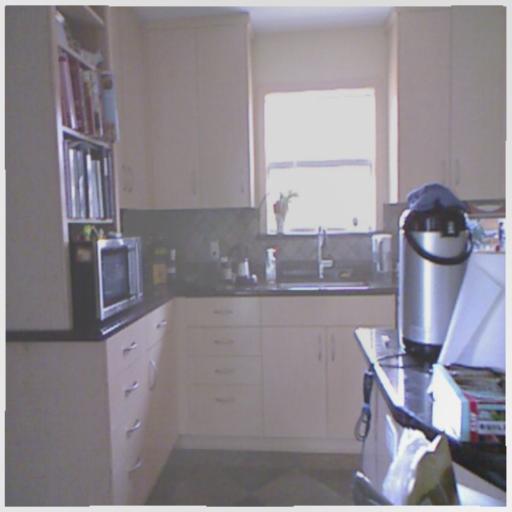}
		\captionsetup{labelformat=empty}
		\captionsetup{justification=centering}
		\caption*{Input}
	\end{minipage} 
	\begin{minipage}{.185\textwidth}
		\centering
		\caption*{SSIM:0.8584}
		\includegraphics[width=1\textwidth]{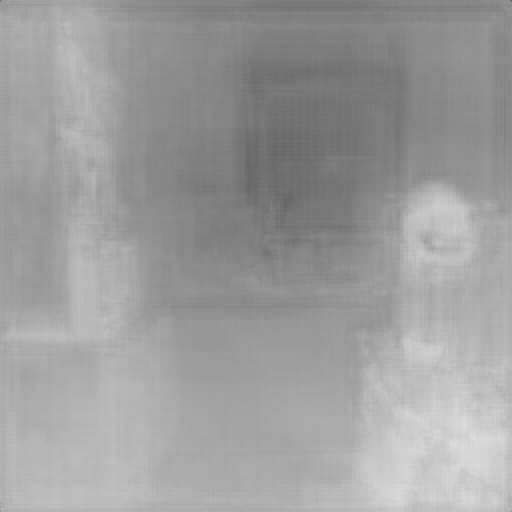}
		\captionsetup{labelformat=empty}
		\captionsetup{justification=centering}
		\caption*{\textbf{\emph{T-L2}}}
	\end{minipage} 
	\begin{minipage}{.185\textwidth}
		\centering
		\caption*{SSIM:0.8654}
		\includegraphics[width=1\textwidth]{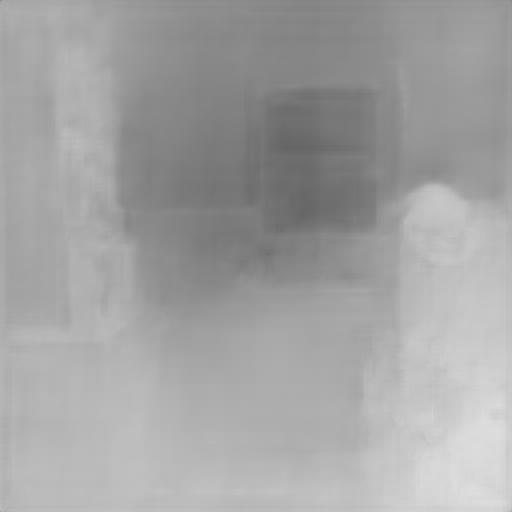}
		\captionsetup{labelformat=empty}
		\captionsetup{justification=centering}
		\caption*{\textbf{\emph{T-L2-G}}}
	\end{minipage} 
	\begin{minipage}{.185\textwidth}
		\centering
		\caption*{SSIM:0.8854}
		\includegraphics[width=1\textwidth]{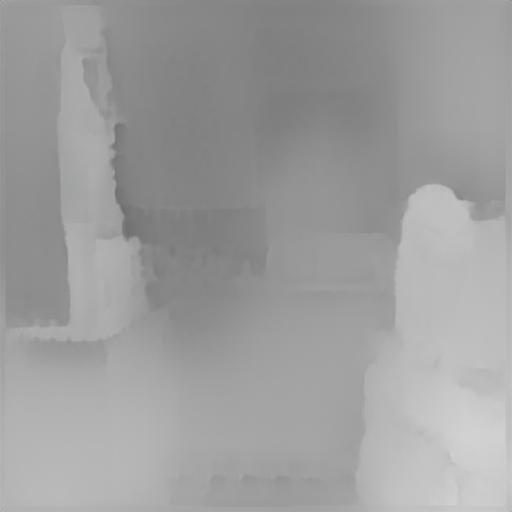}
		\captionsetup{labelformat=empty}
		\captionsetup{justification=centering}
		\caption*{\textbf{\emph{T-L2-G-GAN}}}
	\end{minipage} 
	\begin{minipage}{.185\textwidth}
		\centering
		\caption*{SSIM:1}
		\includegraphics[width=1\textwidth]{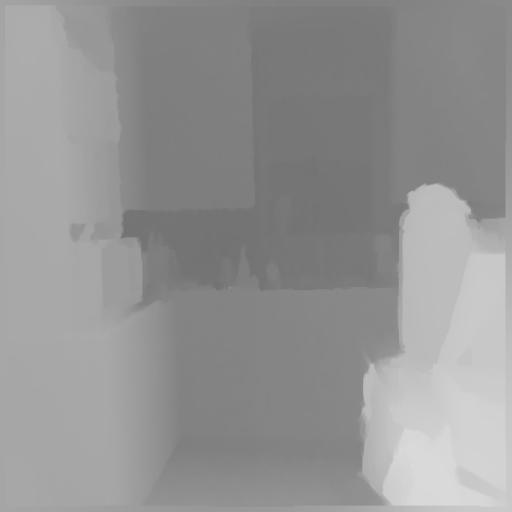}
		\captionsetup{labelformat=empty}
		\captionsetup{justification=centering}
		\caption*{Target}
	\end{minipage}
	\vskip -4pt\caption{Transmission estimation results for \textbf{Ablation 1}. It can be observed that gradient loss enable sharper edges and final GAN framework help to preserved better structure information for each object. }
	\label{fig: base_tran2}
\end{figure*}

\subsection{Training Loss} 
As discussed earlier, the proposed method involves joint learning of two tasks: transmission map estimation and dehazing. Accordingly, to train the network, we define two losses $L^t$ and $L^d$, respectively for the two tasks.\\ 


\subsubsection{Transmission map loss $L^t$}
 To overcome the issue of blurred results due to the minimization of $L_2$ error, the transmission map estimation network is learned by minimizing a weighted combination of $L_2$ error, an adversarial error and a gradient loss. The transmission map loss is defined as
\begin{equation}
L^t = L^t_E + \lambda_a L^t_A + \lambda_G L^t_G,
\label{eq:trans_loss}
\end{equation}
where $\lambda_a$ and $\lambda_G$ are two weights, $L^t_E$ is the pixel-wise Euclidean loss, $L^t_A$ is the adversarial loss and $L^t_G$ is the two-directional gradient loss. These three  losses are defined as follows
\begin{equation}
\label{eq:trans_loss_euc}
 L^t_E= \sum_{w,h} \|(\phi_{G}({\mathbf{I}}))^{w,h}-\mathbf{y}_t^{w,h}\|_2,	\\
 \end{equation} 
 \begin{equation}
 \label{eq:trans_loss_adv}
   L_A^t= -\log(\phi_D(\phi_G({\mathbf{I}})),
 \end{equation}
 where $\mathbf{I}$ is a $C$-channel input hazy image, $\mathbf{y}_t$ is the ground truth transmission map, $w\times h$ indicates the dimension of the input image and transmission map, $\phi_{G}$ is the generator sub-network $G$ for generating the transmission map and $\phi_D$ is the discriminator sub-network $D$.  The directional gradient loss, which has been discussed in other applications \cite{yan2018supervised,yan2018effective},  the  is defined as:
 \begin{equation}
 \label{eq:gra}
 \begin{split}
 L^t_G=&\sum_{w,h}^{}\|(H_x({G_t({\mathbf{I}}})))^{w,h}-(H_x(t))^{w,h}\|_2\\
 &+\|(H_y({G_t({\mathbf{I}}})))^{w,h}-(H_y(t))^{w,h}\|_2,
 \end{split}
 \end{equation}
where $H_x$ and $H_y$ are operators that compute image gradients along rows (horizontal) and columns (vertical), respectively and  $w\times h$ indicates  the width and height  of the output feature map.

{Traditional techniques for transmission map estimation employ only the Euclidean loss ($L_E^t$) to learn the network weighs. However, Euclidean loss is known to introduce blur in the generated output. Hence, the use of additional loss functions (adversarial loss and gradient loss) incorporates further constraints into the learning framework. Specifically, the adversarial loss ($L_A^t$) enforces the network to generate transmission maps that are closer to the input distribution and the gradient loss ($L_G^t$) ensures consistency between the gradients of the target and estimated transmission map. The weights $\lambda_A,\lambda_G$ are set using validation.} 

\subsubsection{Dehazing loss $L^d$} The dehazing network is learned by minimizing  a weighted combination of the pixel-wise Euclidean loss and perceptual loss between the ground-truth dehazed image and the network output and is defined as follows
\begin{equation}
L^d = L^d_E + \lambda_p L^t_P,
\label{eq:dehaze_loss}
\end{equation}
where $\lambda_p$ is a weighting factor, $L^d_E$ is the pixel-wise Euclidean loss and $L^t_P$ is the perceptual loss and are respectively defined as 
 \begin{equation}
 \label{eq:dehaze_loss_euc}
 L^d_E= \sum_{c,w,h}^{} \|\phi_E({\mathbf{I}})^{c,w,h}-\mathbf{J}^{c,w,h}\|_2,
 \end{equation}
 
\begin{equation}
\label{eq:dehaze_loss_perc}
 L^d_P= \sum_{c_i,w_i,h_i}^{} \|V(\phi_E({\mathbf{I}}))^{c_i,w_i,h_i}-V
 (\mathbf{J})^{c_i,w_i,h_i}\|_2,
\end{equation}
 where $\mathbf{I}$ is a $c$-channel input hazy image, $\mathbf{J}$ is the ground truth dehzed image, $w\times h$ is the dimension of the input image and the dehazed image, $\phi_{E}$ is the proposed network, $V$ represents a non-linear CNN transformation and $C_i,W_i,H_i$ are the dimensions of a certain high level layer of $V$. Similar to the idea proposed in \cite{perceptual_loss},  we aim to minimize the distance between high-level features along with pixel-wise Euclidean loss. In our method, we compute the feature loss at layer relu3$\_$1 in VGG-16 model \cite{vgg}.\footnote{https://github.com/ruimashita/caffe-train/blob/master/\\vgg.train$\_$val.prototxt}   Note that the dehazing loss $L^d$ is also to be propagated to the transmission estimation part.

\subsection{Discussion}
{Relaxing the condition of constant atmospheric light enables the network to be trained in an end-to-end fashion, thus allowing the network to implicitly learn the transformation from input hazy image to transmission map and transmission map to dehazed image. While it allows more flexibility in the learning process,  it introduces more complexity on the model. Hence, to efficiently learn the network parameters, the transmission map is  considered since it preserve information about the portion of the light that is not scattered that reaches the camera. Furthermore,  additional losses such as adversarial loss and gradient loss function introduce strong regularization, thus enabling better estimation of transmission map. }
\section{Experiments}
In this section, we present the details and results of various experiments conducted on synthetic and real datasets that contain a variety of hazy conditions. First we describe the datasets used in our experiments. Then, we  discuss the details of the training procedure.  Next, we discuss the results of the ablation study conducted to understand the improvements obtained by various modules of the proposed method. Finally, we compare the results of the proposed network with recent state-of-the-art methods. Through these experiments, we attempt to demonstrate the superiority of the proposed method and the effectiveness of its' various components.

\begin{figure*}[t]
	\centering
	\begin{minipage}{.185\textwidth}
		\centering
		\caption*{SSIM:0.4551}\vskip-6pt
		\includegraphics[width=4cm,height=3.0cm]{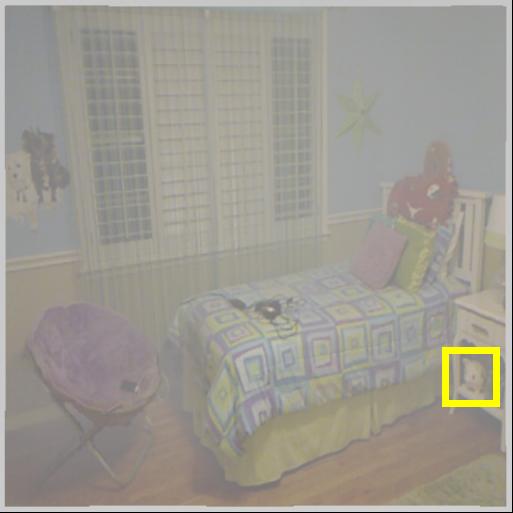}
		\captionsetup{labelformat=empty}
		\captionsetup{justification=centering}
	\end{minipage} 
	\begin{minipage}{.185\textwidth}
		\centering
		\caption*{SSIM:0.8241}\vskip-6pt
		\includegraphics[width=4cm,height=3.0cm]{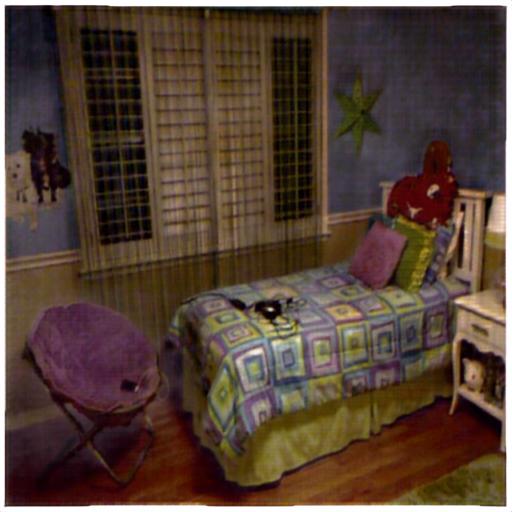}
		\captionsetup{labelformat=empty}
		\captionsetup{justification=centering}
	\end{minipage} 
	\begin{minipage}{.185\textwidth}
		\centering
		\caption*{SSIM:0.8452}\vskip-6pt
		\includegraphics[width=4cm,height=3.0cm]{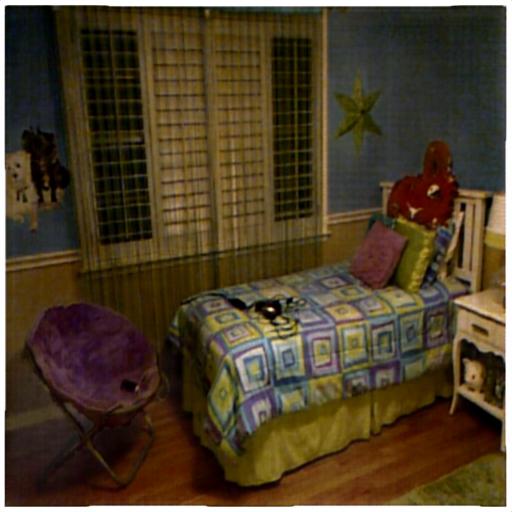}
		\captionsetup{labelformat=empty}
		\captionsetup{justification=centering}
	\end{minipage} 
	\begin{minipage}{.185\textwidth}
		\centering
		\caption*{SSIM:0.8838}		\vskip-6pt

		\includegraphics[width=4cm,height=3.0cm]{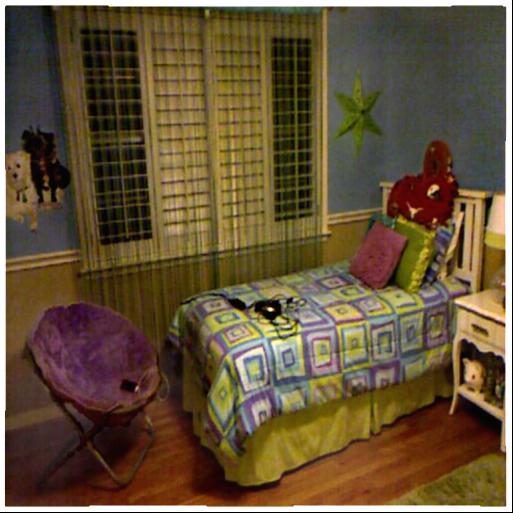}
		\captionsetup{labelformat=empty}
		\captionsetup{justification=centering}
	\end{minipage} 
	\begin{minipage}{.185\textwidth}
		\centering
		\caption*{SSIM:1}\vskip-6pt
		\includegraphics[width=4cm,height=3.0cm]{baseline//11_target.jpg}
		\captionsetup{labelformat=empty}
		\captionsetup{justification=centering}
	\end{minipage}\\
	\begin{minipage}{.185\textwidth}
		\centering
		\includegraphics[width=4cm,height=3.0cm]{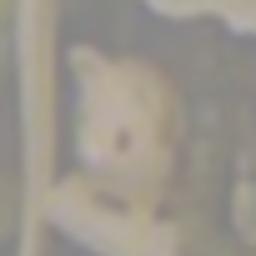}
		\captionsetup{labelformat=empty}
		\captionsetup{justification=centering}
		\caption*{Input}
	\end{minipage} 
	\begin{minipage}{.185\textwidth}
		\centering
		\includegraphics[width=4cm,height=3.0cm]{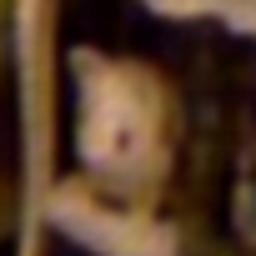}
		\captionsetup{labelformat=empty}
		\captionsetup{justification=centering}
		\caption*{\textbf{\emph{I-L2-noT}}}
	\end{minipage} 
	\begin{minipage}{.185\textwidth}
		\centering
		\includegraphics[width=4cm,height=3.0cm]{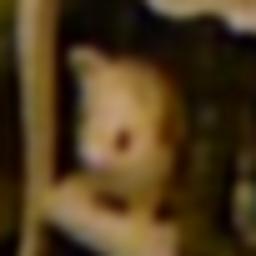}
		\captionsetup{labelformat=empty}
		\captionsetup{justification=centering}
		\caption*{\textbf{\emph{I-L2-T}}}
	\end{minipage} 
	\begin{minipage}{.185\textwidth}
		\centering
		\includegraphics[width=4cm,height=3.0cm]{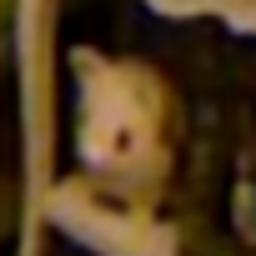}
		\captionsetup{labelformat=empty}
		\captionsetup{justification=centering}
		\caption*{\textbf{\emph{I-L2-Per-T}}}
	\end{minipage} 
	\begin{minipage}{.185\textwidth}
		\centering
		\includegraphics[width=4cm,height=3.0cm]{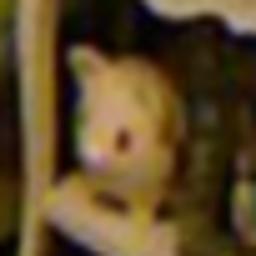}
		\captionsetup{labelformat=empty}
		\captionsetup{justification=centering}
		\caption*{Target}
	\end{minipage}
	\vskip 4pt\caption{Dehazed image results and certain zoomed-in parts for \textbf{Ablation 2}. It can be observed that the introduction of transmission map reduce the color  distortion and the involvement of perceptual loss enable high quality dehazed result.  }
	\label{fig: base_tran}
\end{figure*}

\subsection{Datasets}
Since it is extremely difficult to collect a dataset that contains a large number of hazy/clear/transmission-map image pairs, training and test datasets are synthesized using \eqref{eq: intro} and following the idea proposed in \cite{dehaze_2014_tang,dehaze_2016_tip,dehaze_2016_eccv}.  All the training and test samples are obtained from the NYU Depth dataset \cite{nyudepth}. More specifically, given a haze-free image, we randomly sample four atmosphere light $\mathbf{A(x)} \in[0.5,1.2]$ and the scattering coefficient of the atmosphere $\beta \in[0.4,1.6]$ to generate its corresponding hazy images and transmission maps. An initial set of 600 images are randomly chosen from the NYU dataset. From each image belonging to this initial set, 4 training images are generated by using randomly sampled atmospheric light and scattering coefficient, obtaining a total of 2400 training images. In a similar way, a test dataset consisting of 300 images is obtained. We ensure that none of the training images are in the test set.  By varying $\mathbf{A}$ and $\beta$, we generate our training data with a variety of different conditions.  

As discussed in \cite{dehaze_2016_eccv,dehaze_2014_tang}, the image content is independent of its corresponding depth. Even though the training images are  from the indoor dataset \cite{nyudepth} and hence depths of all the images are relatively shallow, we could modify the value of the attenuation coefficient $\beta$ to vary the haze concentration to make sure the datasets can also used for outdoor image dehazing. Meanwhile, the experimental results have also demonstrated the effectiveness of discussed training datasets.  

To  demonstrate the effectiveness of the proposed method on real-world data, we also created a test dataset including 30 hazy images downloaded from the Internet.

\subsection{Training Details}
The entire network is trained on a Nvidia Titan-X  GPU using the torch framework \cite{torch}. We choose $\lambda_a=0.003$ and $\lambda_G=1$ for the loss in estimating the transmission map and $\lambda_p=1.5$  for the loss in single image dehazing. During training, we use ADAM \cite{adam_opt} as the optimization algorithm with learning rate  of $2\times 10^{-3}$ and batch size of  10 images.   All the training samples are resized to $256\times 256$.  To efficiently train the multi-task network, we leverage the stage-wise training strategy. First, the transmission map estimation module is trained using the loss in Eq.~\ref{eq:trans_loss} . Then, the entire network is fine-tuned using both Eq.~\ref{eq:dehaze_loss} and Eq.~\ref{eq:trans_loss}.
\subsection{Ablation Study}
In order to demonstrate the improvements obtained by different modules for both transmission maps and dehazed images, we conduct two ablation studies for estimating transmission maps and dehazed images, separately. \\

\noindent {\bf{Ablation 1}:} This ablation study demonstrates the effectiveness of different modules in the transmission map estimation block and it consists of the following experiments:\\ 1) \textit{Transmission map estimation using only L2 loss (\textbf{T-L2})},\\
  2) \textit{Transmission map estimation using  L2 loss and gradient loss (\textbf{T-L2-G})},  and\\
    3) \textit{Transmission map estimation using L2 loss, gradient loss and  adversarial loss (\textbf{T-L2-G-GAN})}. 
\begin{table}[]
	\centering
	\caption{Quantitative SSIM results for \textbf{Ablation 1} evaluated on synthetic  datasets for transmission map. }
    \resizebox{0.45\textwidth}{!}{%
\begin{tabular}{cccccc}
\hline
\multicolumn{1}{|c|}{} & \multicolumn{1}{c|}{\emph{Input}} & \multicolumn{1}{c|}{\emph{\textbf{T-L2}}} & \multicolumn{1}{c|}{\emph{\textbf{T-L2-G}}} & \multicolumn{1}{c|}{\emph{\textbf{T-L2-ALL}}} & \multicolumn{1}{c|}{Target} \\ \hline
\multicolumn{1}{|c|}{Transmission Map} & \multicolumn{1}{c|}{0.4523} & \multicolumn{1}{c|}{0.9052} & \multicolumn{1}{c|}{0.9257} & \multicolumn{1}{c|}{\textbf{0.9388}} & \multicolumn{1}{c|}{1.0000} \\ \hline
\end{tabular}}
	\label{ta:base_quan1}
\end{table}

Sample results are shown in Fig~\ref{fig: base_tran2}. It can be observed that the introduction of gradient loss (\emph{\textbf{T-L2-G}}) eliminates halo-artifacts near complicated edges \cite{dehaze_vis_edge}.  Furthermore, the introduction of the discriminator (GAN framework-\emph{\textbf{T-L2-G-GAN}}) effectively refine the local regions and enables sharper reconstructions, thereby preserving the structure for each object. Results of quantitative analysis on synthetic datasets are presented in Table \ref{ta:base_quan1}. The effect of different modules in the proposed network can be clearly observed from this table.\\

\noindent {\bf{Ablation 2:}}  Similarly, another ablation study is conducted to demonstrate the improvements obtained by different modules for dehazing images. This ablation study involves the following experiments:\\
 1) \textit{Image dehazing using L2 loss without estimation of transmission map (\textbf{I-L2-noT})},\\
   2) \textit{Image dehazing using L2 loss with estimation of transmission map (\textbf{I-L2-T})}, and\\
      3) \textit{Image dehazing using L2 loss and perceptual loss with estimation of transmission map (\textbf{I-L2-Per-T})}.

Sample results are shown in Fig~\ref{fig: base_tran}. It can be observed that the method (\emph{\textbf{I-L2-noT}})  is unable to accurately estimate the  haze level and depth (both are inherently captured in the transmission map) and hence the dehazed results tend to contain some color distortion. The introduction of the branch for the estimation of transmission map helps to generate better quality images.  This can be seen by comparing the second column and the third column in Fig~\ref{fig: base_tran}.  Furthermore,  the final involvement of the perceptual loss \emph{\textbf{I-L2-Per-T}} is able to generate better dehazed images with high quality details (observed from the zoom-in parts in Fig~\ref{fig: base_tran}).  {We also compare the inference running time for  each ablation study, as tabulated in Table \ref{ta:ab2_runningtime}. It can be observed that the multi-task learning results in slight increase in complexity of training and inference time. However, it leads to substantial improvements in the dehazing quality. The introduction of different loss functions such as gradient loss and perceptual loss increase the training time, however, it does not affect the inference time. }
\begin{table}[]
	\centering
	\caption{Quantitative SSIM results for \textbf{Ablation 2} evaluated on synthetic  datasets for dehazed image. }
    \resizebox{0.45\textwidth}{!}{%
\begin{tabular}{cccccc}
\hline
\multicolumn{1}{|c|}{} & \multicolumn{1}{c|}{\emph{Input}} & \multicolumn{1}{c|}{\emph{I-L2-noT}} & \multicolumn{1}{c|}{\emph{I-L2-T}} & \multicolumn{1}{c|}{\emph{I-L2-Per-T}} & \multicolumn{1}{c|}{Target} \\ \hline
\multicolumn{1}{|c|}{Dehazed Image} & \multicolumn{1}{c|}{0.7041} & \multicolumn{1}{c|}{0.8835} & \multicolumn{1}{c|}{0.9002} & \multicolumn{1}{c|}{\textbf{0.9133}} & \multicolumn{1}{c|}{1.0000} \\ \hline
\end{tabular}}
	\label{ta:base_quan2}
\end{table}

\begin{table}[]
	\centering
	\caption{Average running time for \textbf{Ablation 2} evaluated on synthetic  datasets for dehazed image. }
	\resizebox{0.45\textwidth}{!}{%
		\begin{tabular}{ccccc}
			\hline
			\multicolumn{1}{|c|}{} & \multicolumn{1}{c|}{\emph{Input}} & \multicolumn{1}{c|}{\emph{I-L2-noT}} & \multicolumn{1}{c|}{\emph{I-L2-T}} & \multicolumn{1}{c|}{\emph{I-L2-Per-T}}  \\ \hline
			\multicolumn{1}{|c|}{Time (s)} & \multicolumn{1}{c|}{2.65} & \multicolumn{1}{c|}{3.33} & \multicolumn{1}{c|}{3.33} & \multicolumn{1}{c|}{3.33}  \\ \hline
	\end{tabular}}
	\label{ta:ab2_runningtime}
\end{table}

\begin{table*}[t]
\vskip -4pt
	\centering
    \caption{Quantitative SSIM results on the synthetic  dataset. }
    \resizebox{0.95\textwidth}{!}{%
\begin{tabular}{|c|c|c|c|c|c|c|c|}
	\hline
	& Input &  \begin{tabular}[c]{@{}l@{}}He. \emph{et al.} \cite{dehaz_2009_dark}\\ \end{tabular} & \begin{tabular}[c]{@{}l@{}}Zhu. \emph{et al.} \cite{dehaze_fast15}\\  \end{tabular}   & \begin{tabular}[c]{@{}l@{}} Ren. \emph{et al.} \cite{dehaze_2016_eccv} \\  \end{tabular} & \begin{tabular}[c]{@{}l@{}}Berman. \emph{et al.} \cite{dehaze_2016_dana,dehaze_iccp217} \\ \end{tabular} & \begin{tabular}[c]{@{}l@{}}Li. \emph{et al.} \cite{dehaze_iccv17} \\  \end{tabular}  & Our\\ \hline\hline
	Transmission & N/A & 0.8739 & 0.8326  & N/A & 0.8675 & N/A & \textbf{0.9388} \\ \hline
	Image &  0.7041  & 0.8642   & 0.8567 & 0.8203 & 0.7959 & {0.8842} & \textbf{0.9133} \\ \hline
\end{tabular}}
	\label{ta:quantitive_depth1}
\vskip -4pt
\end{table*}

\begin{figure*}[!]
	\centering
	\begin{minipage}{.118\textwidth}
		\centering
	 	\caption*{\emph{SSIM: N/A}}
	 	\vskip -6pt
		\includegraphics[width=1\textwidth]{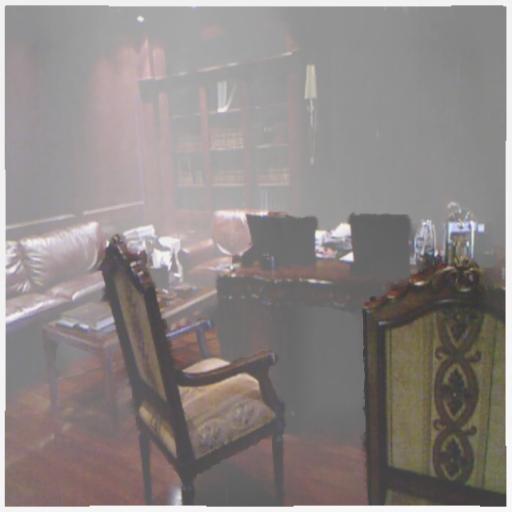}
		\captionsetup{labelformat=empty}
		\captionsetup{justification=centering}
	\end{minipage} 
	\begin{minipage}{.118\textwidth}
		\centering
	 	\caption*{\emph{SSIM: 0.9422}}
	 	\vskip -6pt
		\includegraphics[width=1\textwidth]{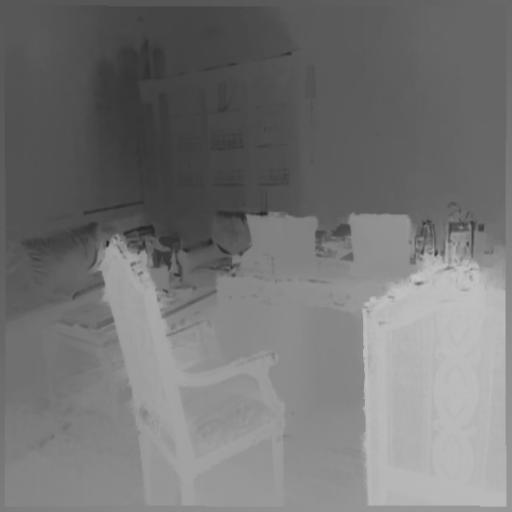}
		\captionsetup{labelformat=empty}
		\captionsetup{justification=centering}
	\end{minipage} 
	\begin{minipage}{.118\textwidth}
		\centering
	 	\caption*{\emph{SSIM: 0.8633}}
	 	\vskip -6pt
		\includegraphics[width=1\textwidth]{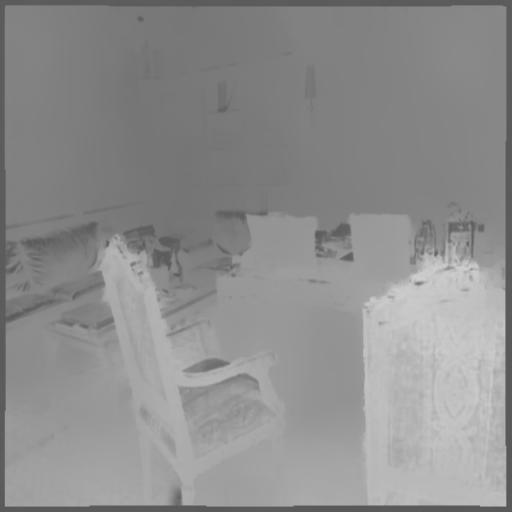}
		\captionsetup{labelformat=empty}
		\captionsetup{justification=centering}
	\end{minipage} 
	\begin{minipage}{.118\textwidth}
		\centering
	 	\caption*{\emph{SSIM: N/A}}
	 	\vskip -6pt
		\includegraphics[width=1\textwidth]{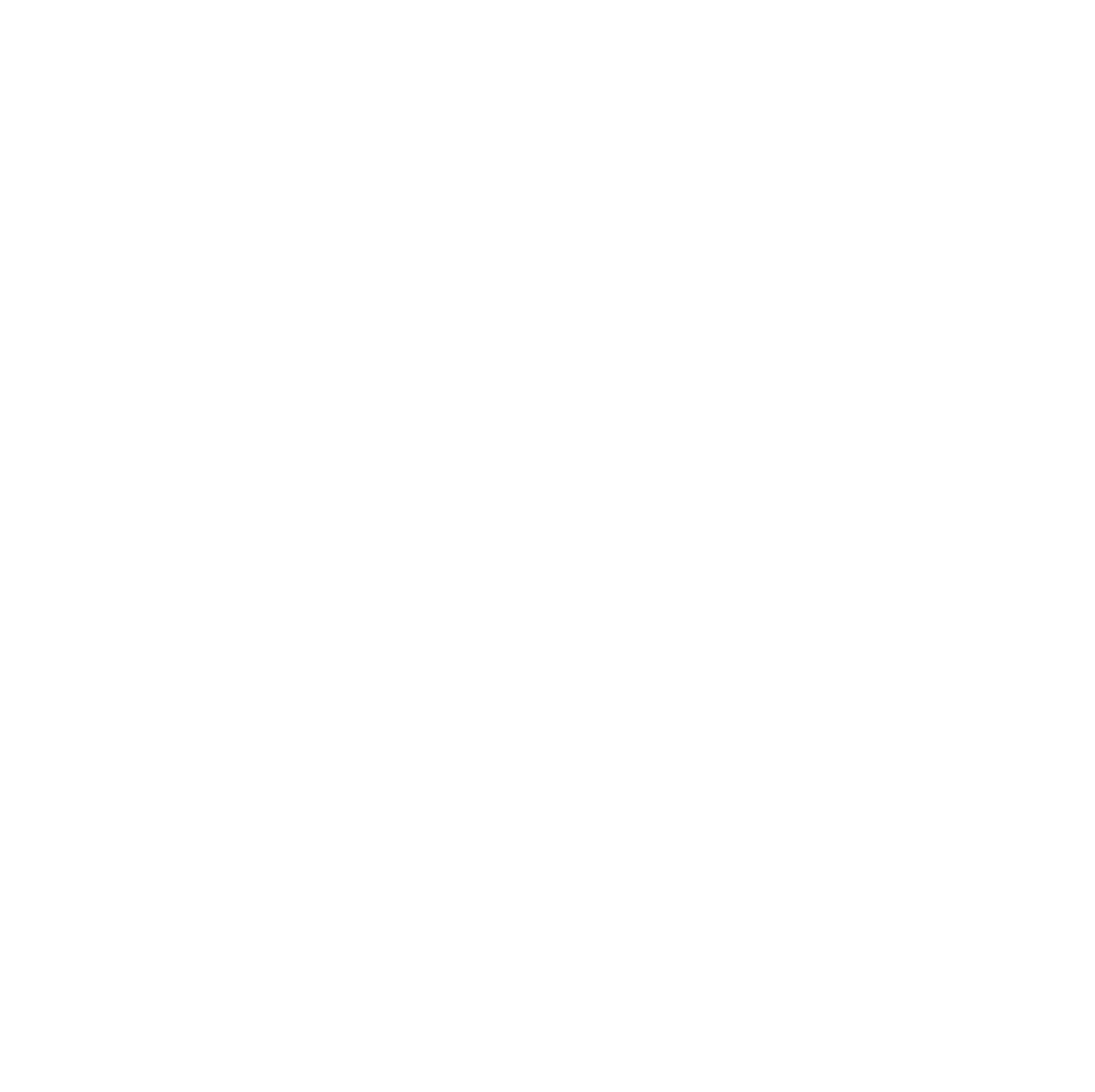}
		\captionsetup{labelformat=empty}
		\captionsetup{justification=centering}
	\end{minipage} 
	\begin{minipage}{.118\textwidth}
		\centering
	 	\caption*{\emph{SSIM: 0.9307}}
	 	\vskip -6pt
		\includegraphics[width=1\textwidth]{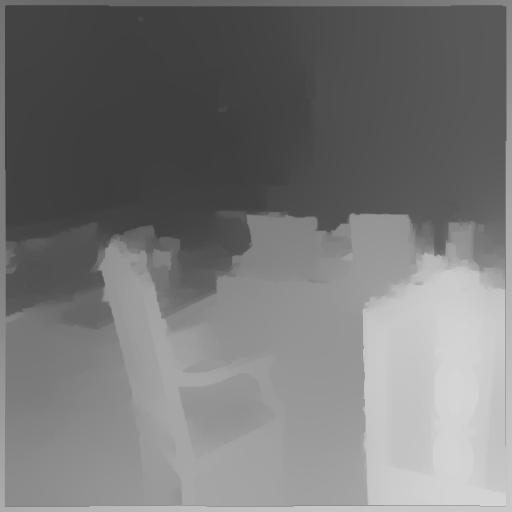}
		\captionsetup{labelformat=empty}
		\captionsetup{justification=centering}
	\end{minipage} 
	\begin{minipage}{.118\textwidth}
		\centering
	 	\caption*{\emph{SSIM: N/A}}
	 	\vskip -6pt
		\includegraphics[width=1\textwidth]{syn_result_image//white.pdf}
		\captionsetup{labelformat=empty}
		\captionsetup{justification=centering}
	\end{minipage} 
	\begin{minipage}{.118\textwidth}
		\centering
	 	\caption*{\emph{\textbf{SSIM: 0.9733}}}
	 	\vskip -6pt
		\includegraphics[width=1\textwidth]{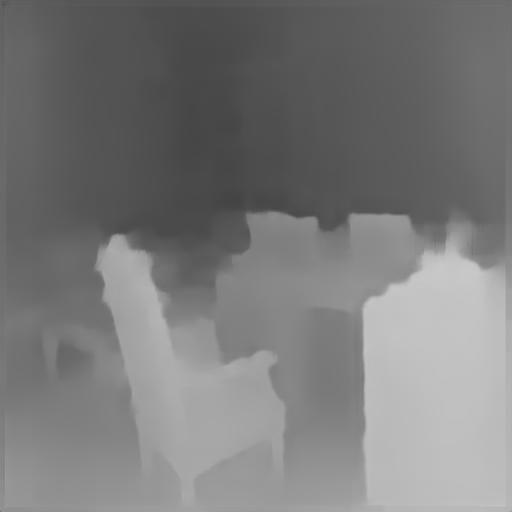}
		\captionsetup{labelformat=empty}
		\captionsetup{justification=centering}
	\end{minipage}
	\begin{minipage}{.118\textwidth}
		\centering
	 	\caption*{\emph{SSIM: 1}}
	 	\vskip -6pt
		\includegraphics[width=1\textwidth]{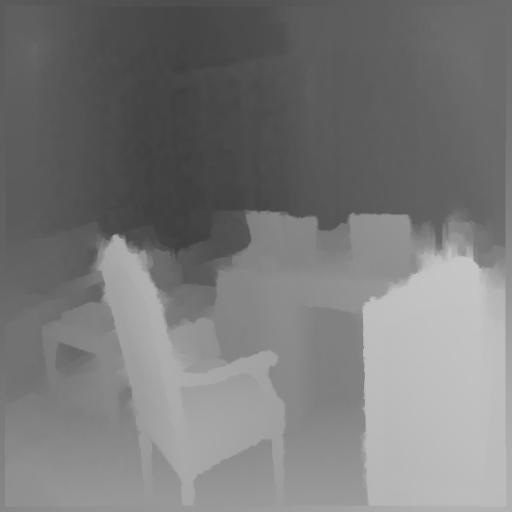}
		\captionsetup{labelformat=empty}
		\captionsetup{justification=centering}
	\end{minipage}\\
	\vskip+5pt
	\begin{minipage}{.118\textwidth}
		\centering
	 	\caption*{\emph{SSIM: 0.5788}}
	 	\vskip -6pt
		\includegraphics[width=1\textwidth]{syn_result_image//14_input.jpg}
		\captionsetup{labelformat=empty}
		\captionsetup{justification=centering}
		\caption*{{Input} \\ \quad}
	\end{minipage} 
	\begin{minipage}{.118\textwidth}
		\centering
	 	\caption*{\emph{SSIM: 0.7169}}
	 	\vskip -6pt
		\includegraphics[width=1\textwidth]{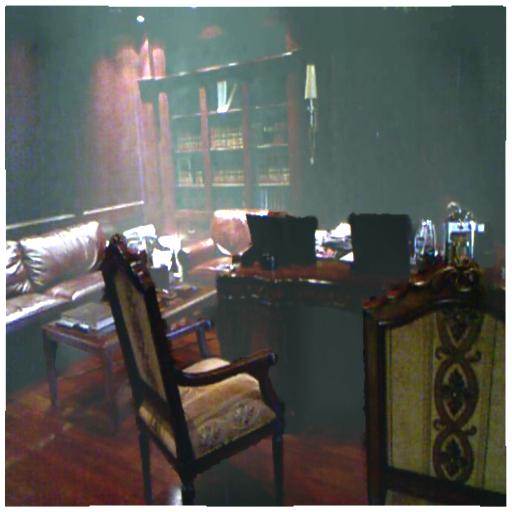}
		\captionsetup{labelformat=empty}
		\captionsetup{justification=centering}
		\caption*{{He \emph{et al.}  \\ \cite{dehaz_2009_dark}}}
	\end{minipage} 
	\begin{minipage}{.118\textwidth}
		\centering
	 	\caption*{\emph{SSIM: 0.7821}}
	 	\vskip -6pt
		\includegraphics[width=1\textwidth]{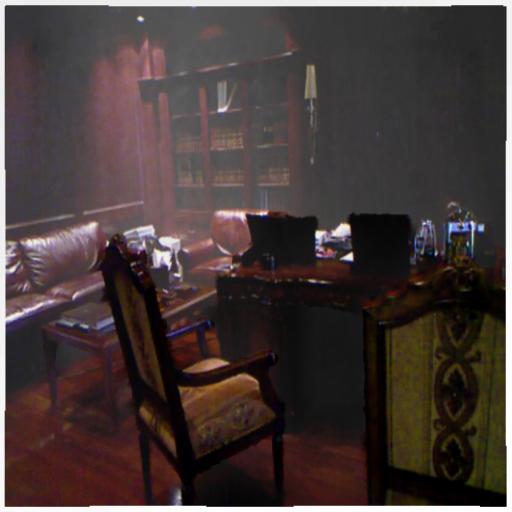}
		\captionsetup{labelformat=empty}
		\captionsetup{justification=centering}
		\caption*{{Zhu \emph{et al.}\\ \cite{dehaze_fast15}}}
	\end{minipage} 
	\begin{minipage}{.118\textwidth}
		\centering
	 	\caption*{\emph{SSIM: 0.7055}}
	 	\vskip -6pt
		\includegraphics[width=1\textwidth]{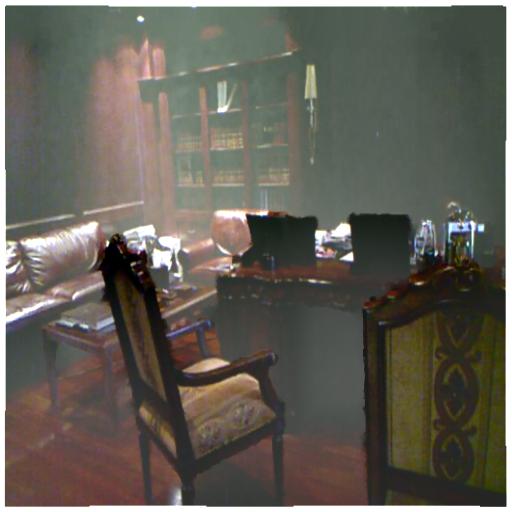}
		\captionsetup{labelformat=empty}
		\captionsetup{justification=centering}
		\caption*{{Ren \emph{et al.} \\ \cite{dehaze_2016_eccv}}}
	\end{minipage}
	\begin{minipage}{.118\textwidth}
		\centering
	 	\caption*{\emph{SSIM: 0.7232}}
	 	\vskip -6pt
		\includegraphics[width=1\textwidth]{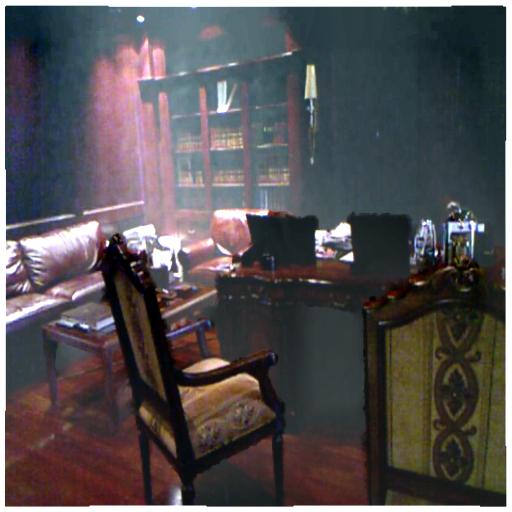}
		\captionsetup{labelformat=empty}
		\captionsetup{justification=centering}
		\caption*{{Berman \emph{et al.} \\ \cite{dehaze_2016_dana,dehaze_iccp217}}}
	\end{minipage} 
	\begin{minipage}{.118\textwidth}
		\centering
	 	\caption*{\emph{SSIM: 0.7267}}
	 	\vskip -6pt
		\includegraphics[width=1\textwidth]{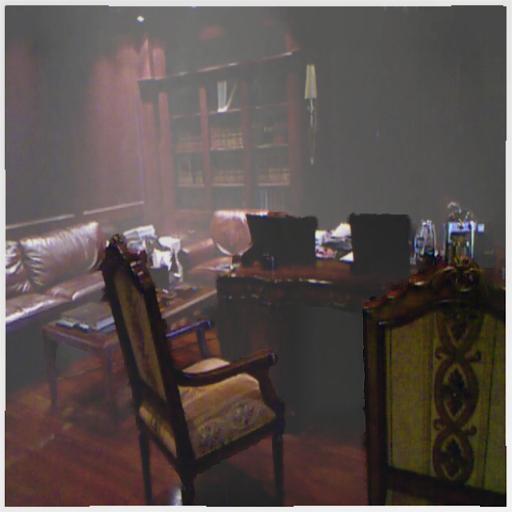}
		\captionsetup{labelformat=empty}
		\captionsetup{justification=centering}
		\caption*{{Li \emph{et al.} \\ \cite{dehaze_iccv17}}}
	\end{minipage} 
	\begin{minipage}{.118\textwidth}
		\centering
	 	\caption*{\textbf{\emph{SSIM: 0.8346}}}
	 	\vskip -6pt
		\includegraphics[width=1\textwidth]{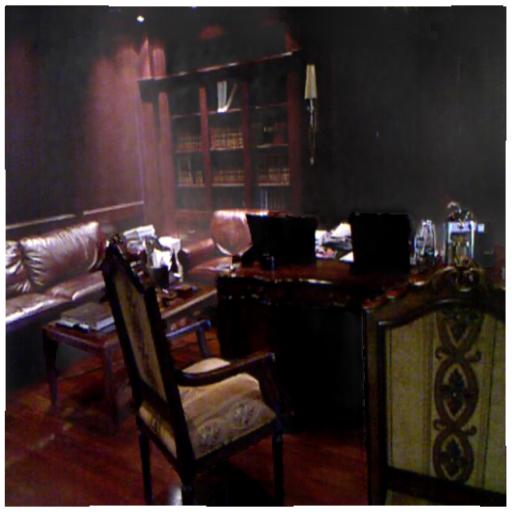}
		\captionsetup{labelformat=empty}
		\captionsetup{justification=centering}
		\caption*{{Our} \\ \qquad}
	\end{minipage}
	\begin{minipage}{.118\textwidth}
		\centering
	 	\caption*{\emph{SSIM: 1}}
	 	\vskip -6pt
		\includegraphics[width=1\textwidth]{syn_result_image//14_gt.jpg}
		\captionsetup{labelformat=empty}
		\captionsetup{justification=centering}
		\caption*{{Target} \\ \qquad}
	\end{minipage}
	
	\vskip -8pt\caption{Dehazing results from our synthetic images, where the first  row correspond to the estimated transmission map and the last  row corresponds to the dehazed image. }\label{syn11}
\end{figure*}

\begin{figure*}[!]
	\centering
	\begin{minipage}{.118\textwidth}
		\centering
	 	\caption*{\emph{SSIM: 0.3882}}
	 	\vskip -8pt
		\includegraphics[width=1\textwidth]{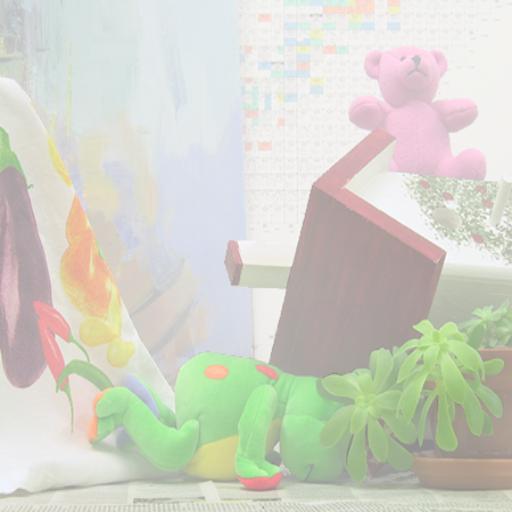}
		\captionsetup{labelformat=empty}
		\captionsetup{justification=centering}
	\end{minipage} 
	\begin{minipage}{.118\textwidth}
		\centering
	 	\caption*{\emph{SSIM: 0.7648}}
	 	\vskip -8pt
		\includegraphics[width=1\textwidth]{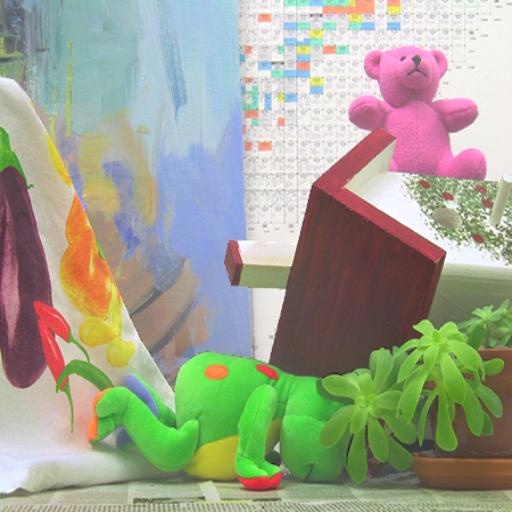}
		\captionsetup{labelformat=empty}
		\captionsetup{justification=centering}
	\end{minipage} 
	\begin{minipage}{.118\textwidth}
		\centering
	 	\caption*{\emph{SSIM: 0.7131}}
	 	\vskip -8pt
		\includegraphics[width=1\textwidth]{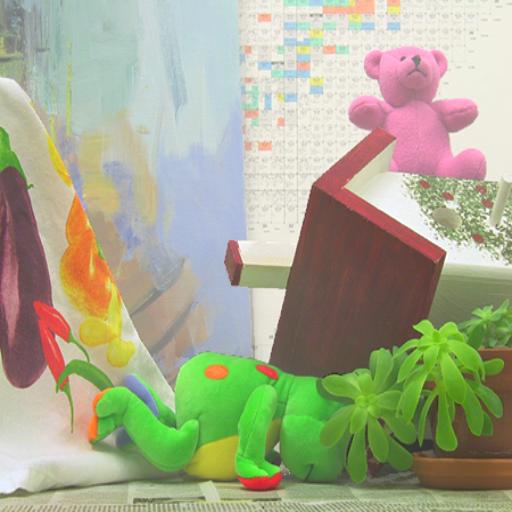}
		\captionsetup{labelformat=empty}
		\captionsetup{justification=centering}
	\end{minipage} 
	\begin{minipage}{.118\textwidth}
		\centering
	 	\caption*{\emph{SSIM: 0.7340}}
	 	\vskip -8pt
		\includegraphics[width=1\textwidth]{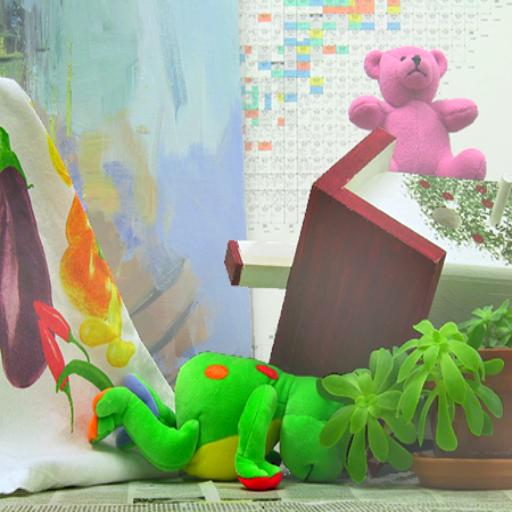}
		\captionsetup{labelformat=empty}
		\captionsetup{justification=centering}
	\end{minipage} 
	\begin{minipage}{.118\textwidth}
		\centering
	 	\caption*{\emph{SSIM: 0.7712}}
	 	\vskip -8pt
		\includegraphics[width=1\textwidth]{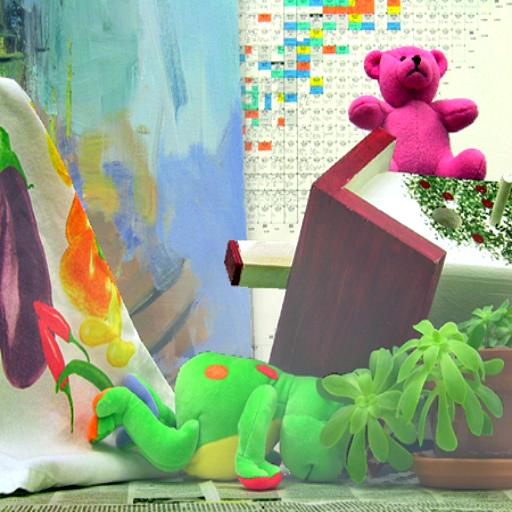}
		\captionsetup{labelformat=empty}
		\captionsetup{justification=centering}
	\end{minipage} 
	\begin{minipage}{.118\textwidth}
		\centering
	 	\caption*{\emph{SSIM: 0.6007}}
	 	\vskip -8pt
		\includegraphics[width=1\textwidth]{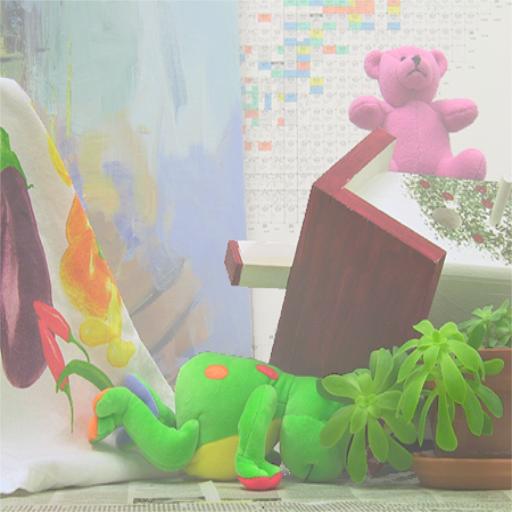}
		\captionsetup{labelformat=empty}
		\captionsetup{justification=centering}
	\end{minipage} 
	\begin{minipage}{.118\textwidth}
		\centering
	 	\caption*{\textbf{\emph{SSIM: 0.8026}}}
	 	\vskip -8pt
		\includegraphics[width=1\textwidth]{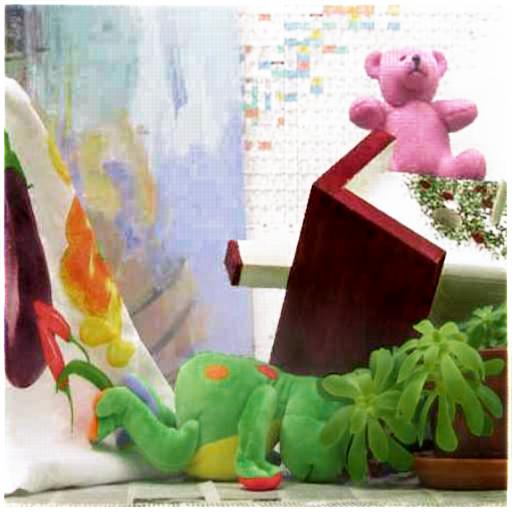}
		\captionsetup{labelformat=empty}
		\captionsetup{justification=centering}
	\end{minipage}
	\begin{minipage}{.118\textwidth}
		\centering
	 	\caption*{{\emph{SSIM: 1}}}
	 	\vskip -8pt
		\includegraphics[width=1\textwidth]{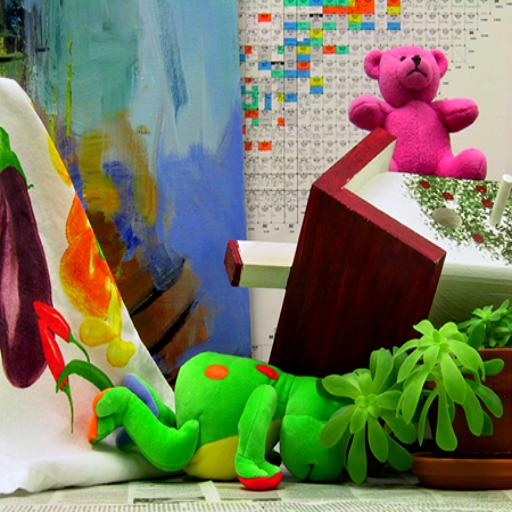}
		\captionsetup{labelformat=empty}
		\captionsetup{justification=centering}
	\end{minipage}\\	 	\vskip +5pt
	\begin{minipage}{.118\textwidth}
		\centering
	 	\caption*{\emph{SSIM: 0.3313}}
	 	\vskip -8pt
		\includegraphics[width=1\textwidth]{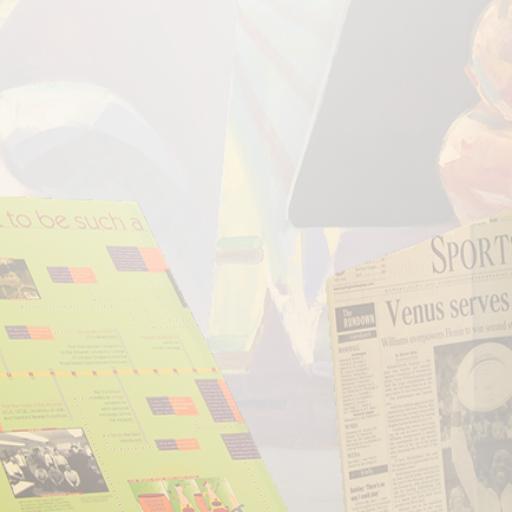}
		\captionsetup{labelformat=empty}
		\captionsetup{justification=centering}
		\caption*{\emph{Input} \\ \quad}
	\end{minipage} 
	\begin{minipage}{.118\textwidth}
		\centering
	 	\caption*{\emph{SSIM: 0.6919}}
	 	\vskip -8pt
		\includegraphics[width=1\textwidth]{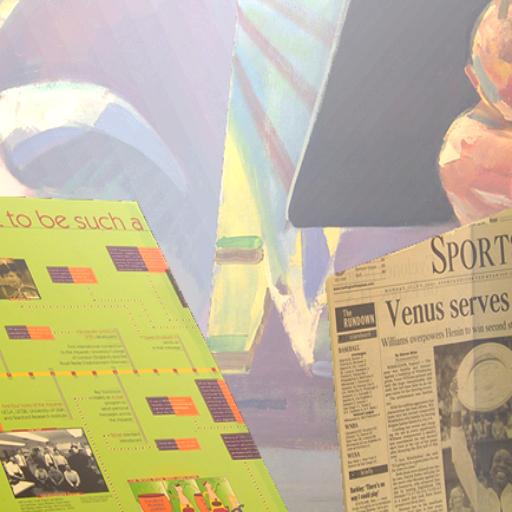}
		\captionsetup{labelformat=empty}
		\captionsetup{justification=centering}
		\caption*{{He \emph{et al.}  \\ \cite{dehaz_2009_dark}}}
	\end{minipage} 
	\begin{minipage}{.118\textwidth}
		\centering
	 	\caption*{\emph{SSIM: 0.6358}}
	 	\vskip -8pt
		\includegraphics[width=1\textwidth]{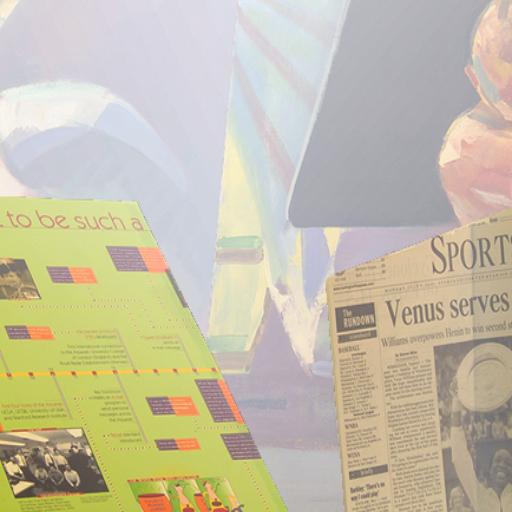}
		\captionsetup{labelformat=empty}
		\captionsetup{justification=centering}
		\caption*{{Zhu \emph{et al.}  \\ \cite{dehaze_fast15}}}
	\end{minipage} 
	\begin{minipage}{.118\textwidth}
		\centering
	 	\caption*{\emph{SSIM: 0.6282}}
	 	\vskip -8pt
		\includegraphics[width=1\textwidth]{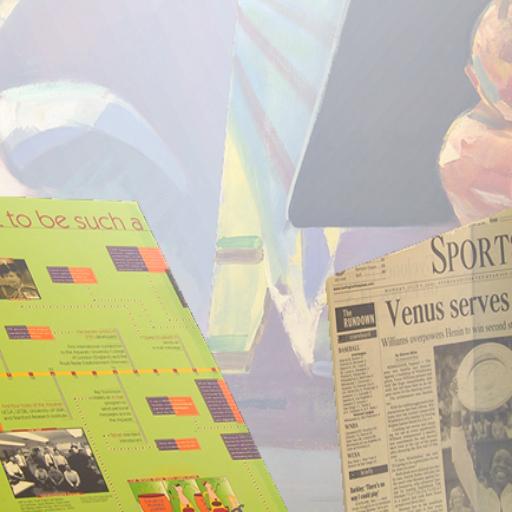}
		\captionsetup{labelformat=empty}
		\captionsetup{justification=centering}
		\caption*{{Ren \emph{et al.}  \\ \cite{dehaze_2016_eccv}}}
	\end{minipage}
	\begin{minipage}{.118\textwidth}
		\centering
	 	\caption*{\emph{SSIM: 0.6680}}
	 	\vskip -8pt
		\includegraphics[width=1\textwidth]{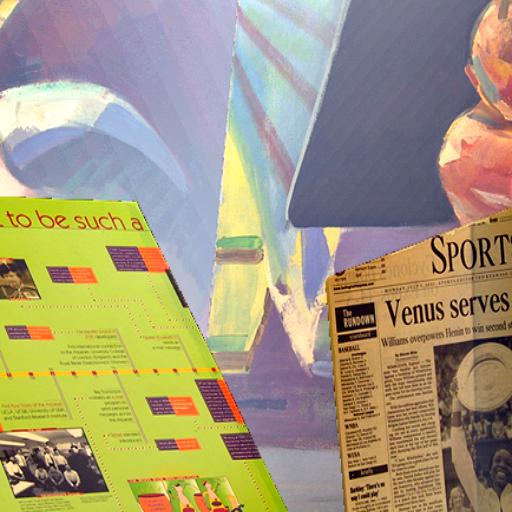}
		\captionsetup{labelformat=empty}
		\captionsetup{justification=centering}
		\caption*{{Berman \emph{et al.}  \\ \cite{dehaze_2016_dana,dehaze_iccp217}}}
	\end{minipage} 
	\begin{minipage}{.118\textwidth}
		\centering
	 	\caption*{\emph{SSIM: 0.6176}}
	 	\vskip -8pt
		\includegraphics[width=1\textwidth]{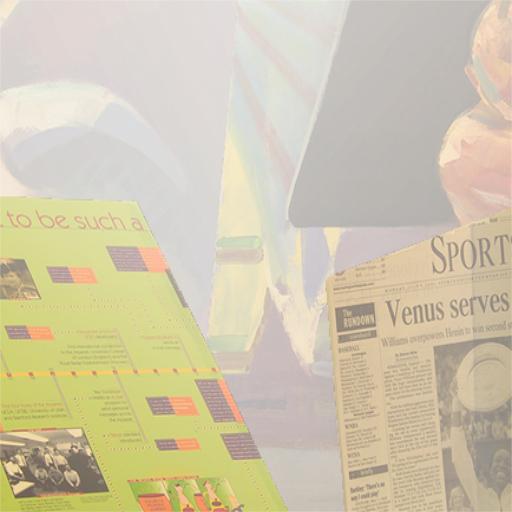}
		\captionsetup{labelformat=empty}
		\captionsetup{justification=centering}
		\caption*{{Li \emph{et al.}\\ \cite{dehaze_iccv17}}}
	\end{minipage} 
	\begin{minipage}{.118\textwidth}
		\centering
	 	\caption*{\textbf{\emph{SSIM: 0.71180}}}
	 	\vskip -8pt
		\includegraphics[width=1\textwidth]{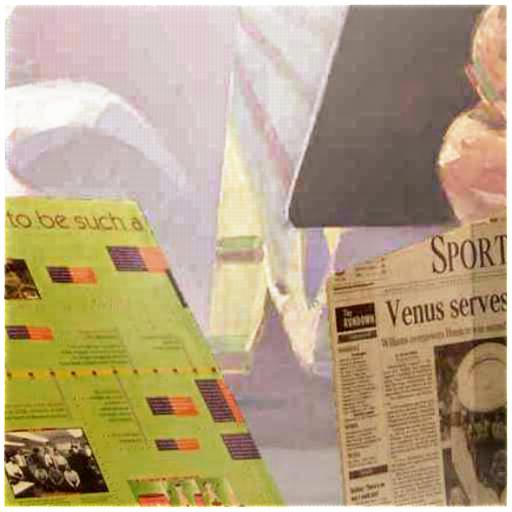}
		\captionsetup{labelformat=empty}
		\captionsetup{justification=centering}
		\caption*{{Our} \\ \qquad}
	\end{minipage}
	\begin{minipage}{.118\textwidth}
		\centering
	 	\caption*{{\emph{SSIM: 1}}}
	 	\vskip -8pt
		\includegraphics[width=1\textwidth]{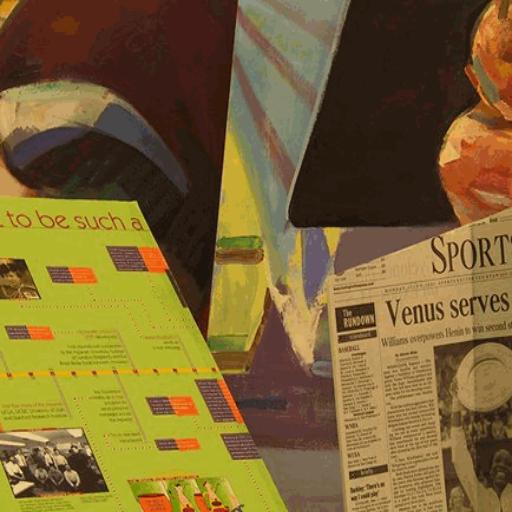}
		\captionsetup{labelformat=empty}
		\captionsetup{justification=centering}
		\caption*{{Target} \\ \qquad}
	\end{minipage}
	\vskip -8pt\caption{Dehazed visual comparisons for results of synthetic image used by previous methods \cite{dehaze_2016_tip,dehaze_iccv17}. }\label{syn51}
\end{figure*}

\begin{figure*}[t]
	\centering
	\begin{minipage}{.122\textwidth}
		\centering
		\includegraphics[width=1\textwidth]{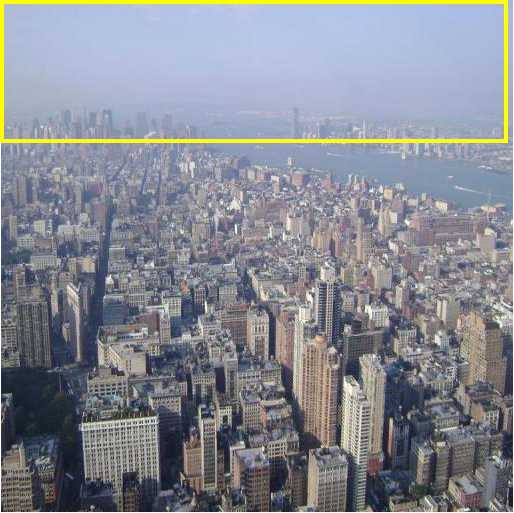}
		\captionsetup{labelformat=empty}
		\captionsetup{justification=centering}
	\end{minipage} 
	\begin{minipage}{.122\textwidth}
		\centering
		\includegraphics[width=1\textwidth]{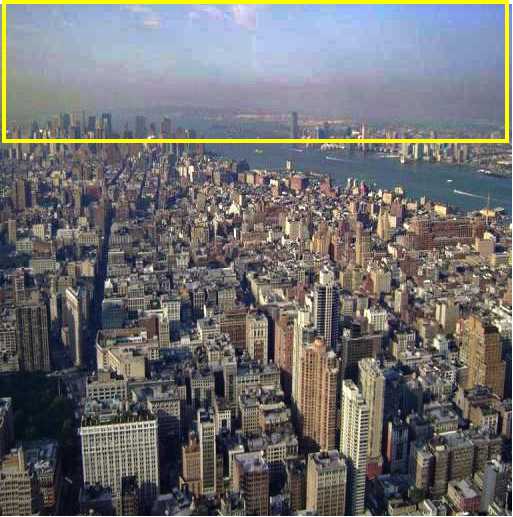}
		\captionsetup{labelformat=empty}
		\captionsetup{justification=centering}
	\end{minipage} 
	\begin{minipage}{.122\textwidth}
		\centering
		\includegraphics[width=1\textwidth]{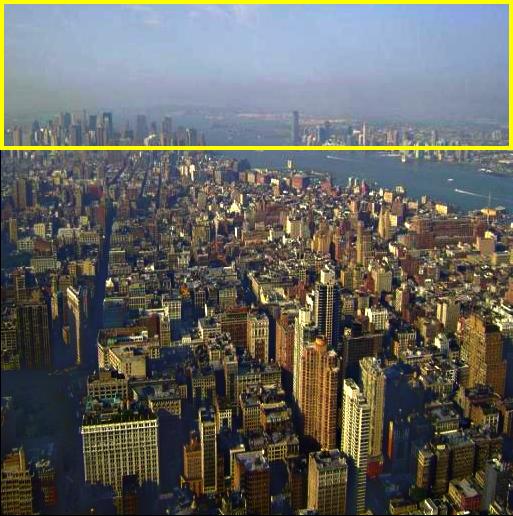}
		\captionsetup{labelformat=empty}
		\captionsetup{justification=centering}
	\end{minipage} 
	\begin{minipage}{.122\textwidth}
		\centering
		\includegraphics[width=1\textwidth]{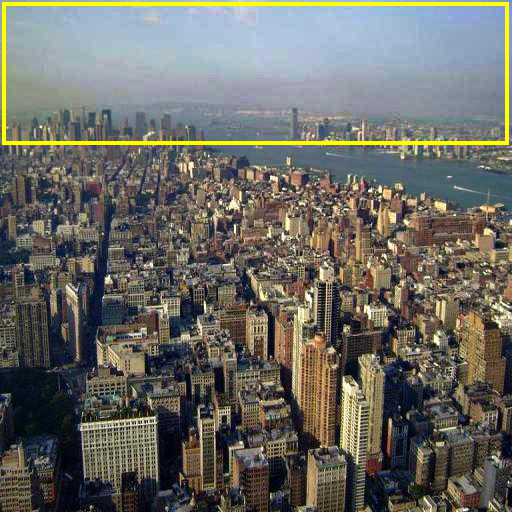}
		\captionsetup{labelformat=empty}
		\captionsetup{justification=centering}
	\end{minipage} 
	\begin{minipage}{.122\textwidth}
		\centering
		\includegraphics[width=1\textwidth]{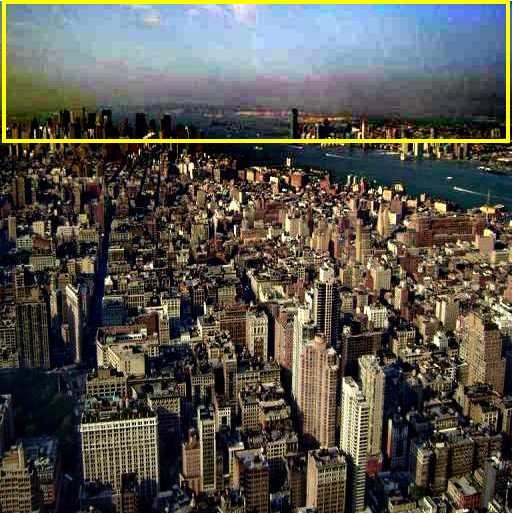}
		\captionsetup{labelformat=empty}
		\captionsetup{justification=centering}
	\end{minipage}
	\begin{minipage}{.122\textwidth}
		\centering
		\includegraphics[width=1\textwidth]{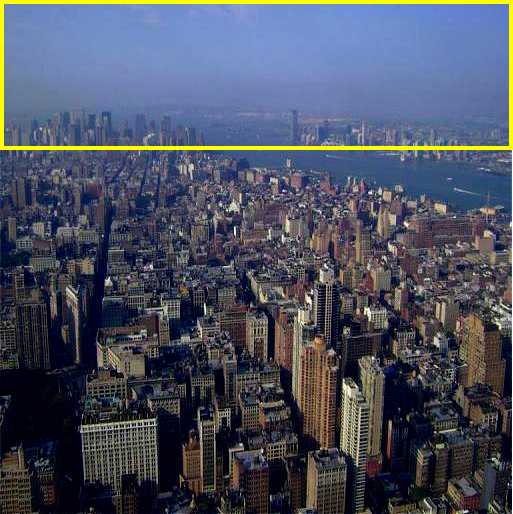}
		\captionsetup{labelformat=empty}
		\captionsetup{justification=centering}
	\end{minipage}
	\begin{minipage}{.122\textwidth}
		\centering
		\includegraphics[width=1\textwidth]{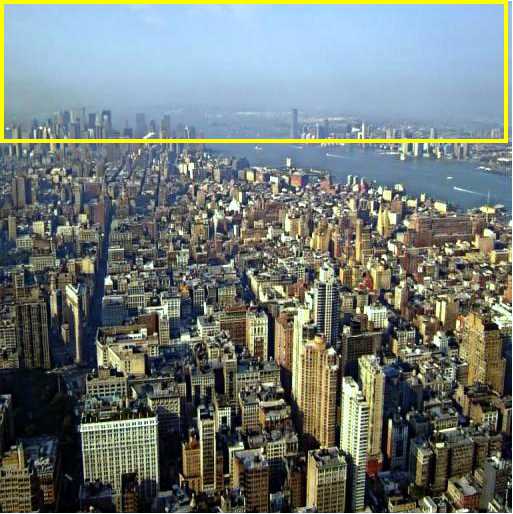}
		\captionsetup{labelformat=empty}
		\captionsetup{justification=centering}
	\end{minipage}
	\vskip+2pt
	\begin{minipage}{.122\textwidth}
		\centering
		\includegraphics[width=1\textwidth]{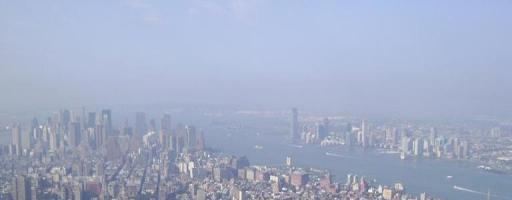}
		\captionsetup{labelformat=empty}
		\captionsetup{justification=centering}
	\end{minipage} 
	\begin{minipage}{.122\textwidth}
		\centering
		\includegraphics[width=1\textwidth]{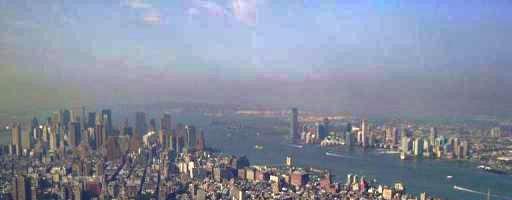}
		\captionsetup{labelformat=empty}
		\captionsetup{justification=centering}
	\end{minipage} 
	\begin{minipage}{.122\textwidth}
		\centering
		\includegraphics[width=1\textwidth]{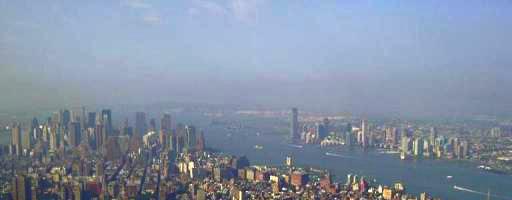}
		\captionsetup{labelformat=empty}
		\captionsetup{justification=centering}
	\end{minipage}
	\begin{minipage}{.122\textwidth}
		\centering
		\includegraphics[width=1\textwidth]{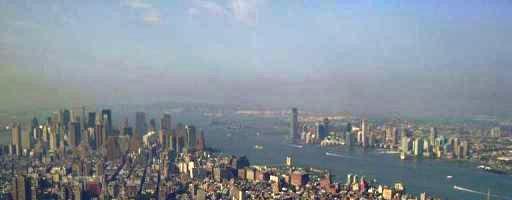}
		\captionsetup{labelformat=empty}
		\captionsetup{justification=centering}
	\end{minipage}  
	\begin{minipage}{.122\textwidth}
		\centering
		\includegraphics[width=1\textwidth]{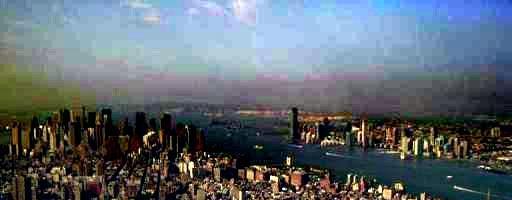}
		\captionsetup{labelformat=empty}
		\captionsetup{justification=centering}
	\end{minipage}
	\begin{minipage}{.122\textwidth}
		\centering
		\includegraphics[width=1\textwidth]{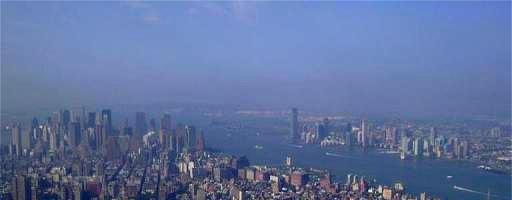}
		\captionsetup{labelformat=empty}
		\captionsetup{justification=centering}
	\end{minipage}
	\begin{minipage}{.122\textwidth}
		\centering
		\includegraphics[width=1\textwidth]{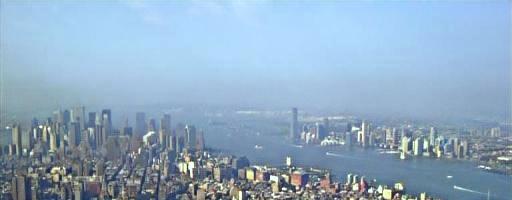}
		\captionsetup{labelformat=empty}
		\captionsetup{justification=centering}
	\end{minipage}
	\vskip+6pt
	\begin{minipage}{.122\textwidth}
		\centering
		\includegraphics[width=2.3cm,height=1.5cm]{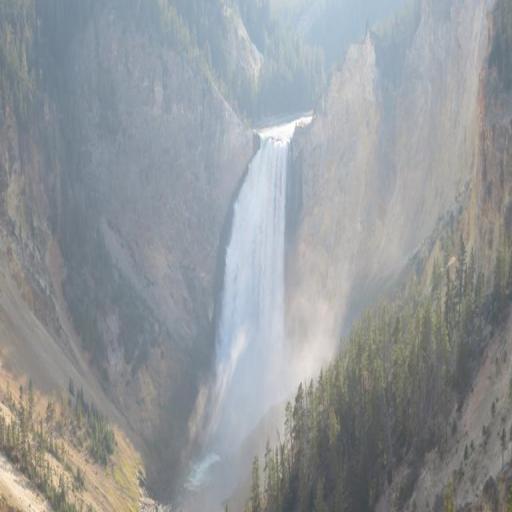}
		\captionsetup{labelformat=empty}
		\captionsetup{justification=centering}
	\end{minipage} 
	\begin{minipage}{.122\textwidth}
		\centering
		\includegraphics[width=2.3cm,height=1.5cm]{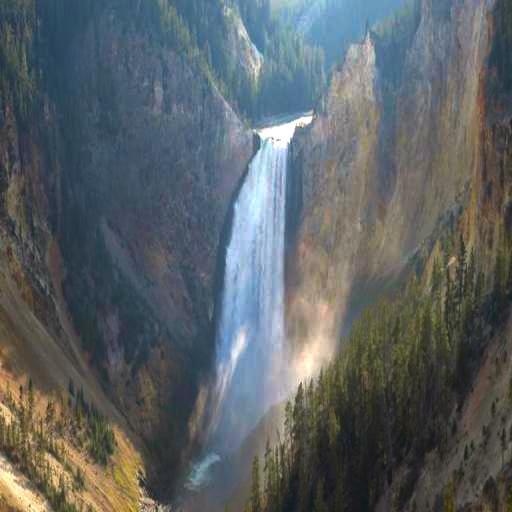}
		\captionsetup{labelformat=empty}
		\captionsetup{justification=centering}
	\end{minipage} 
	\begin{minipage}{.122\textwidth}
		\centering
		\includegraphics[width=2.3cm,height=1.5cm]{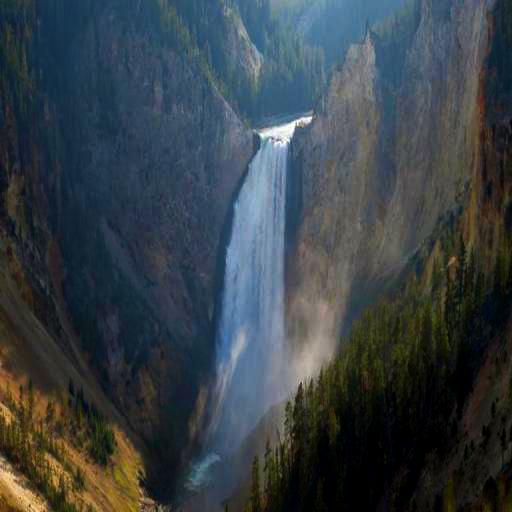}
		\captionsetup{labelformat=empty}
		\captionsetup{justification=centering}
	\end{minipage}
	\begin{minipage}{.122\textwidth}
		\centering
		\includegraphics[width=2.3cm,height=1.5cm]{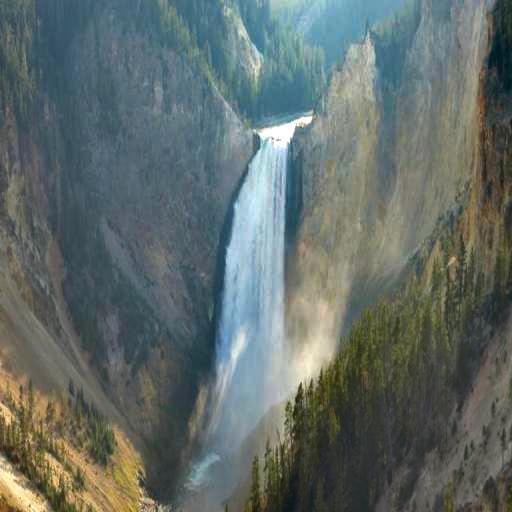}
		\captionsetup{labelformat=empty}
		\captionsetup{justification=centering}
	\end{minipage}  
	\begin{minipage}{.122\textwidth}
		\centering
		\includegraphics[width=2.3cm,height=1.5cm]{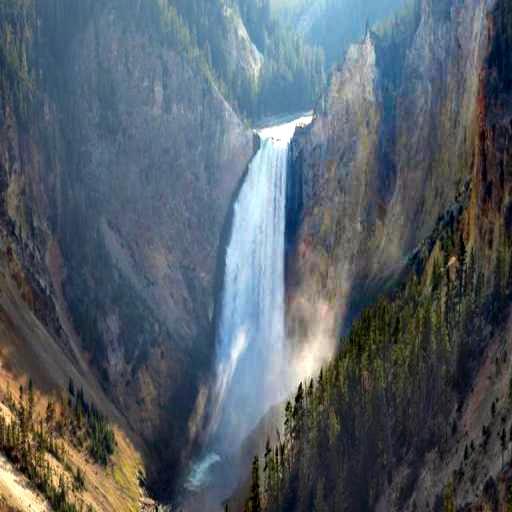}
		\captionsetup{labelformat=empty}
		\captionsetup{justification=centering}
	\end{minipage}
	\begin{minipage}{.122\textwidth}
		\centering
		\includegraphics[width=2.3cm,height=1.5cm]{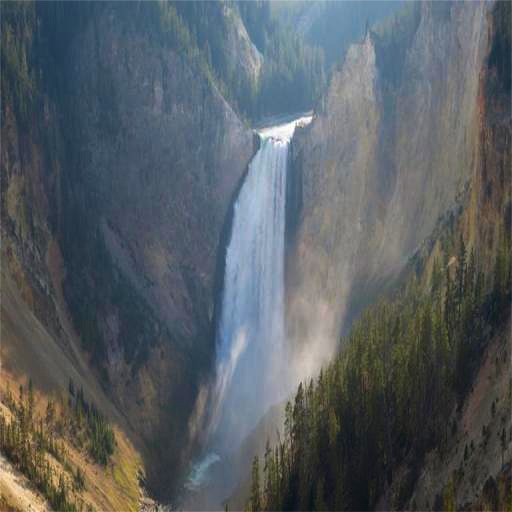}
		\captionsetup{labelformat=empty}
		\captionsetup{justification=centering}
	\end{minipage}
	\begin{minipage}{.122\textwidth}
		\centering
		\includegraphics[width=2.3cm,height=1.5cm]{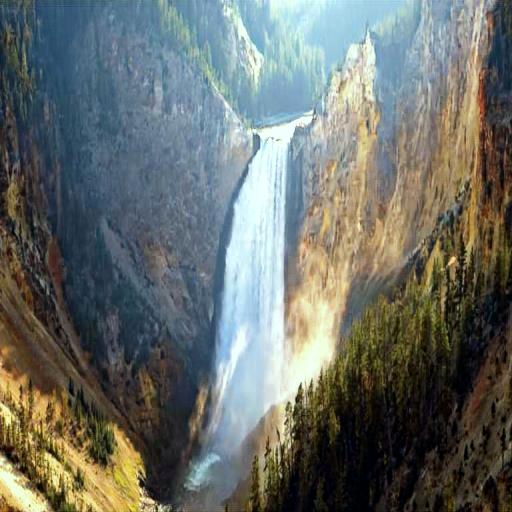}
		\captionsetup{labelformat=empty}
		\captionsetup{justification=centering}
	\end{minipage}\vskip+6pt
	\begin{minipage}{.122\textwidth}
		\centering
		\includegraphics[width=1\textwidth]{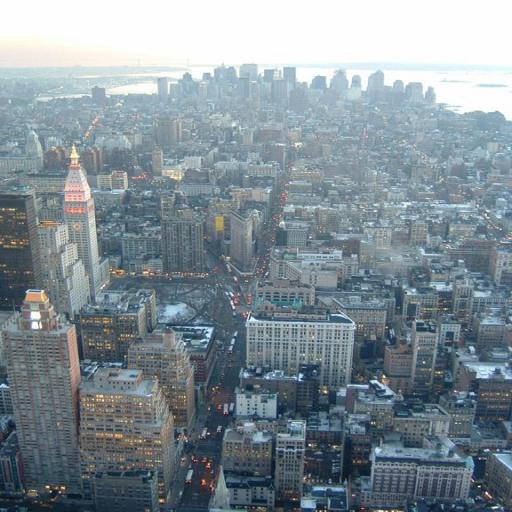}
		\captionsetup{labelformat=empty}
		\captionsetup{justification=centering}
	\end{minipage} 
	\begin{minipage}{.122\textwidth}
		\centering
		\includegraphics[width=1\textwidth]{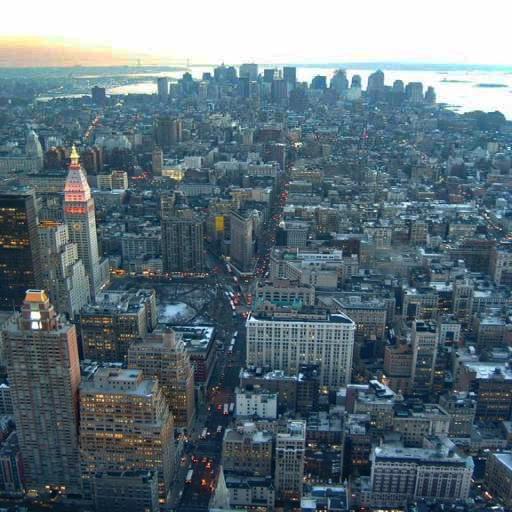}
		\captionsetup{labelformat=empty}
		\captionsetup{justification=centering}
	\end{minipage} 
	\begin{minipage}{.122\textwidth}
		\centering
		\includegraphics[width=1\textwidth]{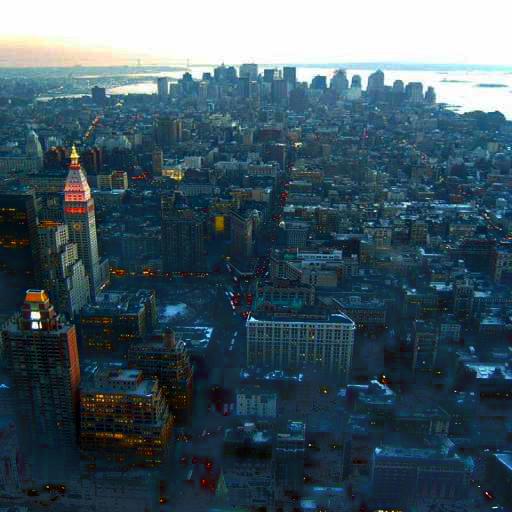}
		\captionsetup{labelformat=empty}
		\captionsetup{justification=centering}
	\end{minipage} 
	\begin{minipage}{.122\textwidth}
		\centering
		\includegraphics[width=1\textwidth]{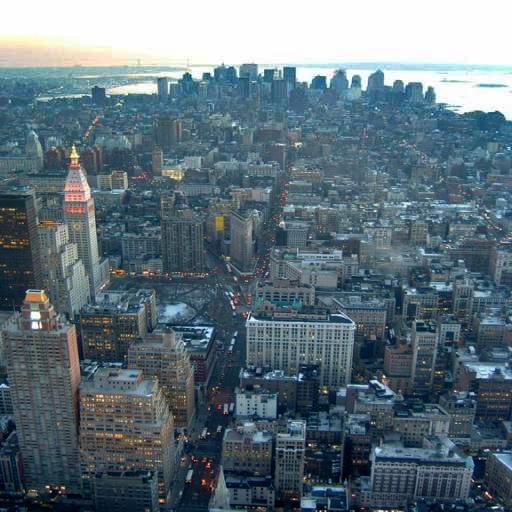}
		\captionsetup{labelformat=empty}
		\captionsetup{justification=centering}
	\end{minipage} 
	\begin{minipage}{.122\textwidth}
		\centering
		\includegraphics[width=1\textwidth]{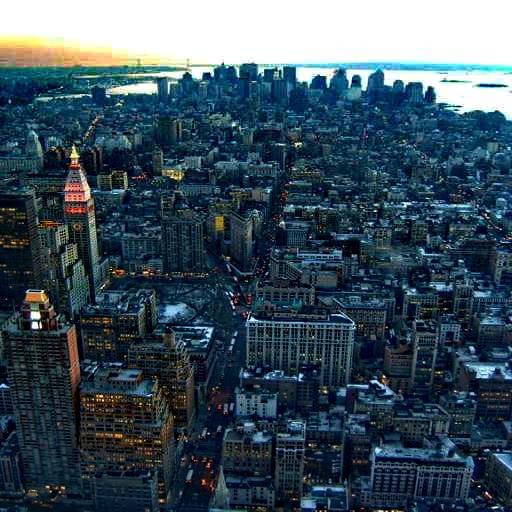}
		\captionsetup{labelformat=empty}
		\captionsetup{justification=centering}
	\end{minipage}
	\begin{minipage}{.122\textwidth}
		\centering
		\includegraphics[width=1\textwidth]{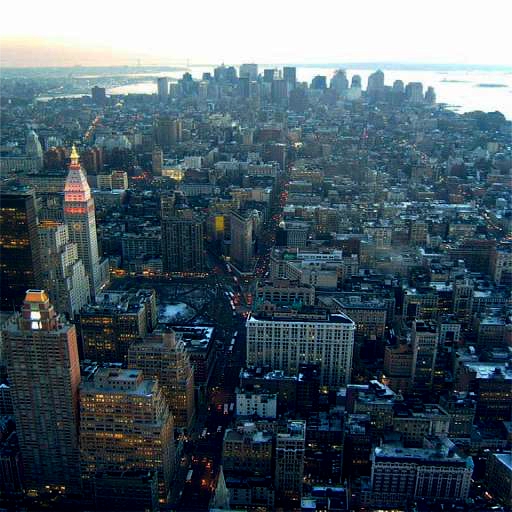}
		\captionsetup{labelformat=empty}
		\captionsetup{justification=centering}
	\end{minipage}
	\begin{minipage}{.122\textwidth}
		\centering
		\includegraphics[width=1\textwidth]{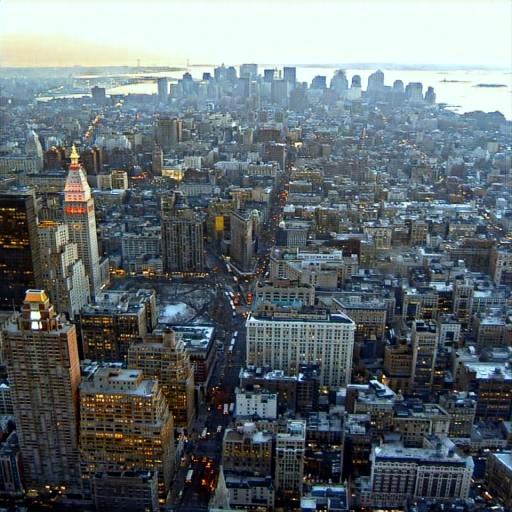}
		\captionsetup{labelformat=empty}
		\captionsetup{justification=centering}
	\end{minipage}		\vskip+6pt
	\begin{minipage}{.122\textwidth}
		\centering
		\includegraphics[width=2.3cm,height=1cm]{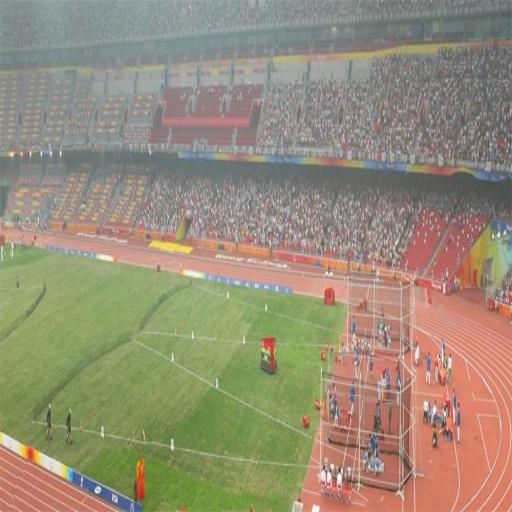}
		\captionsetup{labelformat=empty}
		\captionsetup{justification=centering}
		\caption*{Input\\ \quad}
	\end{minipage} 
	\begin{minipage}{.122\textwidth}
		\centering
		\includegraphics[width=2.3cm,height=1cm]{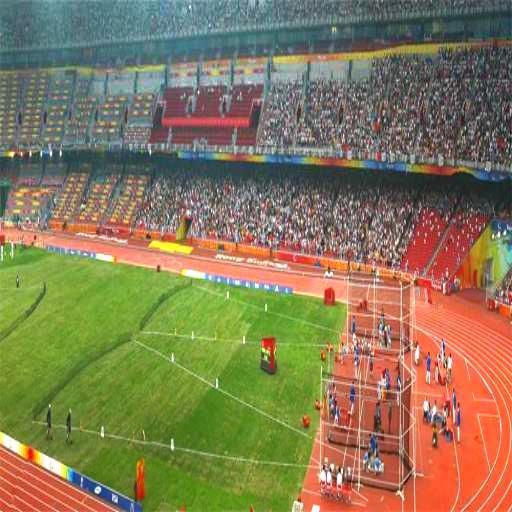}
		\captionsetup{labelformat=empty}
		\captionsetup{justification=centering}
		\caption*{He. \emph{et al.}\\ \cite{dehaz_2009_dark}}
	\end{minipage} 
	\begin{minipage}{.122\textwidth}
		\centering
		\includegraphics[width=2.3cm,height=1cm]{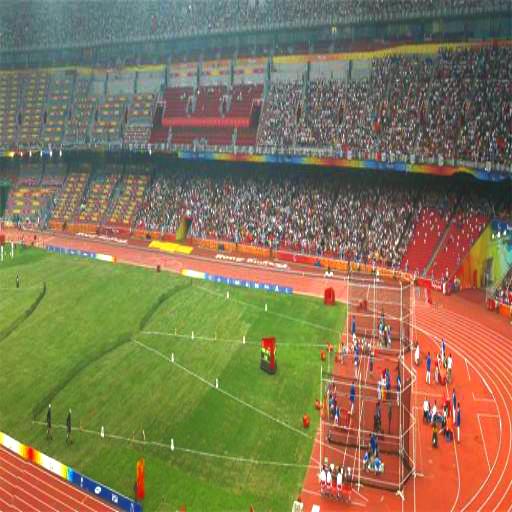}
		\captionsetup{labelformat=empty}
		\captionsetup{justification=centering}
		\caption*{Zhu. \emph{et al.}\\ \cite{dehaze_fast15}}
	\end{minipage} 
	\begin{minipage}{.122\textwidth}
		\centering
		\includegraphics[width=2.3cm,height=1cm]{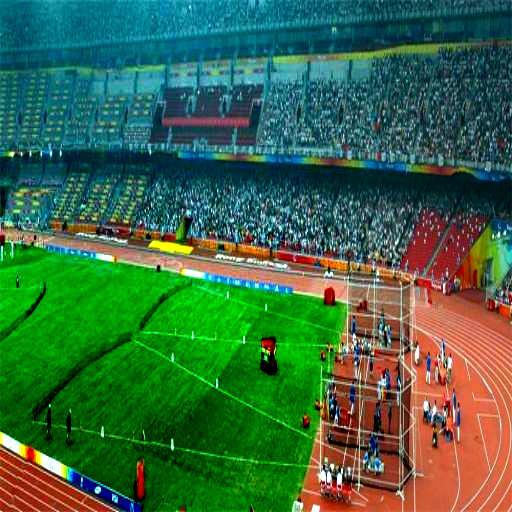}
		\captionsetup{labelformat=empty}
		\captionsetup{justification=centering}
		\caption*{Ren. \emph{et al.}\\ \cite{dehaze_2016_eccv}}
	\end{minipage} 
	\begin{minipage}{.122\textwidth}
		\centering
		\includegraphics[width=2.3cm,height=1cm]{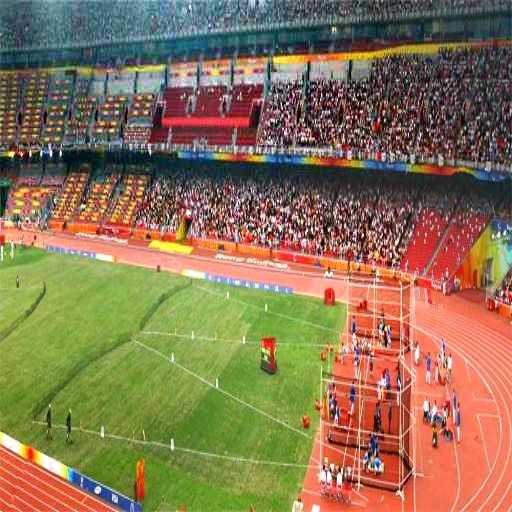}
		\captionsetup{labelformat=empty}
		\captionsetup{justification=centering}
		\caption*{Berman. \emph{et al.}\\ \cite{dehaze_2016_dana,dehaze_iccp217} }
	\end{minipage}
	\begin{minipage}{.122\textwidth}
		\centering
		\includegraphics[width=2.3cm,height=1cm]{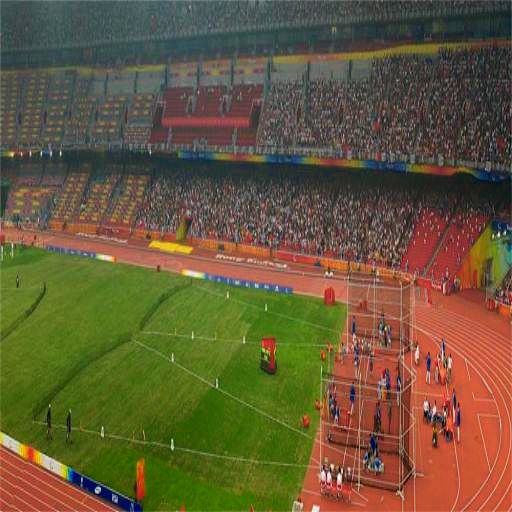}
		\captionsetup{labelformat=empty}
		\captionsetup{justification=centering}
		\caption*{Li. \emph{et.al} \\ \cite{dehaze_iccv17}}
	\end{minipage}
	\begin{minipage}{.122\textwidth}
		\centering
		\includegraphics[width=2.3cm,height=1cm]{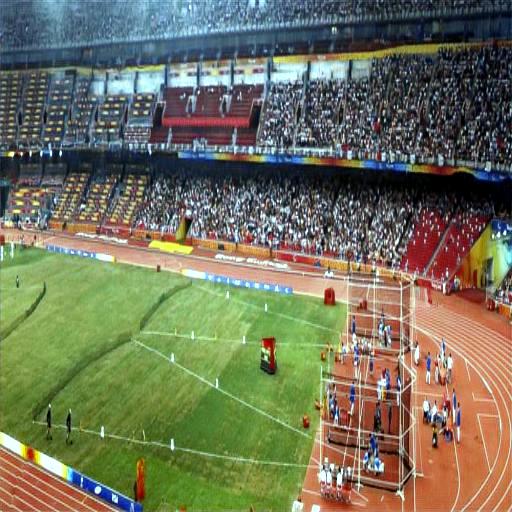}
		\captionsetup{labelformat=empty}
		\captionsetup{justification=centering}
		\caption*{Our \\ \quad}
	\end{minipage}
	
	\caption{Qualitative comparison of dehazing on real-world dataset that is presented in previous dehazing papers. It can be observed from the highlighted region that previous methods may result in undesirable effects such as artifacts and color over-saturation in the output images }\label{nat_hazepre}
	\end{figure*}

\begin{figure*}
\centering
	\begin{minipage}{.122\textwidth}
		\centering
		\includegraphics[width=2.3cm,height=2cm]{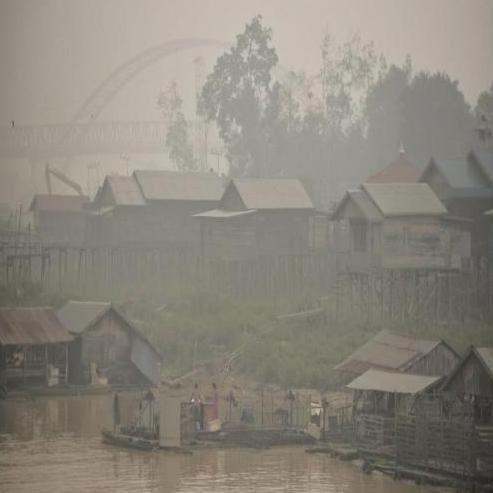}
		\captionsetup{labelformat=empty}
		\captionsetup{justification=centering}
	\end{minipage} 
	\begin{minipage}{.122\textwidth}
		\centering
		\includegraphics[width=2.3cm,height=2cm]{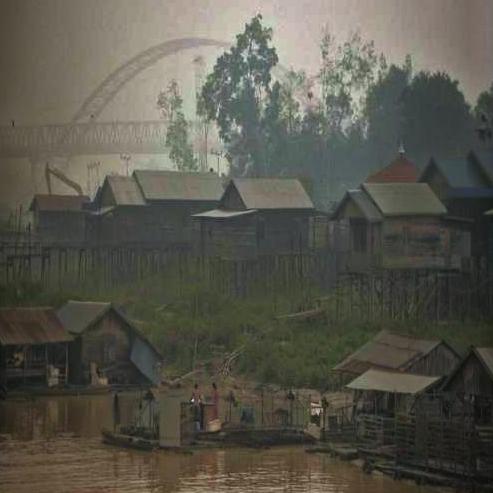}
		\captionsetup{labelformat=empty}
		\captionsetup{justification=centering}
	\end{minipage} 
	\begin{minipage}{.122\textwidth}
		\centering
		\includegraphics[width=2.3cm,height=2cm]{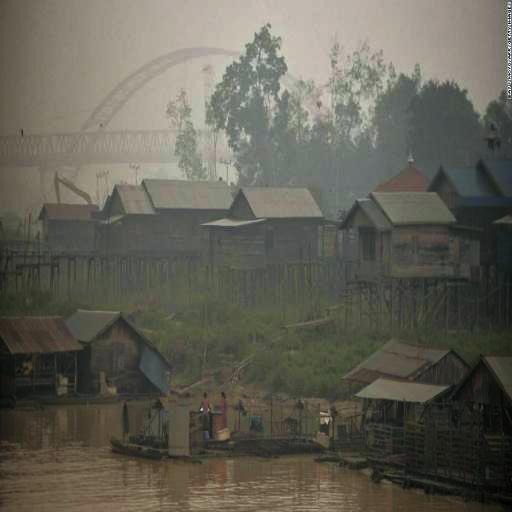}
		\captionsetup{labelformat=empty}
		\captionsetup{justification=centering}
	\end{minipage} 
	\begin{minipage}{.122\textwidth}
		\centering
		\includegraphics[width=2.3cm,height=2cm]{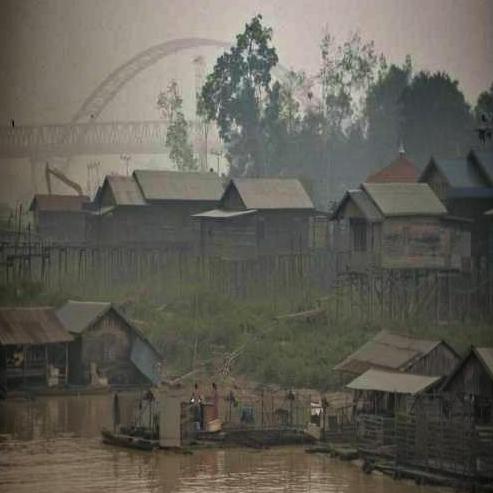}
		\captionsetup{labelformat=empty}
		\captionsetup{justification=centering}
	\end{minipage} 
	\begin{minipage}{.122\textwidth}
		\centering
		\includegraphics[width=2.3cm,height=2cm]{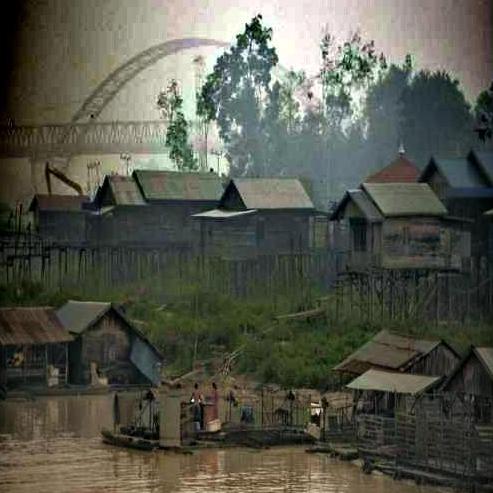}
		\captionsetup{labelformat=empty}
		\captionsetup{justification=centering}
	\end{minipage}
	\begin{minipage}{.122\textwidth}
		\centering
		\includegraphics[width=2.3cm,height=2cm]{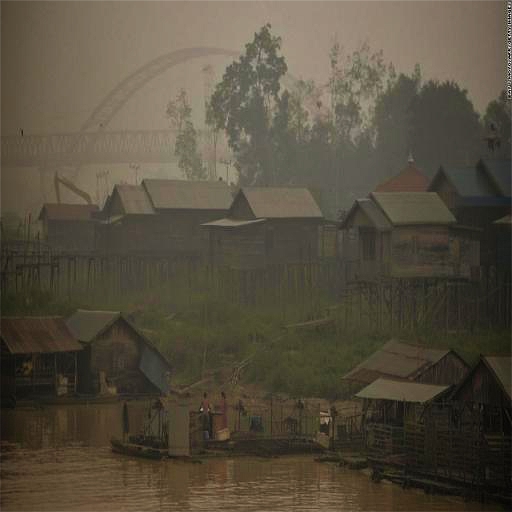}
		\captionsetup{labelformat=empty}
		\captionsetup{justification=centering}
	\end{minipage} 
	\begin{minipage}{.122\textwidth}
		\centering
		\includegraphics[width=2.3cm,height=2cm]{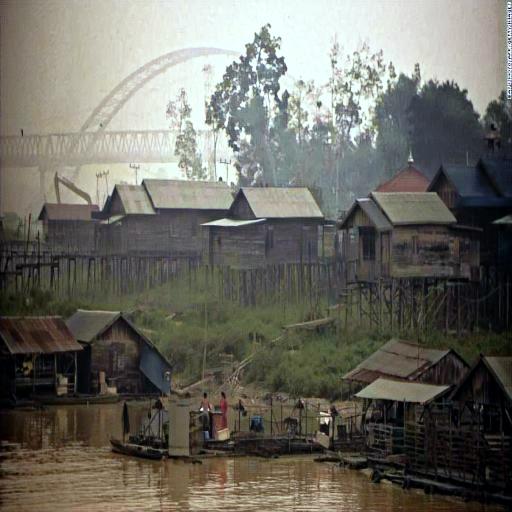}
		\captionsetup{labelformat=empty}
		\captionsetup{justification=centering}
	\end{minipage}	\\		\vskip+6pt
	\begin{minipage}{.122\textwidth}
		\centering
		\includegraphics[width=2.3cm,height=1.5cm]{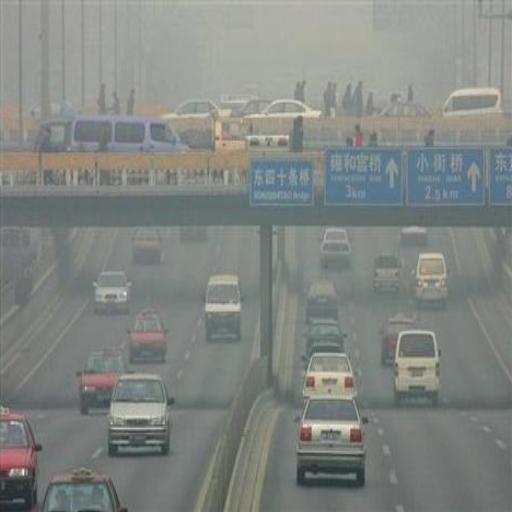}
		\captionsetup{labelformat=empty}
		\captionsetup{justification=centering}
		\caption*{Input \\ \quad}
	\end{minipage} 
	\begin{minipage}{.122\textwidth}
		\centering
		\includegraphics[width=2.3cm,height=1.5cm]{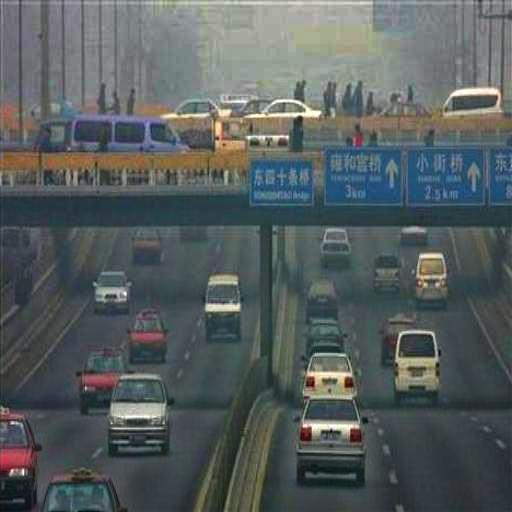}
		\captionsetup{labelformat=empty}
		\captionsetup{justification=centering}
		\caption*{He. \emph{et al.} \\\cite{dehaz_2009_dark}}
	\end{minipage} 
	\begin{minipage}{.122\textwidth}
		\centering
		\includegraphics[width=2.3cm,height=1.5cm]{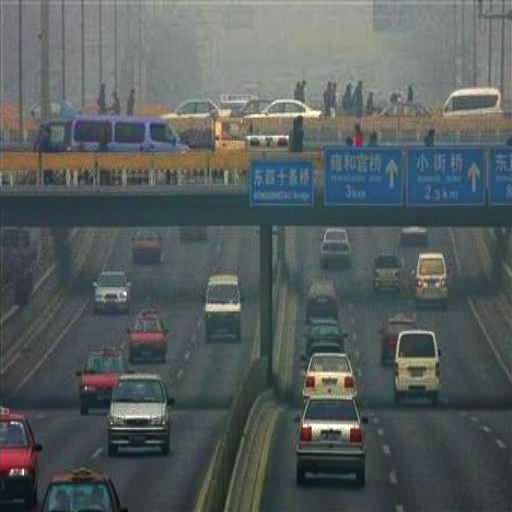}
		\captionsetup{labelformat=empty}
		\captionsetup{justification=centering}
		\caption*{Zhu. \emph{et al.}\\ \cite{dehaze_fast15}}
	\end{minipage} 
	\begin{minipage}{.122\textwidth}
		\centering
		\includegraphics[width=2.3cm,height=1.5cm]{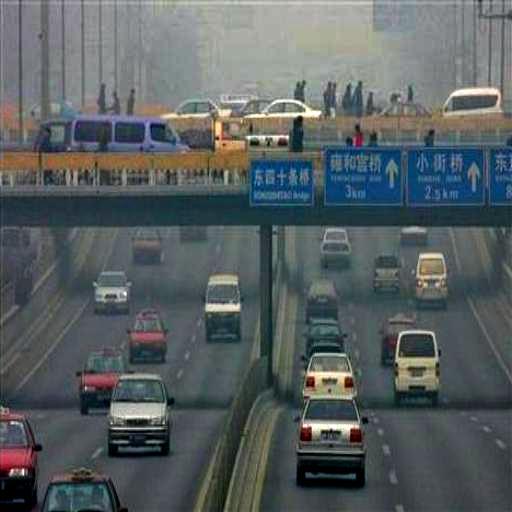}
		\captionsetup{labelformat=empty}
		\captionsetup{justification=centering}
		\caption*{Ren. \emph{et al.}\\ \cite{dehaze_2016_eccv}}
	\end{minipage}
	\begin{minipage}{.122\textwidth}
		\centering
		\includegraphics[width=2.3cm,height=1.5cm]{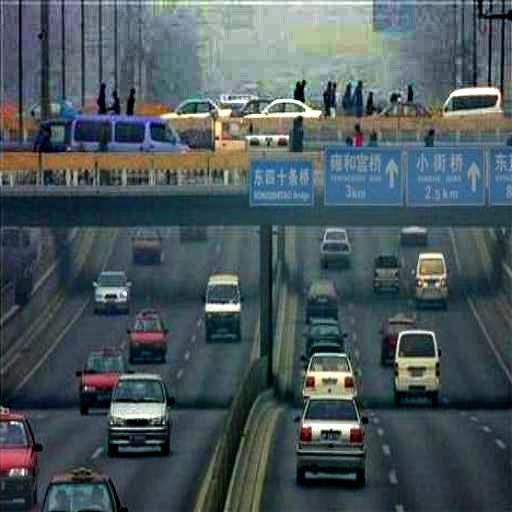}
		\captionsetup{labelformat=empty}
		\captionsetup{justification=centering}
		\caption*{Berman. \emph{et al.}\\ \cite{dehaze_2016_dana,dehaze_iccp217} }
	\end{minipage} 
	\begin{minipage}{.122\textwidth}
		\centering
		\includegraphics[width=2.3cm,height=1.5cm]{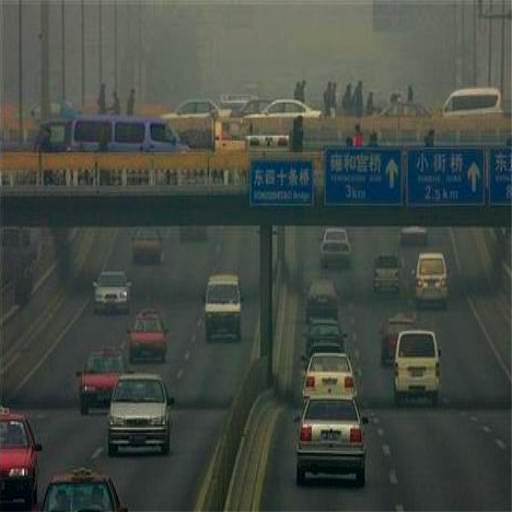}
		\captionsetup{labelformat=empty}
		\captionsetup{justification=centering}
		\caption*{Li. \emph{et.al}\\ \cite{dehaze_iccv17}}
	\end{minipage} 
	\begin{minipage}{.122\textwidth}
		\centering
		\includegraphics[width=2.3cm,height=1.5cm]{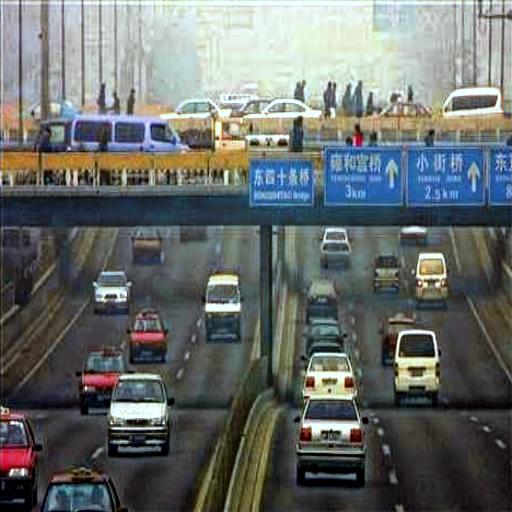}
		\captionsetup{labelformat=empty}
		\captionsetup{justification=centering}
		\caption*{Our \\ \quad}
	\end{minipage}
	\caption{Qualitative comparison of dehazing on real-world dataset. Results on two sample images from a set of images downloaded from the Internet. }\label{nat_haze4}
\end{figure*}

\subsection{Comparison with state-of-the-art Methods}
To demonstrate the improvements achieved by the proposed method, it is compared against recent state-of-the-art methods on synthetic and real datasets.\\ 

\noindent \textbf{Evaluation on synthetic dataset:} Synthetic dataset,   as described in Section IV(A), is used for the purpose of training and evaluating the network. Due to the availability of ground-truth images, we conduct both qualitative and quantitative evaluations. 


Figure \ref{syn11} shows results of the proposed method as compared with recent state-of-the-art methods (\cite{dehaz_2009_dark,dehaze_fast15,dehaze_2016_dana,dehaze_iccp217,dehaze_2016_eccv,dehaze_iccv17} ) on a sample image from the test split of the synthetic dataset. After carefully analyzing these results, we observed that the recent best methods resulted in either incomplete removal of haze or over-correction which reduced the visual appeal of the image. Even though, \cite{dehaze_2016_dana} is able to achieve good performance in the presence of moderate haze, its dehazed results tend to contain color shift. In contrast, the proposed method is able to achieve better dehazing for a variety of haze contents. Similar results can be observed regarding the quality of transmission maps estimated by the proposed multi-task method as compared with the existing methods. It can be noted that the previous methods are unable to accurately estimate the relative depth in a given image, resulting in lower quality of dehazed images. In contrast, the proposed method not only estimates high quality transmission maps, but also achieves better quality dehazing.

The quantitative performance of the proposed method is compared against five state-of-the-art methods \cite{dehaz_2009_dark,dehaze_fast15,dehaze_2016_eccv,dehaze_2016_dana,dehaze_iccv17} using  SSIM \cite{ssim}. The quantitative results are tabulated in Table \ref{ta:quantitive_depth1}.  It can be observed from this table that the proposed method achieves the best performance in terms SSIM. Note that, we have attempted to obtain the best possible results for the other methods by fine-tuning their respective parameters based on the source code released by the authors and kept the parameter consistent for all the experiments. As the code released by \cite{dehaze_2016_eccv,dehaze_iccv17} cannot estimate the predicted transmission map, the results for the transmission estimation corresponding to \cite{dehaze_2016_eccv,dehaze_iccv17} is not included in the discussion.\\

Furthermore, we also evaluate the proposed method on the synthetic images used by previous methods \cite{dehaze_2016_tip,dehaze_iccv17}. Results are shown in Fig~\ref{syn51}. It can be clearly observed that Berman \emph{et al.} \cite{dehaze_2016_dana,dehaze_iccp217} and the proposed methods achieve the best visual performance among all. However, by looking closer at the upper right part of Fig~\ref{syn51}, it can be found that method from Berman \emph{et al.} \cite{dehaze_2016_dana,dehaze_iccp217} tend to bring in the color-shift and hence degrade the overall performance.\\ 

\begin{figure*}
\centering
	\begin{minipage}{.24\textwidth}
		\centering
		\includegraphics[width=1\textwidth]{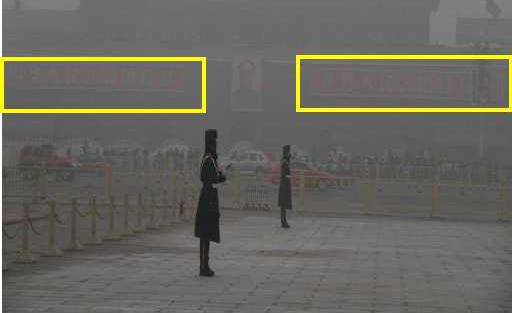}
		\captionsetup{labelformat=empty}
		\captionsetup{justification=centering}
		\caption*{Input\\\quad}
	\end{minipage} 
	\begin{minipage}{.24\textwidth}
		\centering
		\includegraphics[width=1\textwidth]{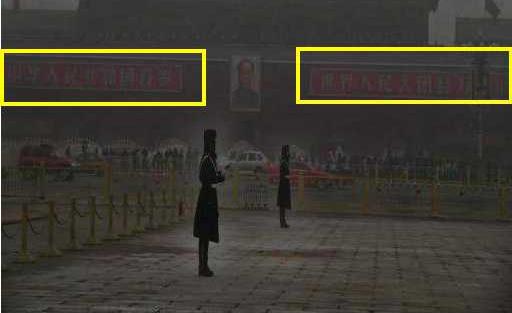}
		\captionsetup{labelformat=empty}
		\captionsetup{justification=centering}
		\caption*{He. \emph{et.al}\\ \cite{dehaz_2009_dark}}
	\end{minipage} 
	\begin{minipage}{.24\textwidth}
		\centering
		\includegraphics[width=1\textwidth]{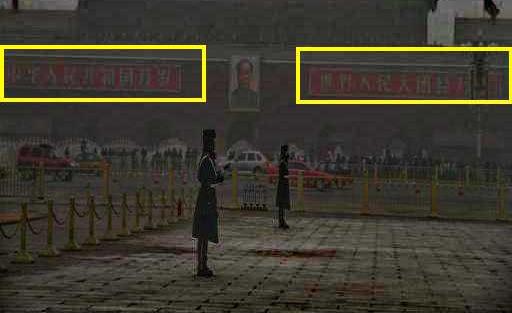}
		\captionsetup{labelformat=empty}
		\captionsetup{justification=centering}
		\caption*{Zhu. \emph{et al.}\\ \cite{dehaze_fast15}}
	\end{minipage} \\		\vskip+6pt
	\begin{minipage}{.24\textwidth}
		\centering
		\includegraphics[width=1\textwidth]{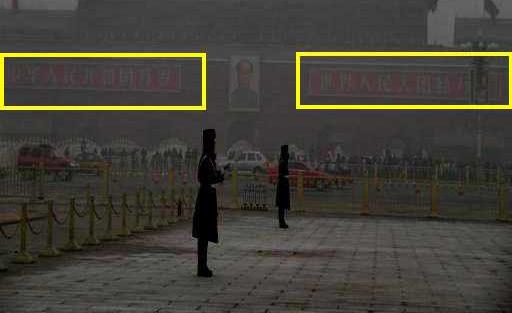}
		\captionsetup{labelformat=empty}
		\captionsetup{justification=centering}
		\caption*{Ren. \emph{et.al}\\ \cite{dehaze_2016_eccv}}
	\end{minipage} 
	\begin{minipage}{.24\textwidth}
		\centering
		\includegraphics[width=1\textwidth]{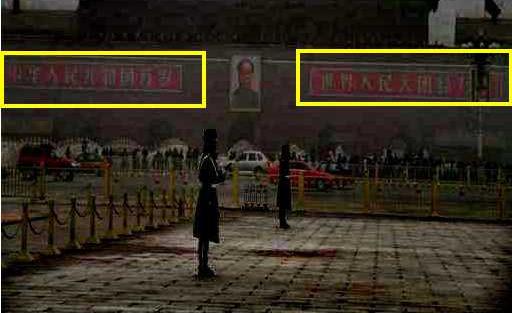}
		\captionsetup{labelformat=empty}
		\captionsetup{justification=centering}
		\caption*{Berman. \emph{et.al} \\\cite{dehaze_2016_dana,dehaze_iccp217} }
	\end{minipage}
	\begin{minipage}{.24\textwidth}
		\centering
		\includegraphics[width=1\textwidth]{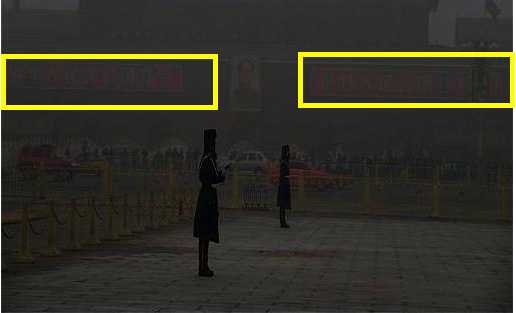}
		\captionsetup{labelformat=empty}
		\captionsetup{justification=centering}
		\caption*{Li. \emph{et.al} \\\cite{dehaze_iccv17}}
	\end{minipage} 
	\begin{minipage}{.24\textwidth}
		\centering
		\includegraphics[width=1\textwidth]{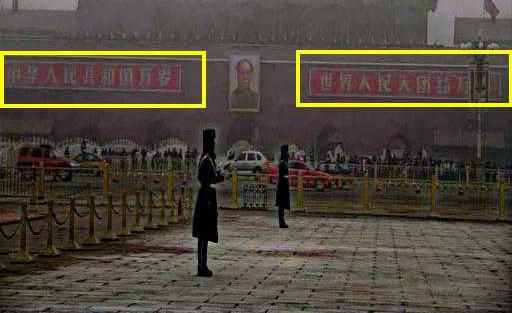}
		\captionsetup{labelformat=empty}
		\captionsetup{justification=centering}
		\caption*{Our \\ \quad}
	\end{minipage}
	\caption{Qualitative comparison of dehazing on real-world dataset. Top row: Results on a sample image from the real-world dataset provided by previous methods. Bottom two rows: Results on two sample images from a set of images downloaded from the Internet.}\label{nat_tough}
\end{figure*}

\noindent \textbf{Evaluation on real dataset:} In addition to the synthetic dataset, we also conducted evaluation experiments on real dataset which consists of hazy images from the real world, collected from the internet. Since the ground truths are not available for such images, we do not use this dataset for training and we perform only qualitative evaluations. 


\begin{figure}
	\centering
	\begin{minipage}{.24\textwidth}
		\centering
		\includegraphics[width=1\textwidth]{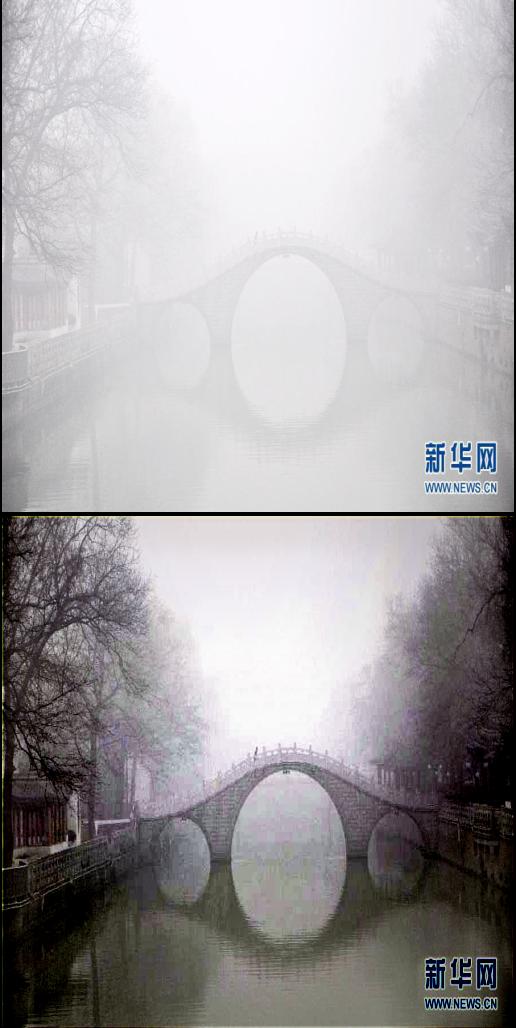}
		\captionsetup{labelformat=empty}
		\captionsetup{justification=centering}
	\end{minipage} 
	\begin{minipage}{.24\textwidth}
	\centering
	\includegraphics[width=1\textwidth]{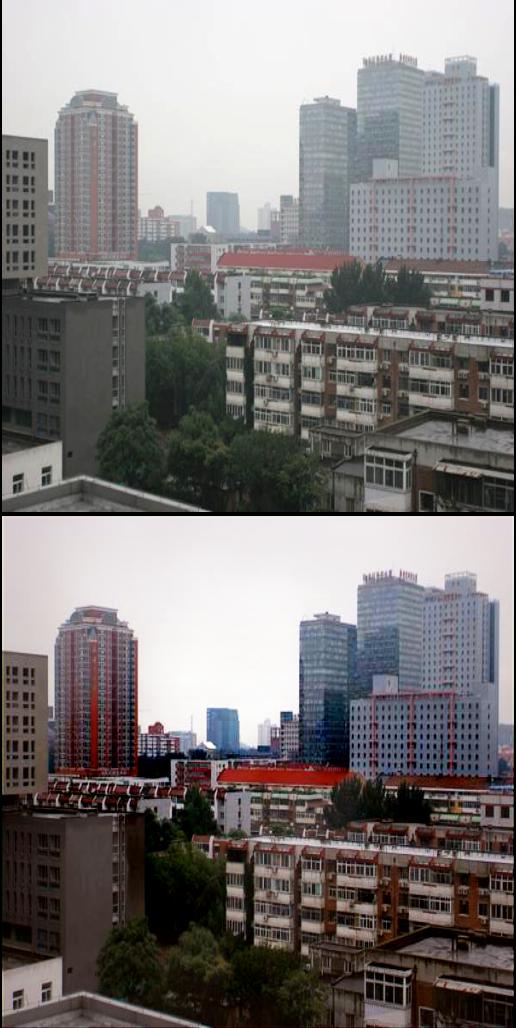}
	\captionsetup{labelformat=empty}
	\captionsetup{justification=centering}
\end{minipage}
	\caption{More dehazing results on the real-world images. The first row show the original hazy image and second row show dehazed images of the proposed method.  }\label{nat_more}
\end{figure}

Comparison of results on four sample images used in earlier methods compared with various approaches is shown in Figure~\ref{nat_hazepre}. Yellow rectangles are used to highlight the improvements obtained using the proposed method. Though the existing methods seem to achieve good visual performance in the top row, it can be observed from the highlighted region that these methods may result in undesirable effects such as artifacts and color over-saturation in the output images. For the bottom two rows, the existing methods either make the image darker due to overestimation of dark pixels or are unable to perform complete dehazing. For example, leaning-based methods \cite{dehaze_2016_eccv,dehaze_iccv17} underestimate the thickness of haze resulting in partial dehazing.   Even though Berman \emph{et al.} \cite{dehaze_2016_dana,dehaze_iccp217} leaves less haze in the output, the resulting image tends to be darker as the haze line is tough to detect under heavy haze conditions. In contrast, the proposed method is able to achieve near-complete dehazing with visually appealing results by avoiding any  undesirable effects in the output images.

Furthermore, we also illustrate three qualitative examples of dehazing results on real-world hazy images by different methods. He. \emph{et al} \cite{dehaz_2009_dark}, Li. \emph{et al} \cite{dehaze_iccv17} and  Ren. \emph{et al} \cite{dehaze_2016_eccv} method  perform well but they tend to leave haze in the output leading to loss in color contrast. Even though Berman \emph{et al} \cite{dehaze_2016_dana,dehaze_iccp217} perform better, they tend to over-estimate the haze level resulting darker output images.  Overall, our proposed method is able to tackle the problems brought by the other methods and achieve the best performance visually. 

In Fig~\ref{nat_tough}, we present a very tough hazy image to illustrate the results. The visual comparison here also confirms our findings in the previous experiments. Particularly, from the highlighted yellow rectangle, it can be observed that the method can better recover the Mandarin characters hidden behind the haze. 

Through these experiments on real dataset, we are able to demonstrate that the proposed method, although trained on synthetic dataset, is able to generalize well to real world conditions.


\noindent\textbf{Run Time Comparison: } {The proposed method is evaluated for its computational complexity. On average, our method  is able to processes 512$\times$512 images at 18 frames per second (fps), thus providing real-time performance. Further more, the proposed method is compared against several recent methods as shown in Table \ref{tab-time}. The proposed method is comparable to the Li. \emph{et.al}  \cite{dehaze_iccv17} but with better performance.  On average, it takes about 3.3s to de-rain an image of size $512\times 512$. }
\begin{table*}[t]\scriptsize
	\centering
	\caption{Average running time on the synthesized dataset. M: Matlab implementation, P: Python implementation.}
	\label{tab-time}
	\begin{tabular}{|c|c|c|c|c|c|c|}
		\hline
		& He. \emph{et.al} (M) \cite{dehaz_2009_dark} (M)& Zhu. \emph{et.al} \cite{dehaze_fast15} (M) & Ren. \emph{et.al}  \cite{dehaze_2016_eccv} (M)  & Berman. \emph{et.al}  \cite{dehaze_2016_dana}  (M) & Li. \emph{et.al}  \cite{dehaze_iccv17} (P)  & Our (P)
		\\
		\hline                                
		Time (s) & 25.08 & 3.92 & 3.75 & 8.41 & \textbf{3.18} & \textbf{3.33}	
		\\
		\hline
	\end{tabular}
	\vspace{-4mm}
\end{table*}

\subsection{Failure Cases}
\begin{figure}
\centering
	\begin{minipage}{.24\textwidth}
		\centering
		\includegraphics[width=4cm,height=3cm]{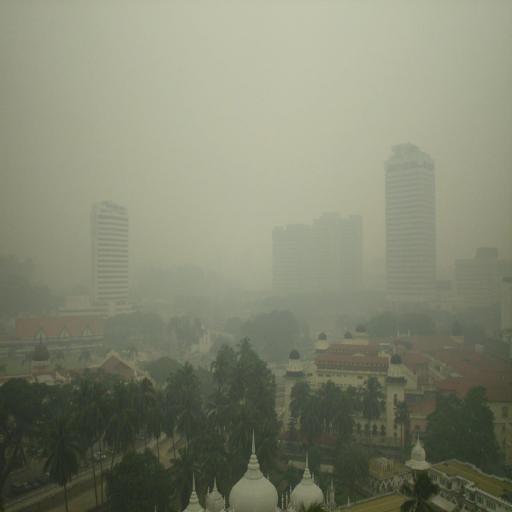}
		\includegraphics[width=4cm,height=3cm]{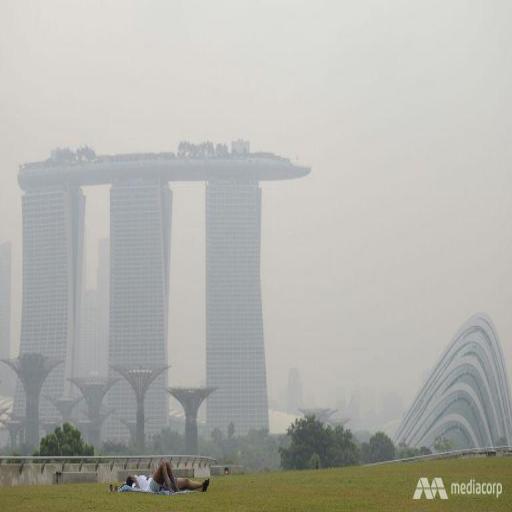}
		\captionsetup{labelformat=empty}
		\captionsetup{justification=centering}
		\caption*{Input image}
	\end{minipage} 
	\begin{minipage}{.24\textwidth}
		\centering
		\includegraphics[width=4cm,height=3cm]{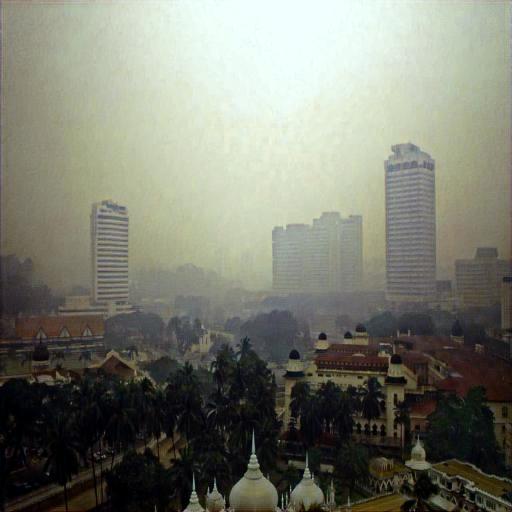}
		\includegraphics[width=4cm,height=3cm]{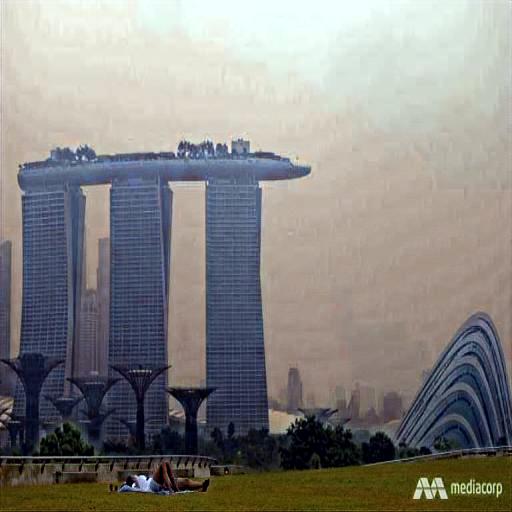}
		\captionsetup{labelformat=empty}
		\captionsetup{justification=centering}
		\caption*{Dehazed image}
	\end{minipage} 
	\caption{Failure case of the proposed method, }\label{faliure}
\end{figure}
{Although the proposed method is able to generalize well to most of the outdoor cases, it results in saturation of certain region of specific images.  For example, as shown in dehazed images in Fig~\ref{faliure}, central part of the sky is not recovered appropriately and it looks over-exposed. This is primarily due to the rarity of similar samples during training. This is a common problem in most existing methods. }

{Though the success of using synthetic samples for avoiding the need of expensive annotations has demonstrated the effectiveness in single image dehazing, the performance gap between the results on synthetic and real-world images illustrates some of the limitations in learning from synthetic data. Hence, it is necessary to  explore new possibilities for leveraging synthetic data in order to obtain better generalization across real world images.}


\section{Conclusion}
This paper presented a new multi-task end-to-end CNN-based network that jointly learns to estimate transmission map and performs image dehazing. In contrast to the existing methods that consider the transmission estimation and single image dehazing as two separate tasks, we bridge the gap between them by using multi-task learning. This is achieved by relaxing the constant atmospheric light assumption in the standard image degradation model. In other words, we enforce the network to estimate the transmission map and use it for further dehazing thereby following the standard image degradation model for image dehazing. Experiments were conducted on multiple datasets (synthetic and real) and the results were compared against several recent methods. Further, detailed ablation studies were conducted to understand the significance of the different components in the proposed. method.




\section*{Acknowledgement} This work was supported by an ARO grant W911NF-16-1-0126.

\bibliographystyle{IEEEtran}
\bibliography{egbib}

\begin{thebibliography}{10}
\providecommand{\url}[1]{#1}
\csname url@samestyle\endcsname
\providecommand{\newblock}{\relax}
\providecommand{\bibinfo}[2]{#2}
\providecommand{\BIBentrySTDinterwordspacing}{\spaceskip=0pt\relax}
\providecommand{\BIBentryALTinterwordstretchfactor}{4}
\providecommand{\BIBentryALTinterwordspacing}{\spaceskip=\fontdimen2\font plus
\BIBentryALTinterwordstretchfactor\fontdimen3\font minus
  \fontdimen4\font\relax}
\providecommand{\BIBforeignlanguage}[2]{{%
\expandafter\ifx\csname l@#1\endcsname\relax
\typeout{** WARNING: IEEEtran.bst: No hyphenation pattern has been}%
\typeout{** loaded for the language `#1'. Using the pattern for}%
\typeout{** the default language instead.}%
\else
\language=\csname l@#1\endcsname
\fi
#2}}
\providecommand{\BIBdecl}{\relax}
\BIBdecl

\bibitem{dehaze_2016_eccv}
W.~Ren, S.~Liu, H.~Zhang, J.~Pan, X.~Cao, and M.-H. Yang, ``Single image
  dehazing via multi-scale convolutional neural networks,'' in
  \emph{ECCV}.\hskip 1em plus 0.5em minus 0.4em\relax Springer, 2016, pp.
  154--169.

\bibitem{dehaze_2016_tip}
B.~Cai, X.~Xu, K.~Jia, C.~Qing, and D.~Tao, ``Dehazenet: An end-to-end system
  for single image haze removal,'' \emph{IEEE TIP}, vol.~25, no.~11, pp.
  5187--5198, 2016.

\bibitem{dehaze_2014_tang}
K.~Tang, J.~Yang, and J.~Wang, ``Investigating haze-relevant features in a
  learning framework for image dehazing,'' in \emph{CVPR}, 2014, pp.
  2995--3000.

\bibitem{dehaze_2016_dana}
D.~Berman, S.~Avidan \emph{et~al.}, ``Non-local image dehazing,'' in
  \emph{CVPR}, 2016, pp. 1674--1682.

\bibitem{dehaz_2009_dark}
K.~He, J.~Sun, and X.~Tang, ``Single image haze removal using dark channel
  prior,'' \emph{IEEE Trans. on PAMI}, vol.~33, no.~12, pp. 2341--2353, 2011.

\bibitem{dehaze_depth}
J.~Kopf, B.~Neubert, B.~Chen, M.~Cohen, D.~Cohen-Or, O.~Deussen,
  M.~Uyttendaele, and D.~Lischinski, ``Deep photo: Model-based photograph
  enhancement and viewing,'' in \emph{ACM TOG}, vol.~27, no.~5.\hskip 1em plus
  0.5em minus 0.4em\relax ACM, 2008, p. 116.

\bibitem{dehaze_multi_image}
Z.~Li, P.~Tan, R.~T. Tan, D.~Zou, S.~Zhiying~Zhou, and L.-F. Cheong,
  ``Simultaneous video defogging and stereo reconstruction,'' in \emph{CVPR},
  2015, pp. 4988--4997.

\bibitem{dehaze_iccv17}
B.~Li, X.~Peng, Z.~Wang, J.~Xu, and D.~Feng, ``An all-in-one network for
  dehazing and beyond,'' \emph{ICCV}, 2017.

\bibitem{yang2018towards}
X.~Yang, Z.~Xu, and J.~Luo, ``Towards perceptual image dehazing by
  physics-based disentanglement and adversarial training,'' 2018.

\bibitem{kim2013optimized}
J.-H. Kim, W.-D. Jang, J.-Y. Sim, and C.-S. Kim, ``Optimized contrast
  enhancement for real-time image and video dehazing,'' \emph{Journal of Visual
  Communication and Image Representation}, vol.~24, no.~3, pp. 410--425, 2013.

\bibitem{galdran2017fusion}
A.~Galdran, J.~Vazquez-Corral, D.~Pardo, and M.~Bertalmio, ``Fusion-based
  variational image dehazing,'' \emph{IEEE Signal Processing Letters}, vol.~24,
  no.~2, pp. 151--155, 2017.

\bibitem{wang2017fast}
W.~Wang, X.~Yuan, X.~Wu, and Y.~Liu, ``Fast image dehazing method based on
  linear transformation,'' \emph{IEEE Transactions on Multimedia}, vol.~19,
  no.~6, pp. 1142--1155, 2017.

\bibitem{dehaze_2018}
H.~Zhang and V.~M. Patel, ``Densely connected pyramid dehazing network,'' in
  \emph{Proceedings of the IEEE Conference on Computer Vision and Pattern
  Recognition}, 2018, pp. 3194--3203.

\bibitem{ren2019deep}
W.~Ren, J.~Zhang, X.~Xu, L.~Ma, X.~Cao, G.~Meng, and W.~Liu, ``Deep video
  dehazing with semantic segmentation,'' \emph{IEEE Transactions on Image
  Processing}, vol.~28, no.~4, pp. 1895--1908, 2019.

\bibitem{zhang2018multi}
H.~Zhang, V.~Sindagi, and V.~M. Patel, ``Multi-scale single image dehazing
  using perceptual pyramid deep network,'' in \emph{Proceedings of the IEEE
  Conference on Computer Vision and Pattern Recognition Workshops}, 2018, pp.
  902--911.

\bibitem{li2018single}
R.~Li, J.~Pan, Z.~Li, and J.~Tang, ``Single image dehazing via conditional
  generative adversarial network,'' in \emph{Proceedings of the IEEE Conference
  on Computer Vision and Pattern Recognition}, 2018, pp. 8202--8211.

\bibitem{yang2018proximal}
D.~Yang and J.~Sun, ``Proximal dehaze-net: A prior learning-based deep network
  for single image dehazing,'' in \emph{Proceedings of the European Conference
  on Computer Vision (ECCV)}, 2018, pp. 702--717.

\bibitem{liu2019dual}
X.~Liu, M.~Suganuma, Z.~Sun, and T.~Okatani, ``Dual residual networks
  leveraging the potential of paired operations for image restoration,''
  \emph{arXiv preprint arXiv:1903.08817}, 2019.

\bibitem{xu2018strong}
Z.~Xu, X.~Yang, X.~Li, X.~Sun, and P.~Harbin, ``Strong baseline for single
  image dehazing with deep features and instance normalization.''

\bibitem{chen2019gated}
D.~Chen, M.~He, Q.~Fan, J.~Liao, L.~Zhang, D.~Hou, L.~Yuan, and G.~Hua, ``Gated
  context aggregation network for image dehazing and deraining,'' in \emph{2019
  IEEE Winter Conference on Applications of Computer Vision (WACV)}.\hskip 1em
  plus 0.5em minus 0.4em\relax IEEE, 2019, pp. 1375--1383.

\bibitem{sakaridis2018semantic}
C.~Sakaridis, D.~Dai, and L.~Van~Gool, ``Semantic foggy scene understanding
  with synthetic data,'' \emph{International Journal of Computer Vision}, pp.
  1--20.

\bibitem{Fattal_2008_siggraph}
\BIBentryALTinterwordspacing
R.~Fattal, ``Single image dehazing,'' in \emph{ACM SIGGRAPH 2008 Papers}, ser.
  SIGGRAPH '08.\hskip 1em plus 0.5em minus 0.4em\relax New York, NY, USA: ACM,
  2008, pp. 72:1--72:9. [Online]. Available:
  \url{http://doi.acm.org/10.1145/1399504.1360671}
\BIBentrySTDinterwordspacing

\bibitem{dehaze_2014acm}
------, ``Dehazing using color-lines,'' vol.~34, no.~13.\hskip 1em plus 0.5em
  minus 0.4em\relax New York, NY, USA: ACM, 2014.

\bibitem{dehaze_iccv15}
Y.~Li, R.~T. Tan, and M.~S. Brown, ``Nighttime haze removal with glow and
  multiple light colors,'' in \emph{Proceedings of the IEEE International
  Conference on Computer Vision}, 2015, pp. 226--234.

\bibitem{dehaze_tif_2012_}
J.-P. Tarel, N.~Hautiere, L.~Caraffa, A.~Cord, H.~Halmaoui, and D.~Gruyer,
  ``Vision enhancement in homogeneous and heterogeneous fog,'' \emph{IEEE
  Intelligent Transportation Systems Magazine}, vol.~4, no.~2, pp. 6--20, 2012.

\bibitem{dehaze_vis_edge}
S.-C. Huang, B.-H. Chen, and W.-J. Wang, ``Visibility restoration of single
  hazy images captured in real-world weather conditions,'' \emph{IEEE
  Transactions on Circuits and Systems for Video Technology}, vol.~24, no.~10,
  pp. 1814--1824, 2014.

\bibitem{dehaze_tan_2008}
R.~T. Tan, ``Visibility in bad weather from a single image,'' in
  \emph{CVPR}.\hskip 1em plus 0.5em minus 0.4em\relax IEEE, 2008, pp. 1--8.

\bibitem{dehaze_2009_kratz}
L.~Kratz and K.~Nishino, ``Factorizing scene albedo and depth from a single
  foggy image,'' in \emph{ICCV}.\hskip 1em plus 0.5em minus 0.4em\relax IEEE,
  2009, pp. 1701--1708.

\bibitem{dehaze_2013_meng}
G.~Meng, Y.~Wang, J.~Duan, S.~Xiang, and C.~Pan, ``Efficient image dehazing
  with boundary constraint and contextual regularization,'' in \emph{ICCV},
  2013, pp. 617--624.

\bibitem{long2015fully}
J.~Long, E.~Shelhamer, and T.~Darrell, ``Fully convolutional networks for
  semantic segmentation,'' in \emph{CVPR}, 2015, pp. 3431--3440.

\bibitem{eigen2015predicting}
D.~Eigen and R.~Fergus, ``Predicting depth, surface normals and semantic labels
  with a common multi-scale convolutional architecture,'' in \emph{ICCV}, 2015,
  pp. 2650--2658.

\bibitem{fu2017clearing}
X.~Fu, J.~Huang, X.~Ding, Y.~Liao, and J.~Paisley, ``Clearing the skies: A deep
  network architecture for single-image rain removal,'' \emph{IEEE Transactions
  on Image Processing}, vol.~26, no.~6, pp. 2944--2956, 2017.

\bibitem{GAN}
I.~Goodfellow, J.~Pouget-Abadie, M.~Mirza, B.~Xu, D.~Warde-Farley, S.~Ozair,
  A.~Courville, and Y.~Bengio, ``Generative adversarial nets,'' in \emph{NIPS},
  2014, pp. 2672--2680.

\bibitem{peng2018jointly}
X.~Peng, Z.~Tang, F.~Yang, R.~Feris, and D.~Metaxas, ``Jointly optimize data
  augmentation and network training: Adversarial data augmentation in human
  pose estimation,'' \emph{arXiv preprint arXiv:1805.09707}, 2018.

\bibitem{zhang2018translating}
Z.~Zhang, L.~Yang, and Y.~Zheng, ``Translating and segmenting multimodal
  medical volumes with cycle-and shape-consistency generative adversarial
  network,'' \emph{arXiv preprint arXiv:1802.09655}, 2018.

\bibitem{pix2pix2016}
P.~Isola, J.-Y. Zhu, T.~Zhou, and A.~A. Efros, ``Image-to-image translation
  with conditional adversarial networks,'' \emph{CVPR}, 2017.

\bibitem{zhang2018synthesis}
H.~Zhang, B.~S. Riggan, S.~Hu, N.~J. Short, and V.~M. Patel, ``Synthesis of
  high-quality visible faces from polarimetric thermal faces using generative
  adversarial networks,'' \emph{International Journal of Computer Vision}, pp.
  1--18.

\bibitem{WEIKAI_GAN}
\BIBentryALTinterwordspacing
R.~Natsume, S.~Saito, Z.~Huang, W.~Chen, C.~Ma, H.~Li, and S.~Morishima,
  ``Siclope: Silhouette-based clothed people,'' \emph{CoRR}, vol.
  abs/1901.00049, 2019. [Online]. Available:
  \url{http://arxiv.org/abs/1901.00049}
\BIBentrySTDinterwordspacing

\bibitem{GAN_sisr}
C.~Ledig, L.~Theis, F.~Husz{\'a}r, J.~Caballero, A.~Cunningham, A.~Acosta,
  A.~Aitken, A.~Tejani, J.~Totz, Z.~Wang \emph{et~al.}, ``Photo-realistic
  single image super-resolution using a generative adversarial network,''
  \emph{arXiv preprint arXiv:1609.04802}, 2016.

\bibitem{yu2018generative}
J.~Yu, Z.~Lin, J.~Yang, X.~Shen, X.~Lu, and T.~S. Huang, ``Generative image
  inpainting with contextual attention,'' in \emph{Proceedings of the IEEE
  Conference on Computer Vision and Pattern Recognition}, 2018, pp. 5505--5514.

\bibitem{yu2018free}
------, ``Free-form image inpainting with gated convolution,'' \emph{arXiv
  preprint arXiv:1806.03589}, 2018.

\bibitem{zhao2018identity}
Y.~Zhao, W.~Chen, J.~Xing, X.~Li, Z.~Bessinger, F.~Liu, W.~Zuo, and R.~Yang,
  ``Identity preserving face completion for large ocular region occlusion,''
  \emph{arXiv preprint arXiv:1807.08772}, 2018.

\bibitem{derain_2017_zhang}
H.~Zhang, V.~Sindagi, and V.~M. Patel, ``Image de-raining using a conditional
  generative adversarial network,'' \emph{arXiv preprint arXiv:1701.05957},
  2017.

\bibitem{nips-generating}
A.~Dosovitskiy and T.~Brox, ``Generating images with perceptual similarity
  metrics based on deep networks,'' in \emph{Advances in Neural Information
  Processing Systems}, 2016, pp. 658--666.

\bibitem{perceptual_loss}
J.~Johnson, A.~Alahi, and L.~Fei-Fei, ``Perceptual losses for real-time style
  transfer and super-resolution,'' in \emph{European Conference on Computer
  Vision}.\hskip 1em plus 0.5em minus 0.4em\relax Springer, 2016, pp. 694--711.

\bibitem{luan2017deep}
F.~Luan, S.~Paris, E.~Shechtman, and K.~Bala, ``Deep photo style transfer.''

\bibitem{yulun_per}
\BIBentryALTinterwordspacing
Y.~Zhang, K.~Li, K.~Li, B.~Zhong, and Y.~Fu, ``Residual non-local attention
  networks for image restoration,'' \emph{CoRR}, vol. abs/1903.10082, 2019.
  [Online]. Available: \url{http://arxiv.org/abs/1903.10082}
\BIBentrySTDinterwordspacing

\bibitem{li2016precomputed}
C.~Li and M.~Wand, ``Precomputed real-time texture synthesis with markovian
  generative adversarial networks,'' in \emph{ECCV}, 2016, pp. 702--716.

\bibitem{gatys2015neural}
L.~A. Gatys, A.~S. Ecker, and M.~Bethge, ``A neural algorithm of artistic
  style,'' \emph{arXiv preprint arXiv:1508.06576}, 2015.

\bibitem{xiong_inpainting}
\BIBentryALTinterwordspacing
W.~Xiong, J.~Yu, Z.~Lin, J.~Yang, X.~Lu, C.~Barnes, and J.~Luo,
  ``Foreground-aware image inpainting,'' \emph{CoRR}, vol. abs/1901.05945,
  2019. [Online]. Available: \url{http://arxiv.org/abs/1901.05945}
\BIBentrySTDinterwordspacing

\bibitem{vgg}
K.~Simonyan and A.~Zisserman, ``Very deep convolutional networks for
  large-scale image recognition,'' \emph{arXiv preprint arXiv:1409.1556}, 2014.

\bibitem{mirza2014conditional}
M.~Mirza and S.~Osindero, ``Conditional generative adversarial nets,''
  \emph{arXiv preprint arXiv:1411.1784}, 2014.

\bibitem{GAN_pix2pix2016}
P.~Isola, J.-Y. Zhu, T.~Zhou, and A.~A. Efros, ``Image-to-image translation
  with conditional adversarial networks,'' \emph{arxiv}, 2016.

\bibitem{unet}
O.~Ronneberger, P.~Fischer, and T.~Brox, ``U-net: Convolutional networks for
  biomedical image segmentation,'' in \emph{International Conference on Medical
  Image Computing and Computer-Assisted Intervention}.\hskip 1em plus 0.5em
  minus 0.4em\relax Springer, 2015, pp. 234--241.

\bibitem{image_res_cnn}
X.-J. Mao, C.~Shen, and Y.-B. Yang, ``Image denoising using very deep fully
  convolutional encoder-decoder networks with symmetric skip connections,''
  \emph{arXiv preprint arXiv:1603.09056}, 2016.

\bibitem{deep_residue}
K.~He, X.~Zhang, S.~Ren, and J.~Sun, ``Deep residual learning for image
  recognition,'' in \emph{Proceedings of the IEEE Conference on Computer Vision
  and Pattern Recognition}, 2016, pp. 770--778.

\bibitem{prelu}
------, ``Delving deep into rectifiers: Surpassing human-level performance on
  imagenet classification,'' in \emph{Proceedings of the IEEE international
  conference on computer vision}, 2015, pp. 1026--1034.

\bibitem{batch_norm}
S.~Ioffe and C.~Szegedy, ``Batch normalization: Accelerating deep network
  training by reducing internal covariate shift,'' in \emph{Proceedings of The
  32nd International Conference on Machine Learning}, 2015, pp. 448--456.

\bibitem{yan2018fast}
C.~Yan, H.~Xie, J.~Chen, Z.~Zha, X.~Hao, Y.~Zhang, and Q.~Dai, ``A fast uyghur
  text detector for complex background images,'' \emph{IEEE Transactions on
  Multimedia}, vol.~20, no.~12, pp. 3389--3398, 2018.

\bibitem{depth_gra_zhou}
B.~Ummenhofer, H.~Zhou, J.~Uhrig, N.~Mayer, E.~Ilg, A.~Dosovitskiy, and
  T.~Brox, ``Demon: Depth and motion network for learning monocular stereo,''
  \emph{CVPR}, 2017.

\bibitem{depth_gra}
J.~Li, R.~Klein, and A.~Yao, ``A two-streamed network for estimating
  fine-scaled depth maps from single rgb images,'' in \emph{The IEEE
  International Conference on Computer Vision (ICCV)}, Oct 2017.

\bibitem{deep_joint_guided}
Y.~Li, J.-B. Huang, N.~Ahuja, and M.-H. Yang, ``Deep joint image filtering,''
  in \emph{European Conference on Computer Vision}.\hskip 1em plus 0.5em minus
  0.4em\relax Springer, 2016, pp. 154--169.

\bibitem{guided_1}
X.~Shen, C.~Zhou, L.~Xu, and J.~Jia, ``Mutual-structure for joint filtering,''
  in \emph{ICCV}, 2015, pp. 3406--3414.

\bibitem{guided_2}
D.~Ferstl, C.~Reinbacher, R.~Ranftl, M.~R{\"u}ther, and H.~Bischof, ``Image
  guided depth upsampling using anisotropic total generalized variation,'' in
  \emph{ICCV}, 2013, pp. 993--1000.

\bibitem{yan2018supervised}
C.~Yan, H.~Xie, D.~Yang, J.~Yin, Y.~Zhang, and Q.~Dai, ``Supervised hash coding
  with deep neural network for environment perception of intelligent
  vehicles,'' \emph{IEEE transactions on intelligent transportation systems},
  vol.~19, no.~1, pp. 284--295, 2018.

\bibitem{yan2018effective}
C.~Yan, H.~Xie, S.~Liu, J.~Yin, Y.~Zhang, and Q.~Dai, ``Effective uyghur
  language text detection in complex background images for traffic prompt
  identification,'' \emph{IEEE transactions on intelligent transportation
  systems}, vol.~19, no.~1, pp. 220--229, 2018.

\bibitem{nyudepth}
P.~K. Nathan~Silberman, Derek~Hoiem and R.~Fergus, ``Indoor segmentation and
  support inference from rgbd images,'' in \emph{ECCV}, 2012.

\bibitem{torch}
R.~Collobert, K.~Kavukcuoglu, and C.~Farabet, ``Torch7: A matlab-like
  environment for machine learning,'' in \emph{BigLearn, NIPS Workshop}, 2011.

\bibitem{adam_opt}
D.~Kingma and J.~Ba, ``Adam: A method for stochastic optimization,''
  \emph{arXiv preprint arXiv:1412.6980}, 2014.

\bibitem{dehaze_fast15}
Q.~Zhu, J.~Mai, and L.~Shao, ``A fast single image haze removal algorithm using
  color attenuation prior,'' \emph{IEEE Transactions on Image Processing},
  vol.~24, no.~11, pp. 3522--3533, 2015.

\bibitem{dehaze_iccp217}
D.~Berman, T.~Treibitz, and S.~Avidan, ``Air-light estimation using
  haze-lines,'' in \emph{Computational Photography (ICCP), 2017 IEEE
  International Conference on}.\hskip 1em plus 0.5em minus 0.4em\relax IEEE,
  2017, pp. 1--9.

\bibitem{ssim}
Z.~Wang, A.~C. Bovik, H.~R. Sheikh, and E.~P. Simoncelli, ``Image quality
  assessment: from error visibility to structural similarity,'' \emph{IEEE
  TIP}, vol.~13, no.~4, pp. 600--612, 2004.

\end{thebibliography}


@article{gharbi2016deep,
  title={Deep joint demosaicking and denoising},
  author={Gharbi, Micha{\"e}l and Chaurasia, Gaurav and Paris, Sylvain and Durand, Fr{\'e}do},
  journal={ACM Transactions on Graphics (TOG)},
  volume={35},
  number={6},
  pages={191},
  year={2016},
  publisher={ACM}
}


@article{sakaridis2018semantic,
	title={Semantic foggy scene understanding with synthetic data},
	author={Sakaridis, Christos and Dai, Dengxin and Van Gool, Luc},
	journal={International Journal of Computer Vision},
	pages={1--20},
	publisher={Springer}
}

@article{liu2019dual,
	title={Dual Residual Networks Leveraging the Potential of Paired Operations for Image Restoration},
	author={Liu, Xing and Suganuma, Masanori and Sun, Zhun and Okatani, Takayuki},
	journal={arXiv preprint arXiv:1903.08817},
	year={2019}
}
×
Drag and Drop
The image will be downloaded by Fatkun


@inproceedings{chen2019gated,
  title={Gated Context Aggregation Network for Image Dehazing and Deraining},
  author={Chen, Dongdong and He, Mingming and Fan, Qingnan and Liao, Jing and Zhang, Liheng and Hou, Dongdong and Yuan, Lu and Hua, Gang},
  booktitle={2019 IEEE Winter Conference on Applications of Computer Vision (WACV)},
  pages={1375--1383},
  year={2019},
  organization={IEEE}
}
×
Drag and Drop
The image will be downloaded by Fatkun


@inproceedings{xu2018strong,
	title={Strong baseline for single image dehazing with deep features and instance normalization},
	author={Xu, Zheng and Yang, Xitong and Li, Xue and Sun, Xiaoshuai and Harbin, PR}
}
×
Drag and Drop
The image will be downloaded by Fatkun

@article{fu2017clearing,
	title={Clearing the skies: A deep network architecture for single-image rain removal},
	author={Fu, Xueyang and Huang, Jiabin and Ding, Xinghao and Liao, Yinghao and Paisley, John},
	journal={IEEE Transactions on Image Processing},
	volume={26},
	number={6},
	pages={2944--2956},
	year={2017},
	publisher={IEEE}
}
×
Drag and Drop
The image will be downloaded by Fatkun

@article{yulun_per,
	author    = {Yulun Zhang and
	Kunpeng Li and
	Kai Li and
	Bineng Zhong and
	Yun Fu},
	title     = {Residual Non-local Attention Networks for Image Restoration},
	journal   = {CoRR},
	volume    = {abs/1903.10082},
	year      = {2019},
	url       = {http://arxiv.org/abs/1903.10082},
	archivePrefix = {arXiv},
}

@inproceedings{yu2018generative,
title={Generative image inpainting with contextual attention},
author={Yu, Jiahui and Lin, Zhe and Yang, Jimei and Shen, Xiaohui and Lu, Xin and Huang, Thomas S},
booktitle={Proceedings of the IEEE Conference on Computer Vision and Pattern Recognition},
pages={5505--5514},
year={2018}
}
×
Drag and Drop
The image will be downloaded by Fatkun


@article{zhang2018synthesis,
title={Synthesis of High-Quality Visible Faces from Polarimetric Thermal Faces using Generative Adversarial Networks},
author={Zhang, He and Riggan, Benjamin S and Hu, Shuowen and Short, Nathaniel J and Patel, Vishal M},
journal={International Journal of Computer Vision},
pages={1--18},
publisher={Springer}
}


@article{yamaguchi2018high,
	title={High-fidelity facial reflectance and geometry inference from an unconstrained image},
	author={Yamaguchi, Shuco and Saito, Shunsuke and Nagano, Koki and Zhao, Yajie and Chen, Weikai and Olszewski, Kyle and Morishima, Shigeo and Li, Hao},
	journal={ACM Transactions on Graphics (TOG)},
	volume={37},
	number={4},
	pages={162},
	year={2018},
	publisher={ACM}
}
×
Drag and Drop
The image will be downloaded by Fatkun

@article{zhao2018identity,
	title={Identity preserving face completion for large ocular region occlusion},
	author={Zhao, Yajie and Chen, Weikai and Xing, Jun and Li, Xiaoming and Bessinger, Zach and Liu, Fuchang and Zuo, Wangmeng and Yang, Ruigang},
	journal={arXiv preprint arXiv:1807.08772},
	year={2018}
}
×
Drag and Drop
The image will be downloaded by Fatkun

@article{WEIKAI_GAN,
	author    = {Ryota Natsume and
	Shunsuke Saito and
	Zeng Huang and
	Weikai Chen and
	Chongyang Ma and
	Hao Li and
	Shigeo Morishima},
	title     = {SiCloPe: Silhouette-Based Clothed People},
	journal   = {CoRR},
	volume    = {abs/1901.00049},
	year      = {2019},
	url       = {http://arxiv.org/abs/1901.00049},
	archivePrefix = {arXiv},
	eprint    = {1901.00049},
	timestamp = {Thu, 31 Jan 2019 13:52:49 +0100},
	biburl    = {https://dblp.org/rec/bib/journals/corr/abs-1901-00049},
	bibsource = {dblp computer science bibliography, https://dblp.org}
}

@article{xiong_inpainting,
	author    = {Wei Xiong and
	Jiahui Yu and
	Zhe Lin and
	Jimei Yang and
	Xin Lu and
	Connelly Barnes and
	Jiebo Luo},
	title     = {Foreground-aware Image Inpainting},
	journal   = {CoRR},
	volume    = {abs/1901.05945},
	year      = {2019},
	url       = {http://arxiv.org/abs/1901.05945},
	archivePrefix = {arXiv},
	eprint    = {1901.05945},
	timestamp = {Fri, 01 Feb 2019 13:39:59 +0100},
	biburl    = {https://dblp.org/rec/bib/journals/corr/abs-1901-05945},
	bibsource = {dblp computer science bibliography, https://dblp.org}
}


@article{yu2018free,
title={Free-form image inpainting with gated convolution},
author={Yu, Jiahui and Lin, Zhe and Yang, Jimei and Shen, Xiaohui and Lu, Xin and Huang, Thomas S},
journal={arXiv preprint arXiv:1806.03589},
year={2018}
}

@article{li2019benchmarking,
	title={Benchmarking single-image dehazing and beyond},
	author={Li, Boyi and Ren, Wenqi and Fu, Dengpan and Tao, Dacheng and Feng, Dan and Zeng, Wenjun and Wang, Zhangyang},
	journal={IEEE Transactions on Image Processing},
	volume={28},
	number={1},
	pages={492--505},
	year={2019},
	publisher={IEEE}
}
×
Drag and Drop
The image will be downloaded by Fatkun

@inproceedings{yang2018proximal,
	title={Proximal Dehaze-Net: A Prior Learning-Based Deep Network for Single Image Dehazing},
	author={Yang, Dong and Sun, Jian},
	booktitle={Proceedings of the European Conference on Computer Vision (ECCV)},
	pages={702--717},
	year={2018}
}
×
Drag and Drop
The image will be downloaded by Fatkun

@inproceedings{li2018single,
	title={Single image dehazing via conditional generative adversarial network},
	author={Li, Runde and Pan, Jinshan and Li, Zechao and Tang, Jinhui},
	booktitle={Proceedings of the IEEE Conference on Computer Vision and Pattern Recognition},
	pages={8202--8211},
	year={2018}
}
×
Drag and Drop
The image will be downloaded by Fatkun
@inproceedings{zhang2018multi,
	title={Multi-scale single image dehazing using perceptual pyramid deep network},
	author={Zhang, He and Sindagi, Vishwanath and Patel, Vishal M},
	booktitle={Proceedings of the IEEE Conference on Computer Vision and Pattern Recognition Workshops},
	pages={902--911},
	year={2018}
}
×
Drag and Drop
The image will be downloaded by Fatkun


@article{ren2019deep,
	title={Deep video dehazing with semantic segmentation},
	author={Ren, Wenqi and Zhang, Jingang and Xu, Xiangyu and Ma, Lin and Cao, Xiaochun and Meng, Gaofeng and Liu, Wei},
	journal={IEEE Transactions on Image Processing},
	volume={28},
	number={4},
	pages={1895--1908},
	year={2019},
	publisher={IEEE}
}
×
Drag and Drop
The image will be downloaded by Fatkun
@article{xing_face,
	author    = {Xing Di and He Zhang and Vishal M. Patel},
	title     = {Polarimetric Thermal to Visible Face Verification via Attribute Preserved
	Synthesis},
	journal   = {CoRR},
	volume    = {abs/1901.00889},
	year      = {2019},
	url       = {http://arxiv.org/abs/1901.00889},
	archivePrefix = {arXiv},
}

@inproceedings{dehaze_2018,
	title={Densely connected pyramid dehazing network},
	author={Zhang, He and Patel, Vishal M},
	booktitle={Proceedings of the IEEE Conference on Computer Vision and Pattern Recognition},
	pages={3194--3203},
	year={2018}
}


@article{yan2018fast,
	title={A fast Uyghur text detector for complex background images},
	author={Yan, Chenggang and Xie, Hongtao and Chen, Jianjun and Zha, Zhengjun and Hao, Xinhong and Zhang, Yongdong and Dai, Qionghai},
	journal={IEEE Transactions on Multimedia},
	volume={20},
	number={12},
	pages={3389--3398},
	year={2018},
	publisher={IEEE}
}
×
Drag and Drop
The image will be downloaded by Fatkun


@article{li2019benchmarking,
	title={Benchmarking single-image dehazing and beyond},
	author={Li, Boyi and Ren, Wenqi and Fu, Dengpan and Tao, Dacheng and Feng, Dan and Zeng, Wenjun and Wang, Zhangyang},
	journal={IEEE Transactions on Image Processing},
	volume={28},
	number={1},
	pages={492--505},
	year={2019},
	publisher={IEEE}
}
@article{yan2018effective,
	title={Effective Uyghur language text detection in complex background images for traffic prompt identification},
	author={Yan, Chenggang and Xie, Hongtao and Liu, Shun and Yin, Jian and Zhang, Yongdong and Dai, Qionghai},
	journal={IEEE transactions on intelligent transportation systems},
	volume={19},
	number={1},
	pages={220--229},
	year={2018},
	publisher={IEEE}
}

@article{yan2018supervised,
	title={Supervised hash coding with deep neural network for environment perception of intelligent vehicles},
	author={Yan, Chenggang and Xie, Hongtao and Yang, Dongbao and Yin, Jian and Zhang, Yongdong and Dai, Qionghai},
	journal={IEEE transactions on intelligent transportation systems},
	volume={19},
	number={1},
	pages={284--295},
	year={2018},
	publisher={IEEE}
}

@article{wang2017fast,
	title={Fast image dehazing method based on linear transformation},
	author={Wang, Wencheng and Yuan, Xiaohui and Wu, Xiaojin and Liu, Yunlong},
	journal={IEEE Transactions on Multimedia},
	volume={19},
	number={6},
	pages={1142--1155},
	year={2017},
	publisher={IEEE}
}

@article{galdran2017fusion,
	title={Fusion-based variational image dehazing},
	author={Galdran, Adrian and Vazquez-Corral, Javier and Pardo, David and Bertalmio, Marcelo},
	journal={IEEE Signal Processing Letters},
	volume={24},
	number={2},
	pages={151--155},
	year={2017},
	publisher={IEEE}
}

@article{kim2013optimized,
	title={Optimized contrast enhancement for real-time image and video dehazing},
	author={Kim, Jin-Hwan and Jang, Won-Dong and Sim, Jae-Young and Kim, Chang-Su},
	journal={Journal of Visual Communication and Image Representation},
	volume={24},
	number={3},
	pages={410--425},
	year={2013},
	publisher={Elsevier}
}

@inproceedings{yang2018towards,
	title={Towards perceptual image dehazing by physics-based disentanglement and adversarial training},
	author={Yang, Xitong and Xu, Zheng and Luo, Jiebo},
	year={2018}
}

@article{zhang2018translating,
	title={Translating and Segmenting Multimodal Medical Volumes with Cycle-and Shape-Consistency Generative Adversarial Network},
	author={Zhang, Zizhao and Yang, Lin and Zheng, Yefeng},
	journal={arXiv preprint arXiv:1802.09655},
	year={2018}
}

@article{peng2018jointly,
	title={Jointly Optimize Data Augmentation and Network Training: Adversarial Data Augmentation in Human Pose Estimation},
	author={Peng, Xi and Tang, Zhiqiang and Yang, Fei and Feris, Rogerio and Metaxas, Dimitris},
	journal={arXiv preprint arXiv:1805.09707},
	year={2018}
}

@article{middlebury_dataset,
  title={A taxonomy and evaluation of dense two-frame stereo correspondence algorithms},
  author={Scharstein, Daniel and Szeliski, Richard},
  journal={International journal of computer vision},
  volume={47},
  number={1-3},
  pages={7--42},
  year={2002},
  publisher={Springer}
}


@inproceedings{stage_wise_recog,
  title={Multimodal deep learning for robust RGB-D object recognition},
  author={Eitel, Andreas and Springenberg, Jost Tobias and Spinello, Luciano and Riedmiller, Martin and Burgard, Wolfram},
  booktitle={Intelligent Robots and Systems (IROS), 2015 IEEE/RSJ International Conference on},
  pages={681--687},
  year={2015},
  organization={IEEE}
}

@inproceedings{stage_wise_feature,
  title={Stage-wise training: An improved feature learning strategy for deep models},
  author={Barshan, Elnaz and Fieguth, Paul},
  booktitle={Feature Extraction: Modern Questions and Challenges},
  pages={49--59},
  year={2015}
}


@article{stage_wise_nvidia,
  title={Progressive Growing of GANs for Improved Quality, Stability, and Variation},
  author={Karras, Tero and Aila, Timo and Laine, Samuli and Lehtinen, Jaakko},
  journal={arXiv preprint arXiv:1710.10196},
  year={2017}
}


@inproceedings{cnn_vis,
  title={Visualizing and understanding convolutional networks},
  author={Zeiler, Matthew D and Fergus, Rob},
  booktitle={European conference on computer vision},
  pages={818--833},
  year={2014},
  organization={Springer}
}


@article{dehaze_AAAI,
  author    = {Boyi Li and
               Xiulian Peng and
               Zhangyang Wang and
               Jizheng Xu and
               Dan Feng},
  title     = {End-to-End United Video Dehazing and Detection},
  journal   = {CoRR},
  volume    = {abs/1709.03919},
  year      = {2017},
  url       = {http://arxiv.org/abs/1709.03919},
  archivePrefix = {arXiv},
  eprint    = {1709.03919},
  timestamp = {Thu, 05 Oct 2017 09:42:58 +0200},
}
@article{dehaze_2008,
  title={Single image dehazing},
  author={Fattal, Raanan},
  journal={ACM transactions on graphics (TOG)},
  volume={27},
  number={3},
  pages={72},
  year={2008},
  publisher={ACM}
}

@article{gan_tutorial,
  title={NIPS 2016 tutorial: Generative adversarial networks},
  author={Goodfellow, Ian},
  journal={arXiv preprint arXiv:1701.00160},
  year={2016}
}


@article{luan2017deep,
  title={Deep photo style transfer},
  author={Luan, Fujun and Paris, Sylvain and Shechtman, Eli and Bala, Kavita}
}




@article{yan2016automatic,
	  title={Automatic photo adjustment using deep neural networks},
	  author={Yan, Zhicheng and Zhang, Hao and Wang, Baoyuan and Paris, Sylvain and Yu, Yizhou},
	  journal={ACM Transactions on Graphics (TOG)},
	  volume={35},
	  number={2},
	  pages={11},
	  year={2016},
	  publisher={ACM}
	}

@InProceedings{depth_gra,
author = {Li, Jun and Klein, Reinhard and Yao, Angela},
title = {A Two-Streamed Network for Estimating Fine-Scaled Depth Maps From Single RGB Images},
booktitle = {The IEEE International Conference on Computer Vision (ICCV)},
month = {Oct},
year = {2017}
}}


@inproceedings{dense_fully,
  title={The one hundred layers tiramisu: Fully convolutional densenets for semantic segmentation},
  author={J{\'e}gou, Simon and Drozdzal, Michal and Vazquez, David and Romero, Adriana and Bengio, Yoshua},
  booktitle={Computer Vision and Pattern Recognition Workshops (CVPRW), 2017 IEEE Conference on},
  pages={1175--1183},
  year={2017},
  organization={IEEE}
}



@article{gan_humanpose,
  title={Adversarial PoseNet: A Structure-aware Convolutional Network for Human Pose Estimation},
  author={Chen, Yu and Shen, Chunhua and Wei, Xiu-Shen and Liu, Lingqiao and Yang, Jian},
  journal={ICCV},
  year={2017}
}




@article{pix2pix2016,
  title={Image-to-Image Translation with Conditional Adversarial Networks},
  author={Isola, Phillip and Zhu, Jun-Yan and Zhou, Tinghui and Efros, Alexei A},
  journal={CVPR},
  year={2017}
}


@inproceedings{zhang2018multi,
	title={Multi-style generative network for real-time transfer},
	author={Zhang, Hang and Dana, Kristin},
	booktitle={Proceedings of the European Conference on Computer Vision (ECCV)},
	pages={0--0},
	year={2018}
}
×
Drag and Drop
The image will be downloaded by Fatkun

@article{Spatial_pymaid,
  title={Spatial pyramid pooling in deep convolutional networks for visual recognition},
  author={He, Kaiming and Zhang, Xiangyu and Ren, Shaoqing and Sun, Jian},
  journal={IEEE transactions on pattern analysis and machine intelligence},
  volume={37},
  number={9},
  pages={1904--1916},
  year={2015},
  publisher={IEEE}
}

@article{gan_conditioanl,
  title={Conditional generative adversarial nets},
  author={Mirza, Mehdi and Osindero, Simon},
  journal={arXiv preprint arXiv:1411.1784},
  year={2014}
}

@article{gan_stack_zhang,
  title={Stackgan: Text to photo-realistic image synthesis with stacked generative adversarial networks},
  author={Zhang, Han and Xu, Tao and Li, Hongsheng and Zhang, Shaoting and Huang, Xiaolei and Wang, Xiaogang and Metaxas, Dimitris},
  journal={ICCV},
  year={2017}
}


@inproceedings{gan_domain_liu,
  title={Coupled generative adversarial networks},
  author={Liu, Ming-Yu and Tuzel, Oncel},
  booktitle={NIPS	},
  pages={469--477},
  year={2016}
}

@inproceedings{dehaze_iccp14_light,
  title={Automatic recovery of the atmospheric light in hazy images},
  author={Sulami, Matan and Glatzer, Itamar and Fattal, Raanan and Werman, Mike},
  booktitle={Computational Photography (ICCP), 2014 IEEE International Conference on},
  pages={1--11},
  year={2014},
  organization={IEEE}
}


@inproceedings{dehaze_iccp217,
  title={Air-light estimation using haze-lines},
  author={Berman, Dana and Treibitz, Tali and Avidan, Shai},
  booktitle={Computational Photography (ICCP), 2017 IEEE International Conference on},
  pages={1--9},
  year={2017},
  organization={IEEE}
}

@inproceedings{depth_middlebbury,
  title={High-resolution stereo datasets with subpixel-accurate ground truth},
  author={Scharstein, Daniel and Hirschm{\"u}ller, Heiko and Kitajima, York and Krathwohl, Greg and Ne{\v{s}}i{\'c}, Nera and Wang, Xi and Westling, Porter},
  booktitle={German Conference on Pattern Recognition},
  pages={31--42},
  year={2014},
  organization={Springer}
}

@inproceedings{sunny_3d_depth,
  title={Sun rgb-d: A rgb-d scene understanding benchmark suite},
  author={Song, Shuran and Lichtenberg, Samuel P and Xiao, Jianxiong},
  booktitle={Proceedings of the IEEE conference on computer vision and pattern recognition},
  pages={567--576},
  year={2015}
}


@article{dehaze_2017_joint,
  title={Joint Transmission Map Estimation and Dehazing using Deep Networks},
  author={Zhang, He and Sindagi, Vishwanath and Patel, Vishal M},
  journal={arXiv preprint arXiv:1708.00581},
  year={2017}
}
@article{depth_gra_zhou,
  title={DeMoN: Depth and motion network for learning monocular stereo},
  author={Ummenhofer, Benjamin and Zhou, Huizhong and Uhrig, Jonas and Mayer, Nikolaus and Ilg, Eddy and Dosovitskiy, Alexey and Brox, Thomas},
  journal={CVPR},
  year={2017}
}

@article{dehaze_vis_edge,
  title={Visibility restoration of single hazy images captured in real-world weather conditions},
  author={Huang, Shih-Chia and Chen, Bo-Hao and Wang, Wei-Jheng},
  journal={IEEE Transactions on Circuits and Systems for Video Technology},
  volume={24},
  number={10},
  pages={1814--1824},
  year={2014},
  publisher={IEEE}
}

@article{dehaze_fast15,
  title={A fast single image haze removal algorithm using color attenuation prior},
  author={Zhu, Qingsong and Mai, Jiaming and Shao, Ling},
  journal={IEEE Transactions on Image Processing},
  volume={24},
  number={11},
  pages={3522--3533},
  year={2015},
  publisher={IEEE}
}

@article{dense_net,
  title={Densely connected convolutional networks},
  author={Huang, Gao and Liu, Zhuang and Weinberger, Kilian Q and van der Maaten, Laurens},
  journal={CVPR},
  year={2017}
}
@article{dehaze_iccv17,
  title={An All-in-One Network for Dehazing and Beyond},
  author={Li, Boyi and Peng, Xiulian and Wang, Zhangyang and Xu, Jizheng and Feng, Dan},
  journal={ICCV},
  year={2017}
}

@article{psp_net,
  title={Pyramid scene parsing network},
  author={Zhao, Hengshuang and Shi, Jianping and Qi, Xiaojuan and Wang, Xiaogang and Jia, Jiaya},
  journal={arXiv preprint arXiv:1612.01105},
  year={2016}
}

@article{dehaze_mul_fusion,
  title={Single image dehazing by multi-scale fusion},
  author={Ancuti, Codruta Orniana and Ancuti, Cosmin},
  journal={IEEE Transactions on Image Processing},
  volume={22},
  number={8},
  pages={3271--3282},
  year={2013},
  publisher={IEEE}
}

@inproceedings{prelu,
  title={Delving deep into rectifiers: Surpassing human-level performance on imagenet classification},
  author={He, Kaiming and Zhang, Xiangyu and Ren, Shaoqing and Sun, Jian},
  booktitle={Proceedings of the IEEE international conference on computer vision},
  pages={1026--1034},
  year={2015}
}

@inproceedings{batch_norm,
  title={Batch Normalization: Accelerating Deep Network Training by Reducing Internal Covariate Shift},
  author={Ioffe, Sergey and Szegedy, Christian},
  booktitle={Proceedings of The 32nd International Conference on Machine Learning},
  pages={448--456},
  year={2015}
}
@article{dehaze_tif_2012_,
  title={Vision enhancement in homogeneous and heterogeneous fog},
  author={Tarel, Jean-Philippe and Hautiere, Nicholas and Caraffa, Laurent and Cord, Aur{\'e}lien and Halmaoui, Houssam and Gruyer, Dominique},
  journal={IEEE Intelligent Transportation Systems Magazine},
  volume={4},
  number={2},
  pages={6--20},
  year={2012},
  publisher={IEEE}
}

@inproceedings{dehaze_iccv15,
  title={Nighttime haze removal with glow and multiple light colors},
  author={Li, Yu and Tan, Robby T and Brown, Michael S},
  booktitle={Proceedings of the IEEE International Conference on Computer Vision},
  pages={226--234},
  year={2015}
}

@inproceedings{dehaze_depth,
  title={Deep photo: Model-based photograph enhancement and viewing},
  author={Kopf, Johannes and Neubert, Boris and Chen, Billy and Cohen, Michael and Cohen-Or, Daniel and Deussen, Oliver and Uyttendaele, Matt and Lischinski, Dani},
  booktitle={ACM TOG},
  volume={27},
  number={5},
  pages={116},
  year={2008},
  organization={ACM}
}

@article{dehaze_multiple,
  title={Contrast restoration of weather degraded images},
  author={Narasimhan, Srinivasa G. and Nayar, Shree K.},
  journal={IEEE Trans on PAMI},
  volume={25},
  number={6},
  pages={713--724},
  year={2003},
  publisher={IEEE}
}

@inproceedings{guided_2,
  title={Image guided depth upsampling using anisotropic total generalized variation},
  author={Ferstl, David and Reinbacher, Christian and Ranftl, Rene and R{\"u}ther, Matthias and Bischof, Horst},
  booktitle={ICCV},
  pages={993--1000},
  year={2013}
}


@inproceedings{guided_1,
  title={Mutual-structure for joint filtering},
  author={Shen, Xiaoyong and Zhou, Chao and Xu, Li and Jia, Jiaya},
  booktitle={ICCV},
  pages={3406--3414},
  year={2015}
}

@inproceedings{deep_joint_guided,
  title={Deep joint image filtering},
  author={Li, Yijun and Huang, Jia-Bin and Ahuja, Narendra and Yang, Ming-Hsuan},
  booktitle={European Conference on Computer Vision},
  pages={154--169},
  year={2016},
  organization={Springer}
}

@inproceedings{nyudepth,
  author    = {Nathan Silberman, Derek Hoiem, Pushmeet Kohli and Rob Fergus},
  title     = {Indoor Segmentation and Support Inference from RGBD Images},
  booktitle = {ECCV},
  year      = {2012}
}


@inproceedings{dehaze_2014acm,   
author = {Fattal, Raanan},   
title = {Dehazing using Color-Lines},
journal = {ACM Transaction on Graphics},   
volume = {34},
issue = {1}, 
number = {13},   
year = {2014},  
publisher = {ACM},   
address = {New York, NY, USA},   
}  


@inproceedings{Fattal_2008_siggraph,
 author = {Fattal, Raanan},
 title = {Single Image Dehazing},
 booktitle = {ACM SIGGRAPH 2008 Papers},
 series = {SIGGRAPH '08},
 year = {2008},
 isbn = {978-1-4503-0112-1},
 location = {Los Angeles, California},
 pages = {72:1--72:9},
 articleno = {72},
 numpages = {9},
 url = {http://doi.acm.org/10.1145/1399504.1360671},
 doi = {10.1145/1399504.1360671},
 acmid = {1360671},
 publisher = {ACM},
 address = {New York, NY, USA}
} 


@inproceedings{dehaze_known_depth,
  title={Deep photo: Model-based photograph enhancement and viewing},
  author={Kopf, Johannes and Neubert, Boris and Chen, Billy and Cohen, Michael and Cohen-Or, Daniel and Deussen, Oliver and Uyttendaele, Matt and Lischinski, Dani},
  booktitle={ACM TOG},
  volume={27},
  number={5},
  pages={116},
  year={2008},
  organization={ACM}
}


@inproceedings{dehaze_multi_image,
  title={Simultaneous video defogging and stereo reconstruction},
  author={Li, Zhuwen and Tan, Ping and Tan, Robby T and Zou, Danping and Zhiying Zhou, Steven and Cheong, Loong-Fah},
  booktitle={CVPR},
  pages={4988--4997},
  year={2015}
}


@inproceedings{unet,
  title={U-net: Convolutional networks for biomedical image segmentation},
  author={Ronneberger, Olaf and Fischer, Philipp and Brox, Thomas},
  booktitle={International Conference on Medical Image Computing and Computer-Assisted Intervention},
  pages={234--241},
  year={2015},
  organization={Springer}
}

@inproceedings{depth_nips_14,
  title={Depth map prediction from a single image using a multi-scale deep network},
  author={Eigen, David and Puhrsch, Christian and Fergus, Rob},
  booktitle={NIPS},
  pages={2366--2374},
  year={2014}
}

@article{dehaze_survey,
  author    = {Yu Li and
               Shaodi You and
               Michael S. Brown and
               Robby T. Tan},
  title     = {Haze Visibility Enhancement: {A} Survey and Quantitative Benchmarking},
  journal   = {CoRR},
  volume    = {abs/1607.06235},
  year      = {2016},
  url       = {http://arxiv.org/abs/1607.06235},
  timestamp = {Tue, 02 Aug 2016 12:59:27 +0200},
  biburl    = {http://dblp.uni-trier.de/rec/bib/journals/corr/LiYBT16},
  bibsource = {dblp computer science bibliography, http://dblp.org}
}

@inproceedings{dehaze_2016_eccv,
  title={Single image dehazing via multi-scale convolutional neural networks},
  author={Ren, Wenqi and Liu, Si and Zhang, Hua and Pan, Jinshan and Cao, Xiaochun and Yang, Ming-Hsuan},
  booktitle={ECCV},
  pages={154--169},
  year={2016},
  organization={Springer}
}



@article{dehaze_2016_tip,
  title={Dehazenet: An end-to-end system for single image haze removal},
  author={Cai, Bolun and Xu, Xiangmin and Jia, Kui and Qing, Chunmei and Tao, Dacheng},
  journal={IEEE TIP},
  volume={25},
  number={11},
  pages={5187--5198},
  year={2016},
  publisher={IEEE}
}


@inproceedings{dehaze_2016_dana,
  title={Non-local image dehazing},
  author={Berman, Dana and Avidan, Shai and others},
  booktitle={CVPR},
  pages={1674--1682},
  year={2016}
}


@article{derain_2017_zhang,
  title={Image De-raining Using a Conditional Generative Adversarial Network},
  author={Zhang, He and Sindagi, Vishwanath and Patel, Vishal M},
  journal={arXiv preprint arXiv:1701.05957},
  year={2017}
} 


@inproceedings{dehaze_2014_tang,
  title={Investigating haze-relevant features in a learning framework for image dehazing},
  author={Tang, Ketan and Yang, Jianchao and Wang, Jue},
  booktitle={CVPR},
  pages={2995--3000},
  year={2014}
}

@inproceedings{dehaze_2013_meng,
  title={Efficient image dehazing with boundary constraint and contextual regularization},
  author={Meng, Gaofeng and Wang, Ying and Duan, Jiangyong and Xiang, Shiming and Pan, Chunhong},
  booktitle={ICCV},
  pages={617--624},
  year={2013}
}

@inproceedings{dehaze_2009_kratz,
  title={Factorizing scene albedo and depth from a single foggy image},
  author={Kratz, Louis and Nishino, Ko},
  booktitle={ICCV},
  pages={1701--1708},
  year={2009},
  organization={IEEE}
}


@inproceedings{dehaze_tan_2008,
  title={Visibility in bad weather from a single image},
  author={Tan, Robby T},
  booktitle={CVPR},
  pages={1--8},
  year={2008},
  organization={IEEE}
}


@inproceedings{nips-generating,
  title={Generating Images with Perceptual Similarity Metrics based on Deep Networks},
  author={Dosovitskiy, Alexey and Brox, Thomas},
  booktitle={Advances in Neural Information Processing Systems},
  pages={658--666},
  year={2016}
}

@inproceedings{rain_wacv2017,
	title={Convolutional Sparse and Low-Rank Coding-Based Rain Streak Removal},
	author={Zhang, He and Patel, Vishal M },
	booktitle={2017 IEEE WACV},
	pages={1--9},
	year={2017},
	organization={IEEE}
}

@article{vgg,
  title={Very deep convolutional networks for large-scale image recognition},
  author={Simonyan, Karen and Zisserman, Andrew},
  journal={arXiv preprint arXiv:1409.1556},
  year={2014}
}



@inproceedings{image_sisr_acc,
  title={Accelerating the super-resolution convolutional neural network},
  author={Dong, Chao and Loy, Chen Change and Tang, Xiaoou},
  booktitle={ECCV},
  pages={391--407},
  year={2016},
  organization={Springer}
}


@article{rain_haze_joint,
  author    = {Wenhan Yang and
               Robby T. Tan and
               Jiashi Feng and
               Jiaying Liu and
               Zongming Guo and
               Shuicheng Yan},
  title     = {Joint Rain Detection and Removal via Iterative Region Dependent Multi-Task
               Learning},
  journal   = {CoRR},
  volume    = {abs/1609.07769},
  year      = {2016},
  url       = {http://arxiv.org/abs/1609.07769},
  timestamp = {Mon, 03 Oct 2016 17:51:10 +0200},
  biburl    = {http://dblp.uni-trier.de/rec/bib/journals/corr/YangTFLGY16},
  bibsource = {dblp computer science bibliography, http://dblp.org}
}



@article{GAN_sisr,
  title={Photo-realistic single image super-resolution using a generative adversarial network},
  author={Ledig, Christian and Theis, Lucas and Husz{\'a}r, Ferenc and Caballero, Jose and Cunningham, Andrew and Acosta, Alejandro and Aitken, Andrew and Tejani, Alykhan and Totz, Johannes and Wang, Zehan and others},
  journal={arXiv preprint arXiv:1609.04802},
  year={2016}
}

@inproceedings{torch,
  title = {Torch7: A Matlab-like Environment for Machine Learning},
  author = {R. Collobert and K. Kavukcuoglu and C. Farabet},
  booktitle = {BigLearn, NIPS Workshop},
  year = {2011}
}

@inproceedings{denoise_auto,
  title={Image denoising and inpainting with deep neural networks},
  author={Xie, Junyuan and Xu, Linli and Chen, Enhong},
  booktitle={NIPS},
  pages={341--349},
  year={2012}
}

@article{VIF,
	title={Image information and visual quality},
	author={Sheikh, Hamid R and Bovik, Alan C},
	journal={IEEE TIP},
	volume={15},
	number={2},
	pages={430--444},
	year={2006},
	publisher={IEEE}
}


@article{SSIM,
	title={Image quality assessment: from error visibility to structural similarity},
	author={Wang, Zhou and Bovik, Alan C and Sheikh, Hamid R and Simoncelli, Eero P},
	journal={IEEE TIP},
	volume={13},
	number={4},
	pages={600--612},
	year={2004},
	publisher={IEEE}
}

@article{UIQ,
	title={A universal image quality index},
	author={Wang, Zhou and Bovik, Alan C},
	journal={IEEE Signal Processing Letters},
	volume={9},
	number={3},
	pages={81--84},
	year={2002},
	publisher={IEEE}
}


@inproceedings{perceptual_loss,
  title={Perceptual losses for real-time style transfer and super-resolution},
  author={Johnson, Justin and Alahi, Alexandre and Fei-Fei, Li},
  booktitle={European Conference on Computer Vision},
  pages={694--711},
  year={2016},
  organization={Springer}
}

@inproceedings{cnn_derain2,
  title={Restoring an image taken through a window covered with dirt or rain},
  author={Eigen, David and Krishnan, Dilip and Fergus, Rob},
  booktitle={ICCV},
  pages={633--640},
  year={2013}
}

@article{Gan_context,
  title={Context Encoders: Feature Learning by Inpainting},
  author={Pathak, Deepak and Krahenbuhl, Philipp and Donahue, Jeff and Darrell, Trevor and Efros, Alexei A},
  journal={arXiv preprint arXiv:1604.07379},
  year={2016}
}


@article{GAN_pix2pix2016,
  title={Image-to-Image Translation with Conditional Adversarial Networks},
  author={Isola, Phillip and Zhu, Jun-Yan and Zhou, Tinghui and Efros, Alexei A},
  journal={arxiv},
  year={2016}
}

@article{GAN_unsupervised,
  title={Unsupervised representation learning with deep convolutional generative adversarial networks},
  author={Radford, Alec and Metz, Luke and Chintala, Soumith},
  journal={arXiv preprint arXiv:1511.06434},
  year={2015}
}

@inproceedings{GAN,
  title={Generative adversarial nets},
  author={Goodfellow, Ian and Pouget-Abadie, Jean and Mirza, Mehdi and Xu, Bing and Warde-Farley, David and Ozair, Sherjil and Courville, Aaron and Bengio, Yoshua},
  booktitle={NIPS},
  pages={2672--2680},
  year={2014}
}

@article{nist4,
	title={NIST special database 4},
	author={Watson, CI and Wilson, CL},
	year={1992},
	publisher={Citeseer}
}

@inproceedings{fast_con,
  title={Fast convolutional sparse coding},
  author={Bristow, Hilton and Eriksson, Anders and Lucey, Simon},
  booktitle={CVPR},
  pages={391--398},
  year={2013}
}

@inproceedings{fast_flec,
  title={Fast and flexible convolutional sparse coding},
  author={Heide, Felix and Heidrich, Wolfgang and Wetzstein, Gordon},
  booktitle={CVPR},
  pages={5135--5143},
  year={2015},
  organization={IEEE}
}



@book{Meyer_200,
 author = {Meyer, Yves},
 title = {Oscillating Patterns in Image Processing and Nonlinear Evolution Equations: The Fifteenth Dean Jacqueline B. Lewis Memorial Lectures},
 year = {2001},
 isbn = {0821829203},
 publisher = {American Mathematical Society},
 address = {Boston, MA, USA},
} 


@Article{Vese2003,
author="Vese, Luminita A.
and Osher, Stanley J.",
title="Modeling Textures with Total Variation Minimization and Oscillating Patterns in Image Processing",
journal="Journal of Scientific Computing",
year="2003",
volume="19",
number="1",
pages="553--572"
}



@article{Shearlet_crack,
  author    = {Xavier Gibert and
               Vishal M. Patel and
               Demetrio Labate and
               Rama Chellappa},
  title     = {Discrete shearlet transform on {GPU} with applications in anomaly
               detection and denoising},
  journal   = {{EURASIP} Journal on Advances in Signal Processing},
  volume    = {2014},
  pages     = {64},
  year      = {2014}}

@inproceedings{deconv_segment,
	title={Learning deconvolution network for semantic segmentation},
	author={Noh, Hyeonwoo and Hong, Seunghoon and Han, Bohyung},
	booktitle={ICCV},
	pages={1520--1528},
	year={2015}
}

@inproceedings{image_guided,
	title={Guided image filtering},
	author={He, Kaiming and Sun, Jian and Tang, Xiaoou},
	booktitle={ECCV},
	pages={1--14},
	year={2010},
	organization={Springer}
}

@article{tv-norm,
	title={Edge-preserving and scale-dependent properties of total variation regularization},
	author={Strong, David and Chan, Tony},
	journal={Inverse problems},
	volume={19},
	number={6},
	pages={S165},
	year={2003},
	publisher={IOP Publishing}
}

@article{bsd_dataset,
	title={Contour detection and hierarchical image segmentation},
	author={Arbelaez, Pablo and Maire, Michael and Fowlkes, Charless and Malik, Jitendra},
	journal={IEEE Trans. on PAMI},
	volume={33},
	number={5},
	pages={898--916},
	year={2011},
	publisher={IEEE}
}
@inproceedings{gan,
  title={Generative adversarial nets},
  author={Goodfellow, Ian and Pouget-Abadie, Jean and Mirza, Mehdi and Xu, Bing and Warde-Farley, David and Ozair, Sherjil and Courville, Aaron and Bengio, Yoshua},
  booktitle={Advances in neural information processing systems},
  pages={2672--2680},
  year={2014}
}

@inproceedings{alex_net,
	title={Imagenet classification with deep convolutional neural networks},
	author={Krizhevsky, Alex and Sutskever, Ilya and Hinton, Geoffrey E},
	booktitle={NIPS},
	pages={1097--1105},
	year={2012}
}
@inproceedings{low_rank-re,
	title={Repairing sparse low-rank texture},
	author={Liang, Xiao and Ren, Xiang and Zhang, Zhengdong and Ma, Yi},
	booktitle={ECCV},
	pages={482--495},
	year={2012},
	organization={Springer}
}


@inproceedings{gmm_refec,
	title={Exploiting reflection change for automatic reflection removal},
	author={Yu, L and Brown, MS},
	booktitle={ICCV},
	pages={2432--2439},
	year={2013}
}

@inproceedings{ucid,
  title={UCID: an uncompressed color image database},
  author={Schaefer, Gerald and Stich, Michal},
  booktitle={Electronic Imaging 2004},
  pages={472--480},
  year={2003},
  organization={International Society for Optics and Photonics}
}


@article{cal_tech101,
	title={Learning generative visual models from few training examples: An incremental bayesian approach tested on 101 object categories},
	author={Fei-Fei, Li and Fergus, Rob and Perona, Pietro},
	journal={Computer Vision and Image Understanding},
	volume={106},
	number={1},
	pages={59--70},
	year={2007},
	publisher={Elsevier}
}

@article{adam_opt,
	title={Adam: A method for stochastic optimization},
	author={Kingma, Diederik and Ba, Jimmy},
	journal={arXiv preprint arXiv:1412.6980},
	year={2014}
}


@article{image_super_re_sp,
	title={Image super-resolution via sparse representation},
	author={Yang, Jianchao and Wright, John and Huang, Thomas S and Ma, Yi},
	journal={IEEE TIP},
	volume={19},
	number={11},
	pages={2861--2873},
	year={2010},
	publisher={IEEE}
}

@article{image_denoise_ksvd,
	title={Image denoising via sparse and redundant representations over learned dictionaries},
	author={Elad, Michael and Aharon, Michal},
	journal={IEEE TIP},
	volume={15},
	number={12},
	pages={3736--3745},
	year={2006},
	publisher={IEEE}
}

@inproceedings{cnn_su_acc,
	title={Accelerating the -resolution convolutional neural network},
	author={Dong, Chao and Loy, Chen Change and Tang, Xiaoou},
	booktitle={ECCV},
	pages={391--407},
	year={2016},
	organization={Springer}
}


@ARTICLE{cnn_derain,
	author = {{Fu}, X. and {Huang}, J. and {Ding}, X. and {Liao}, Y. and {Paisley}, J.
	},
	title = "{Clearing the Skies: A deep network architecture for single-image rain removal}",
	journal = {ArXiv e-prints},
	archivePrefix = "arXiv",
	eprint = {1609.02087},
	primaryClass = "cs.CV",
	keywords = {Computer Science - Computer Vision and Pattern Recognition},
	year = 2016,
	month = sep,
	adsurl = {http://adsabs.harvard.edu/abs/2016arXiv160902087F},
	adsnote = {Provided by the SAO/NASA Astrophysics Data System}
}

@article{image_srcnn,
	title={Image super-resolution using deep convolutional networks},
	author={Dong, Chao and Loy, Chen Change and He, Kaiming and Tang, Xiaoou},
	journal={arXiv preprint arXiv:1501.00092},
	year={2014}
}


@inproceedings{cnn_denoise_,
	title={Natural image denoising with convolutional networks},
	author={Jain, Viren and Seung, Sebastian},
	booktitle={NIPS},
	pages={769--776},
	year={2009}
}

@article{cnn_denois_gaussin,
	title={Beyond a Gaussian Denoiser: Residual Learning of Deep CNN for Image Denoising},
	author={Zhang, Kai and Zuo, Wangmeng and Chen, Yunjin and Meng, Deyu and Zhang, Lei},
	journal={arXiv preprint arXiv:1608.03981},
	year={2016}
}


@inproceedings{deep_residue,
  title={Deep residual learning for image recognition},
  author={He, Kaiming and Zhang, Xiangyu and Ren, Shaoqing and Sun, Jian},
  booktitle={Proceedings of the IEEE Conference on Computer Vision and Pattern Recognition},
  pages={770--778},
  year={2016}
}
@article{image_res_cnn,
	title={Image Denoising Using Very Deep Fully Convolutional Encoder-Decoder Networks with Symmetric Skip Connections},
	author={Mao, Xiao-Jiao and Shen, Chunhua and Yang, Yu-Bin},
	journal={arXiv preprint arXiv:1603.09056},
	year={2016}
}


@inproceedings{image_VDSR,
  title={Accurate image super-resolution using very deep convolutional networks},
  author={Kim, Jiwon and Kwon Lee, Jung and Mu Lee, Kyoung},
  booktitle={Proceedings of the IEEE Conference on Computer Vision and Pattern Recognition},
  pages={1646--1654},
  year={2016}
}


@article{SRCNN_su,
	title={Image super-resolution using deep convolutional networks},
	author={Dong, Chao and Loy, Chen Change and He, Kaiming and Tang, Xiaoou},
	journal={arXiv preprint arXiv:1501.00092},
	year={2014}
}

@article{imagede_total,
	title={Image decomposition and restoration using total variation minimization and the H 1},
	author={Osher, Stanley and Sol{\'e}, Andr{\'e}s and Vese, Luminita},
	journal={Multiscale Modeling \& Simulation},
	volume={1},
	number={3},
	pages={349--370},
	year={2003},
	publisher={SIAM}
}


@ARTICLE{fog_PAMI,
author={S. G. Narasimhan and S. K. Nayar},
journal={IEEE Trans. on PAMI},
title={Contrast restoration of weather degraded images},
year={2003},
volume={25},
number={6},
pages={713-724},
month={June}}



@ARTICLE{cnn_derain,
   author = {{Fu}, X. and {Huang}, J. and {Ding}, X. and {Liao}, Y. and {Paisley}, J.
	},
    title = "{Clearing the Skies: A deep network architecture for single-image rain removal}",
  journal = {ArXiv e-prints},
archivePrefix = "arXiv",
   eprint = {1609.02087},
 primaryClass = "cs.CV",
 keywords = {Computer Science - Computer Vision and Pattern Recognition},
     year = 2016,
    month = sep,
   adsurl = {http://adsabs.harvard.edu/abs/2016arXiv160902087F},
  adsnote = {Provided by the SAO/NASA Astrophysics Data System}
}


@article{haze_bayesian,
  title={Bayesian defogging},
  author={Nishino, Ko and Kratz, Louis and Lombardi, Stephen},
  journal={IJCV},
  volume={98},
  number={3},
  pages={263--278},
  year={2012},
  publisher={Springer}
}
@inproceedings{foggy,
  title={Factorizing scene albedo and depth from a single foggy image},
  author={Kratz, Louis and Nishino, Ko},
  booktitle={ICCV},
  pages={1701--1708},
  year={2009},
  organization={IEEE}
}

@inproceedings{image_separation_BMVC2016,
  title={Convolutional sparse coding-based image decomposition},
  author={H. Zhang and V. M. Patel},
  booktitle={BMVC},
  pages={},
  year={2016}
}


@inproceedings{dehaze_2012_yuli,
	title={Nighttime haze removal using color transfer pre-processing and dark channel prior},
	author={Pei, Soo-Chang and Lee, Tzu-Yen},
	booktitle={IEEE ICIP},
	pages={957--960},
	year={2012},
	organization={IEEE}
}
@inproceedings{snow_removal,
	title={Single-image-based rain and snow removal using multi-guided filter},
	author={Zheng, Xianhui and Liao, Yinghao and Guo, Wei and Fu, Xueyang and Ding, Xinghao},
	booktitle={International Conference on Neural Information Processing},
	pages={258--265},
	year={2013},
	organization={Springer}
}
@article{dehaz_2009_dark,
	title={Single image haze removal using dark channel prior},
	author={He, Kaiming and Sun, Jian and Tang, Xiaoou},
	journal={IEEE Trans. on PAMI},
	volume={33},
	number={12},
	pages={2341--2353},
	year={2011},
	publisher={IEEE}
}
@inproceedings{rain_video_2,
  title={Rain removal in video by combining temporal and chromatic properties},
  author={Zhang, Xiaopeng and Li, Hao and Qi, Yingyi and Leow, Wee Kheng and Ng, Teck Khim},
  booktitle={IEEE International Conference on Multimedia and Expo},
  pages={461--464},
  year={2006},
  organization={IEEE}
}
@inproceedings{rain_video1,
  title={Detection and removal of rain from videos},
  author={Garg, Kshitiz and Nayar, Shree K},
  booktitle={CVPR},
  volume={1},
  pages={I--528},
  organization={IEEE}
}
@article{rain_video3,
  title={Vision and rain},
  author={Garg, Kshitiz and Nayar, Shree K},
  journal={IJCV},
  volume={75},
  number={1},
  pages={3--27},
  year={2007},
  publisher={Springer}
}
@inproceedings{rain_kim_filter,
  title={Single-image deraining using an adaptive nonlocal means filter},
  author={Kim, Jin-Hwan and Lee, Chul and Sim, Jae-Young and Kim, Chang-Su},
  booktitle={ICIP},
  pages={914--917},
  year={2013},
  organization={IEEE}
}

@inproceedings{rain_struc,
  title={Exploiting image structural similarity for single image rain removal},
  author={Sun, Shao-Hua and Fan, Shang-Pu and Wang, Yu-Chiang Frank},
  booktitle={IEEE ICIP},
  pages={4482--4486},
  year={2014},
  organization={IEEE}
}

@article{self_spa_rain,
  title={Self-learning based image decomposition with applications to single image denoising},
  author={Huang, De-An and Kang, Li-Wei and Wang, Yu-Chiang Frank and Lin, Chia-Wen},
  journal={IEEE Transactions on Multimedia},
  volume={16},
  number={1},
  pages={83--93},
  year={2014},
  publisher={IEEE}
}

@inproceedings{rain_2016_gmm,
	title={Rain Streak Removal Using Layer Priors},
	author={Li, Yu and Tan, Robby T and Guo, Xiaojie and Lu, Jiangbo and Brown, Michael S},
	booktitle={CVPR},
	pages={2736--2744},
	year={2016}
}

@article{ssim,
	title={Image quality assessment: from error visibility to structural similarity},
	author={Wang, Zhou and Bovik, Alan C and Sheikh, Hamid R and Simoncelli, Eero P},
	journal={IEEE TIP},
	volume={13},
	number={4},
	pages={600--612},
	year={2004},
	publisher={IEEE}
}

@inproceedings{dis_rain_2015,
  title={Removing rain from a single image via discriminative sparse coding},
  author={Luo, Yu and Xu, Yong and Ji, Hui},
  booktitle={ICCV},
  pages={3397--3405},
  year={2015}
}

@inproceedings{low_rank_rain,
  title={A generalized low-rank appearance model for spatio-temporally correlated rain streaks},
  author={Chen, Yi-Lei and Hsu, Chiou-Ting},
  booktitle={IEEE ICCV},
  pages={1968--1975},
  year={2013}
}

@article{automatic_rain,
  title={Automatic single-image-based rain streaks removal via image decomposition},
  author={Kang, Li-Wei and Lin, Chia-Wen and Fu, Yu-Hsiang},
  journal={IEEE TIP},
  volume={21},
  number={4},
  pages={1742--1755},
  year={2012},
  publisher={IEEE}
}

@article{proximal_boyd,
 author = {Parikh, Neal and Boyd, Stephen},
 title = {Proximal Algorithms},
 journal = {Foundations and Trends in Optimization},
 volume = {1},
 number = {3},
 month = {jan},
 year = {2014},
 pages = {127--239}
} 


@inproceedings{vishalrobust,
  title={Robust multimodal recognition via multitask multivariate low-rank representations},
  author={Zhang, Heng and Patel, Vishal M and Chellappa, Rama},
  booktitle={Automatic Face and Gesture Recognition (FG), 2015 11th IEEE International Conference and Workshops on},
  volume={1},
  pages={1--8},
  year={2015},
  organization={IEEE}
}
@inproceedings{salient_detection,
  title={A unified approach to salient object detection via low rank matrix recovery},
  author={Shen, Xiaohui and Wu, Ying},
  booktitle={CVPR},
  pages={853--860},
  year={2012},
  organization={IEEE}
}
@inproceedings{face_rank,
  title={Sparse representation for face recognition based on discriminative low-rank dictionary learning},
  author={Ma, Long and Wang, Chunheng and Xiao, Baihua and Zhou, Wen},
  booktitle={CVPR},
  pages={2586--2593},
  year={2012},
  organization={IEEE}
}
@inproceedings{image_seg,
  title={Multi-task low-rank affinity pursuit for image segmentation},
  author={Cheng, Bin and Liu, Guangcan and Wang, Jingdong and Huang, Zhongyang and Yan, Shuicheng},
  booktitle={ICCV},
  pages={2439--2446},
  year={2011},
  organization={IEEE}
}
@article{rasl,
  title={RASL: Robust alignment by sparse and low-rank decomposition for linearly correlated images},
  author={Peng, Yigang and Ganesh, Arvind and Wright, John and Xu, Wenli and Ma, Yi},
  journal={IEEE Trans. on PAMI},
  volume={34},
  number={11},
  pages={2233--2246},
  year={2012},
  publisher={IEEE}
}
@article{rank_nuclear,
  title={Guaranteed minimum-rank solutions of linear matrix equations via nuclear norm minimization},
  author={Recht, Benjamin and Fazel, Maryam and Parrilo, Pablo A},
  journal={SIAM review},
  volume={52},
  number={3},
  pages={471--501},
  year={2010},
  publisher={SIAM}
}
@article{robust_pca,
  title={Robust principal component analysis?},
  author={Cand{\`e}s, Emmanuel J and Li, Xiaodong and Ma, Yi and Wright, John},
  journal={Journal of the ACM (JACM)},
  volume={58},
  number={3},
  pages={11},
  year={2011},
  publisher={ACM}
}
@article{robust_subspace,
  title={Robust recovery of subspace structures by low-rank representation},
  author={Liu, Guangcan and Lin, Zhouchen and Yan, Shuicheng and Sun, Ju and Yu, Yong and Ma, Yi},
  journal={IEEE Trans. on PAMI},
  volume={35},
  number={1},
  pages={171--184},
  year={2013},
  publisher={IEEE}
}
@inproceedings{dic_low_rank,
  title={Learning structured low-rank representations for image classification},
  author={Zhang, Yangmuzi and Jiang, Zhuolin and Davis, Larry},
  booktitle={CVPR},
  pages={676--683},
  year={2013}
}


@article{svt_nuclear,
  title={A singular value thresholding algorithm for matrix completion},
  author={Cai, Jian-Feng and Cand{\`e}s, Emmanuel J and Shen, Zuowei},
  journal={SIAM Journal on Optimization},
  volume={20},
  number={4},
  pages={1956--1982},
  year={2010},
  publisher={SIAM}
}

@Inbook{Aujol2003,
author="Aujol, Jean-Fran{\c{c}}ois
and Aubert, Gilles
and Blanc-F{\'e}raud, Laure
and Chambolle, Antonin",
editor="Griffin, Lewis D.
and Lillholm, Martin",
chapter="Image Decomposition Application to SAR Images",
title="Scale Space Methods in Computer Vision",
year="2003",
publisher="Springer Berlin Heidelberg",
address="Berlin, Heidelberg",
pages="297--312",
isbn="978-3-540-44935-5"
}




@article{Shearlet_crack,
  author    = {Xavier Gibert and
               Vishal M. Patel and
               Demetrio Labate and
               Rama Chellappa},
  title     = {Discrete shearlet transform on {GPU} with applications in anomaly
               detection and denoising},
  journal   = {{EURASIP} Journal on Advances in Signal Processing},
  volume    = {2014},
  pages     = {64},
  year      = {2014}}
  
  @INPROCEEDINGS{ACSC_feature_learning_ICCV2011,
author={M. D. Zeiler and G. W. Taylor and R. Fergus},
booktitle={ICCV},
title={Adaptive deconvolutional networks for mid and high level feature learning},
year={2011},
pages={2018-2025},
month={Nov}}



@inproceedings{decon_net,
  title={Deconvolutional networks},
  author={Zeiler, Matthew D and Krishnan, Dilip and Taylor, Graham W and Fergus, Rob},
  booktitle={CVPR},
  pages={2528--2535},
  year={2010},
  organization={IEEE}
}

@article{vishal_SAR_separa,
  title={Separated component-based restoration of speckled SAR images},
  author={Patel, Vishal M and Easley, Glenn R and Chellappa, Rama and Nasrabadi, Nasser M},
  journal={ IEEE Transactions on Geoscience and Remote Sensing},
  volume={52},
  number={2},
  pages={1019--1029},
  year={2014},
  publisher={IEEE}
}

@INPROCEEDINGS{CSC_SR_ICCV2015,
author={S. Gu and W. Zuo and Q. Xie and D. Meng and X. Feng and L. Zhang},
booktitle={ICCV},
title={Convolutional Sparse Coding for Image Super-Resolution},
year={2015},
pages={1823-1831},
month={Dec}}

@article{TV_Haar,
  author    = {Gabriele Steidl and
               Joachim Weickert and
               Thomas Brox and
               Pavel Mr{\'{a}}zek and
               Martin Welk},
  title     = {On the Equivalence of Soft Wavelet Shrinkage, Total Variation Diffusion,
               Total Variation Regularization, and SIDEs},
  journal   = {{SIAM} J. Numerical Analysis},
  volume    = {42},
  number    = {2},
  pages     = {686--713},
  year      = {2004}}
  
  @ARTICLE{Donoho93idealspatial,
    author = {D. Donoho and I. Johnstone and Iain M. Johnstone},
    title = {Ideal Spatial Adaptation by Wavelet Shrinkage},
    journal = {Biometrika},
    year = {1993},
    volume = {81},
    pages = {425--455}
}



@ARTICLE{CSC_TIP2016,
author={Wohlberg, Brendt},
journal={IEEE Transactions on Image Processing},
title={Efficient Algorithms for Convolutional Sparse Representations},
year={2016},
volume={25},
number={1},
pages={301-315},
month={Jan}}


@ARTICLE{MCA_ATS,
author={J. Bobin and J. L. Starck and J. M. Fadili and Y. Moudden and D. L. Donoho},
journal={IEEE Transactions on Image Processing},
title={Morphological Component Analysis: An Adaptive Thresholding Strategy},
year={2007},
volume={16},
number={11},
pages={2675-2681},
month={Nov}}

@book{Elad_book_2010,
 author = {Elad, Michael},
 title = {Sparse and Redundant Representations: From Theory to Applications in Signal and Image Processing},
 year = {2010},
 isbn = {144197010X, 9781441970107},
 edition = {1st},
 publisher = {Springer}
} 

@ARTICLE{ksvd,
author={M. Aharon and M. Elad and A. Bruckstein},
journal={IEEE Transactions on Signal Processing},
title={\rm K -SVD: An Algorithm for Designing Overcomplete Dictionaries for Sparse Representation},
year={2006},
volume={54},
number={11},
pages={4311-4322},
month={Nov}}


@article{admm,
  title={Distributed optimization and statistical learning via the alternating direction method of multipliers},
  author={Boyd, Stephen and Parikh, Neal and Chu, Eric and Peleato, Borja and Eckstein, Jonathan},
  journal={Foundations and Trends{\textregistered} in Machine Learning},
  volume={3},
  number={1},
  pages={1--122},
  year={2011},
  publisher={Now Publishers Inc.}
}
@article{wright2009robust,
  title={Robust face recognition via sparse representation},
  author={Wright, John and Yang, Allen Y and Ganesh, Arvind and Sastry, Shankar S and Ma, Yi},
  journal={IEEE Trans. on PAMI},
  volume={31},
  number={2},
  pages={210--227},
  year={2009},
  publisher={IEEE}
}
@inproceedings{vishal_clust,
  title={Latent space sparse subspace clustering},
  author={Patel, Vishal and Nguyen, Hien and Vidal, Ren{\'e}},
  booktitle={ICCV},
  pages={225--232},
  year={2013}
}
@inproceedings{classification,
  title={Analysis sparse coding models for image-based classification},
  author={Shekhar, Shashi and Patel, Vishal M and Chellappa, Rama},
  booktitle={IEEE ICIP},
  pages={5207--5211},
  year={2014},
  organization={IEEE}
}
@article{image_denoise,
  title={Image denoising via sparse and redundant representations over learned dictionaries},
  author={Elad, Michael and Aharon, Michal},
  journal={IEEE TIP},
  volume={15},
  number={12},
  pages={3736--3745},
  year={2006},
  publisher={IEEE}
}
@article{proximal,
  title={Proximal Algorithms.},
  author={Parikh, Neal and Boyd, Stephen P}
}
@article{finger,
  title={Filterbank-based fingerprint matching},
  author={Jain, Anil K and Prabhakar, Salil and Hong, Lin and Pankanti, Sharath},
  journal={IEEE TIP},
  volume={9},
  number={5},
  pages={846--859},
  year={2000},
  publisher={IEEE}
}

@article{trajectory,
  title={Convolutional sparse coding for trajectory reconstruction},
  author={Zhu, Yingying and Lucey, Simon},
  journal={IEEE Trans. on PAMI},
  volume={37},
  number={3},
  pages={529--540},
  year={2015},
  publisher={IEEE}
}



@inproceedings{con_super,
  title={Convolutional Sparse Coding for Image Super-Resolution},
  author={Gu, Shuhang and Zuo, Wangmeng and Xie, Qi and Meng, Deyu and Feng, Xiangchu and Zhang, Lei},
  booktitle={ICCV},
  pages={1823--1831},
  year={2015}
}
@article{luan2017deep,
  title={Deep photo style transfer},
  author={Luan, Fujun and Paris, Sylvain and Shechtman, Eli and Bala, Kavita}
}

@article{multichannel_survey,
  title={The multichannel deconvolution problem: a discrete analysis},
  author={Colonna, Flavia and Easley, Glenn R},
  journal={Journal of Fourier Analysis and Applications},
  volume={10},
  number={4},
  pages={351--376},
  year={2004},
  publisher={Springer}
}
@article{zhu2017unpaired,
  title={Unpaired image-to-image translation using cycle-consistent adversarial networks},
  author={Zhu, Jun-Yan and Park, Taesung and Isola, Phillip and Efros, Alexei A}
}

@inproceedings{msyn,
  title={An M-channel directional filter bank compatible with the contourlet and shearlet frequency tiling},
  author={Easley, Glenn R and Patel, Vishal and Healy Jr, Dennis M},
  booktitle={Optical Engineering+ Applications},
  pages={67010C--67010C},
  year={2007},
  organization={International Society for Optics and Photonics}
}

@article{tv_norm,
  title={Nonlinear total variation based noise removal algorithms},
  author={Rudin, Leonid I and Osher, Stanley and Fatemi, Emad},
  journal={Physica D: Nonlinear Phenomena},
  volume={60},
  number={1},
  pages={259--268},
  year={1992},
  publisher={Elsevier}
}
@article{lowrank_path,
  title={A low patch-rank interpretation of texture},
  author={Schaeffer, Hayden and Osher, Stanley},
  journal={SIAM Journal on Imaging Sciences},
  volume={6},
  number={1},
  pages={226--262},
  year={2013},
  publisher={SIAM}
}

@article{robust_pca,
  title={Robust principal component analysis?},
  author={Cand{\`e}s, Emmanuel J and Li, Xiaodong and Ma, Yi and Wright, John},
  journal={Journal of the ACM (JACM)},
  volume={58},
  number={3},
  pages={11},
  year={2011},
  publisher={ACM}
}

@inproceedings{fast_con,
  title={Fast convolutional sparse coding},
  author={Bristow, Hilton and Eriksson, Anders and Lucey, Simon},
  booktitle={CVPR},
  pages={391--398},
  year={2013}
}


@book{Meyer_200,
 author = {Meyer, Yves},
 title = {Oscillating Patterns in Image Processing and Nonlinear Evolution Equations: The Fifteenth Dean Jacqueline B. Lewis Memorial Lectures},
 year = {2001},
 isbn = {0821829203},
 publisher = {American Mathematical Society},
 address = {Boston, MA, USA},
} 


@Article{Vese2003,
author="Vese, Luminita A.
and Osher, Stanley J.",
title="Modeling Textures with Total Variation Minimization and Oscillating Patterns in Image Processing",
journal="Journal of Scientific Computing",
year="2003",
volume="19",
number="1",
pages="553--572"
}





@Inbook{Osendorfer2014,
author="Osendorfer, Christian
and Soyer, Hubert
and van der Smagt, Patrick",
editor="Loo, Chu Kiong
and Yap, Keem Siah
and Wong, Kok Wai
and Beng Jin, Andrew Teoh
and Huang, Kaizhu",
chapter="Image Super-Resolution with Fast Approximate Convolutional Sparse Coding",
title="Neural Information Processing: 21st International Conference, ICONIP 2014, Kuching, Malaysia, November 3-6, 2014. Proceedings, Part III",
year="2014",
publisher="Springer International Publishing",
address="Cham",
pages="250--257"}



@inproceedings{wohlberg2014efficient,
  title={Efficient convolutional sparse coding},
  author={Wohlberg, Brendt},
  booktitle={IEEE ICASSP},
  pages={7173--7177},
  year={2014},
  organization={IEEE}
}
@inproceedings{fast_flec,
  title={Fast and flexible convolutional sparse coding},
  author={Heide, Felix and Heidrich, Wolfgang and Wetzstein, Gordon},
  booktitle={CVPR},
  pages={5135--5143},
  year={2015},
  organization={IEEE}
}

@article{bristow2014optimization,
  title={Optimization methods for convolutional sparse coding},
  author={Bristow, Hilton and Lucey, Simon},
  journal={arXiv preprint arXiv:1406.2407},
  year={2014}
}

@article{mca_image,
  title={Image decomposition via the combination of sparse representations and a variational approach},
  author={Starck, Jean-Luc and Elad, Michael and Donoho, David L},
  journal={IEEE TIP},
  volume={14},
  number={10},
  pages={1570--1582},
  year={2005},
  publisher={IEEE}
}
@article{mca_diversity,
  title={Learning the morphological diversity},
  author={Peyr{\'e}, Gabriel and Fadili, Jalal and Starck, Jean-Luc},
  journal={SIAM Journal on Imaging Sciences},
  volume={3},
  number={3},
  pages={646--669},
  year={2010},
  publisher={SIAM}
}

@article{tvnorm,
  title={Image decomposition and restoration using total variation minimization and the H 1},
  author={Osher, Stanley and Sol{\'e}, Andr{\'e}s and Vese, Luminita},
  journal={Multiscale Modeling \& Simulation},
  volume={1},
  number={3},
  pages={349--370},
  year={2003},
  publisher={SIAM}
}

@article{BNN_rank,
  title={Cartoon-texture image decomposition using blockwise low-rank texture characterization},
  author={Ono, Shintaro and Miyata, Takamichi and Yamada, Isao},
  journal={IEEE TIP},
  volume={23},
  number={3},
  pages={1128--1142},
  year={2014},
  publisher={IEEE}
}

@article{mirza2014conditional,
  title={Conditional generative adversarial nets},
  author={Mirza, Mehdi and Osindero, Simon},
  journal={arXiv preprint arXiv:1411.1784},
  year={2014}
}

@article{thibodeau2014deep,
  title={Deep generative stochastic networks trainable by backprop},
  author={Thibodeau-Laufer, Eric and Alain, Guillaume and Yosinski, Jason},
  year={2014}
}

@article{karacan2016learning,
  title={Learning to Generate Images of Outdoor Scenes from Attributes and Semantic Layouts},
  author={Karacan, Levent and Akata, Zeynep and Erdem, Aykut and Erdem, Erkut},
  journal={arXiv preprint arXiv:1612.00215},
  year={2016}
}

@inproceedings{salimans2016improved,
  title={Improved techniques for training gans},
  author={Salimans, Tim and Goodfellow, Ian and Zaremba, Wojciech and Cheung, Vicki and Radford, Alec and Chen, Xi},
  booktitle={NIPS},
  pages={2226--2234},
  year={2016}
}
@article{creswell2016task,
  title={Task Specific Adversarial Cost Function},
  author={Creswell, Antonia and Bharath, Anil A},
  journal={arXiv preprint arXiv:1609.08661},
  year={2016}
}

@article{zhao2016energy,
  title={Energy-based generative adversarial network},
  author={Zhao, Junbo and Mathieu, Michael and LeCun, Yann},
  journal={arXiv preprint arXiv:1609.03126},
  year={2016}
}
@article{jetchev2016texture,
  title={Texture Synthesis with Spatial Generative Adversarial Networks},
  author={Jetchev, Nikolay and Bergmann, Urs and Vollgraf, Roland},
  journal={arXiv preprint arXiv:1611.08207},
  year={2016}
}

@inproceedings{long2015fully,
  title={Fully convolutional networks for semantic segmentation},
  author={Long, Jonathan and Shelhamer, Evan and Darrell, Trevor},
  booktitle={CVPR},
  pages={3431--3440},
  year={2015}
}
@inproceedings{eigen2015predicting,
  title={Predicting depth, surface normals and semantic labels with a common multi-scale convolutional architecture},
  author={Eigen, David and Fergus, Rob},
  booktitle={ICCV},
  pages={2650--2658},
  year={2015}
}
@inproceedings{	mahendran2015understanding,
  title={Understanding deep image representations by inverting them},
  author={Mahendran, Aravindh and Vedaldi, Andrea},
  booktitle={CVPR},
  pages={5188--5196},
  year={2015},
  organization={IEEE}
}

@inproceedings{zheng2015conditional,
  title={Conditional random fields as recurrent neural networks},
  author={Zheng, Shuai and Jayasumana, Sadeep and Romera-Paredes, Bernardino and Vineet, Vibhav and Su, Zhizhong and Du, Dalong and Huang, Chang and Torr, Philip HS},
  booktitle={ICCV},
  pages={1529--1537},
  year={2015}
}

@article{dosovitskiy2015inverting,
  title={Inverting visual representations with convolutional networks},
  author={Dosovitskiy, Alexey and Brox, Thomas},
  journal={arXiv preprint arXiv:1506.02753},
  year={2015}
}

@article{gatys2015neural,
  title={A neural algorithm of artistic style},
  author={Gatys, Leon A and Ecker, Alexander S and Bethge, Matthias},
  journal={arXiv preprint arXiv:1508.06576},
  year={2015}
}

@inproceedings{li2016precomputed,
  title={Precomputed Real-Time Texture Synthesis with Markovian Generative Adversarial Networks},
  author={Li, Chuan and Wand, Michael},
  booktitle={ECCV},
  pages={702--716},
  year={2016}
}

@inproceedings{pathak2016context,
  title={Context Encoders: Feature Learning by Inpainting},
  author={Pathak, Deepak and Krahenbuhl, Philipp and Donahue, Jeff and Darrell, Trevor and Efros, Alexei A},
  booktitle={CVPR},
  year={2016}
}

@article{reed2016generative,
  title={Generative adversarial text to image synthesis},
  author={Reed, Scott and Akata, Zeynep and Yan, Xinchen and Logeswaran, Lajanugen and Schiele, Bernt and Lee, Honglak},
  journal={arXiv preprint arXiv:1605.05396},
  year={2016}
}

\end{document}